\documentclass{article} 
\usepackage{iclr2021_conference,times}

\usepackage{amsmath,amsfonts,bm}

\def\eqref#1{equation~\ref{#1}}

\def\1{\bm{1}}

\DeclareMathAlphabet{\mathsfit}{\encodingdefault}{\sfdefault}{m}{sl}
\SetMathAlphabet{\mathsfit}{bold}{\encodingdefault}{\sfdefault}{bx}{n}

\newcommand{\R}{\mathbb{R}}

\newcommand{\Cov}{\mathrm{Cov}}

\definecolor{Red}{rgb}{1,0,0}
\definecolor{Green}{rgb}{0,0.7,0}
\definecolor{Blue}{rgb}{0,0,1}
\definecolor{Red}{rgb}{0.6,0,0}
\definecolor{Orange}{rgb}{1,0.5,0}

\def\B#1{\mathbf{#1}}
\def\C#1{\mathcal{#1}}
\def\R#1{\mathrm{#1}}

\usepackage{hyperref}
\usepackage{url}
\usepackage{booktabs}
\usepackage{graphicx}
\usepackage{tabularx}
\usepackage{subcaption}
\usepackage{xargs}
\usepackage{array}
\usepackage{tikz}
\usepackage{mathtools}
\usetikzlibrary{automata,decorations.markings,positioning,arrows}

\title{Explaining the Efficacy of 
Counterfactually Augmented Data}

\author{Divyansh Kaushik, Amrith Setlur, Eduard Hovy, Zachary C. Lipton\\ 
Carnegie Mellon University\\
Pittsburgh, PA, USA\\
\{\texttt{dkaushik, asetlur, hovy, zlipton\}@cmu.edu}}

\newcommand{\expnumber}[2]{{#1}\mathrm{e}{#2}}

\iclrfinalcopy 
\begin{document}
\newcommand{\ind}{\perp\!\!\!\!\perp} 

\maketitle

\begin{abstract}
In attempts to produce machine learning models
less reliant on spurious patterns in NLP datasets,
researchers have recently proposed 
curating counterfactually augmented data (CAD)
via a human-in-the-loop process
in which given some documents 
and their (initial) labels,
humans must revise the text 
to make a counterfactual label applicable. 
Importantly, edits that are not necessary 
to flip the applicable label are prohibited.
Models trained on the augmented 
(original \emph{and} revised) data
appear, empirically, to rely less on 
semantically irrelevant words
and to generalize better out of domain.
While this work draws loosely on causal thinking,
the underlying causal model (even at an abstract level)
and the principles underlying the 
observed out-of-domain improvements remain unclear.
In this paper, we introduce a toy analog
based on linear Gaussian models,
observing interesting relationships 
between causal models, measurement noise,
out-of-domain generalization, 
and reliance on spurious signals.
Our analysis provides some insights
that help to explain the efficacy of CAD.
Moreover, we develop the hypothesis 
that while adding noise to causal features 
should degrade both in-domain 
and out-of-domain performance,
adding noise to non-causal features 
should lead to relative improvements 
in out-of-domain performance.
This idea inspires a speculative test 
for determining whether a feature attribution
technique has identified the \emph{causal spans}. 
If adding noise (e.g., by random word flips)
to the highlighted spans degrades both in-domain
and out-of-domain performance
on a battery of challenge datasets,
but adding noise to the complement 
gives improvements out-of-domain,
this suggests we have identified causal spans.  
Thus, we present a large-scale empirical study 
comparing spans edited to create CAD
to those selected by attention and saliency maps.
Across numerous challenge domains and models,
we find that the hypothesized phenomenon 
is pronounced for CAD. 

\end{abstract}

\section{Introduction}
Despite machine learning (ML)'s many practical breakthroughs,
formidable obstacles obstruct its deployment 
in consequential applications. 
Of particular concern, 
these models have been shown
to rely on spurious signals, 
such as surface-level textures in images
\citep{jo2017measuring, geirhos2018imagenet},
and background scenery---even when the task 
is to recognize foreground objects \citep{beery2018recognition}.
Other studies have uncovered 
a worrisome reliance on gender
in models trained for the purpose
of recommending jobs \citep{dastin2018amazon}, 
and on race in prioritizing patients 
for medical care \citep{obermeyer2019dissecting}. 
Moreover, while modern ML performs remarkably well on 
independent and identically distributed (iid) holdout data,
performance often decays catastrophically 
under both naturally occurring 
and adversarial distribution shift
\citep{quionero2009dataset, sugiyama2012machine, szegedy2013intriguing, ovadia2019can, Filos2020CanAV}.

These two problems: 
(i) reliance on semantically irrelevant signals, 
raising concerns about bias;
and (ii) the brittleness of models under distributions shift; 
might appear unrelated, 
but share important conceptual features. 
Concerns about bias
stem in part from principles of procedural fairness \citep{blader2003constitutes,sep-justice,grgic2018beyond,lipton2018does},
according to which decisions 
should be based on qualifications,
not on distant proxies 
that are spuriously associated with the outcome of interest.
Arguably one key distinction of an actual qualification 
might be that it actually exerts causal influence 
on the outcome of interest. 
In an interesting parallel, 
one line of work on distribution shift
has focused on causal graphical models,
addressing settings 
where some parts of the model
remain stable over time but others do not.
One common assumption is that the
relationship between the target 
and its direct causal ancestors
remains invariant
\citep{peters2016causal,ghassami2017learning,rojas2018invariant,kuang2018stable,magliacane2018domain,christiansen2020switching,weichwald2020distributional}.
While these papers contribute insight,
they focus on toy settings,
with few variables related by a known model.
However, in complex domains with high-dimensional data, 
what variables are relevant 
and what graph relates them
is typically unclear.

Recently in NLP, \citet{kaushik2020learning}
proposed \emph{Counterfactually Augmented Data (CAD)},
injecting causal thinking into real world settings
by leveraging human-in-the-loop feedback 
to identify causally relevant features
(versus those that merely happen
to be predictive due to confounding). 
Human editors are presented with document-label pairs
and tasked with editing documents
to render counterfactual labels applicable. 
The instructions restrict editors 
to only make modifications that are necessary 
to flip the label's applicability. 
The key result is that many spurious correlations 
present in the original dataset are absent in the CAD.
In case of sentiment analysis, 
\citet{kaushik2020learning} demonstrated 
that linear classifiers trained 
to predict the sentiment of movie reviews
based on bag-of-words representations 
assign high-magnitude weights 
to seemingly irrelevant terms,
including ``will'', ``my'', ``has'', 
``especially'', and ``script'', among others. 
Notably, ``horror'' featured among the most negative terms,
while ``romance'' featured among the most positive, 
despite both communicating genre, not sentiment. 
Interestingly, in the revised data,
each ``horror'' review retains the word ``horror''
(per the instruction not to make unnecessary edits)
but is associated with the opposite sentiment label. 
Models trained on the augmented data 
(original \emph{and} revised)
perform well on both original and revised data,
and assign little weight to the associated but irrelevant terms.
Intuitively, one might imagine 
that the spurious patterns 
would generalize less reliably out of domain. 
Most consumer products do not belong to movie genres,
but words like ``excellent'' and ``awful''
continue to connote positive and negative sentiment, respectively.
Indeed, \citet{kaushik2020learning} demonstrated 
that models trained on CAD enjoyed 
out-of-domain performance benefits 
on Tweets, and Amazon and Yelp reviews.

In this paper, 
we make some initial attempts 
towards explaining CAD's efficacy.
While CAD plainly draws on causal thinking,
(invoking interventions and counterfactuals), 
foundational questions remain open:
What is the assumed causal structure underlying 
settings where CAD might be effective?
What are the principles underlying 
its out-of-domain benefits? 
Must humans \emph{really} intervene,
or could automatic feature attribution methods,
e.g., attention \citep{deyoung2020eraser}, 
or cheaper feedback mechanisms,
e.g., feature feedback \citep{zaidan2007using},
produce similar results?

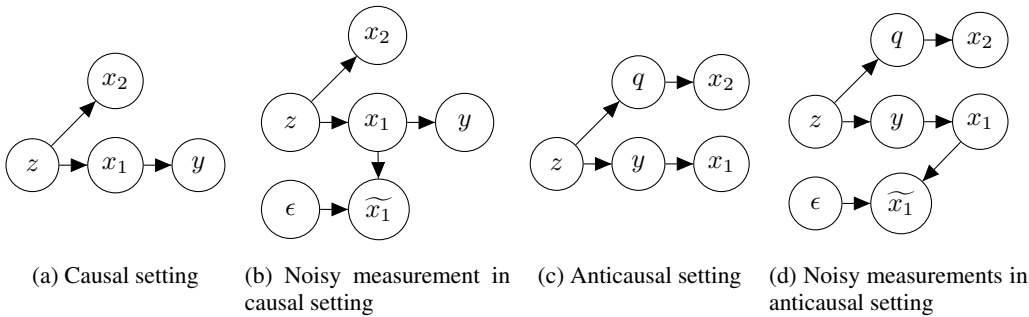
\begin{figure}[t]
\center
\subfloat[Causal setting]{
	\begin{minipage}[c][1\width]{
	   0.24\textwidth}
	   \centering
	   \begin{tikzpicture}[
        midarr/.style 2 args={
          decoration={
            markings,
            mark=at position 1 with {\arrow[xshift=0pt]{triangle 45} \node[#1] {#2};},
          },
            postaction=decorate,
        },
        state/.append style={minimum size=20pt},
        node distance=0.37cm and 0.37cm
        ]
        \node[state]
          (Z) {$z$};
        \node[state, right=of Z]
          (X1) {$x_1$};
        \node[state, above=of X1]
          (X2) {$x_2$};
        \node[state,right=of X1]
          (Y) {$y$};
        
        \draw[midarr={right=2pt}{}]
          (Z) -- (X1);  
        \draw[midarr={right=2pt}{}]
          (Z) -- (X2);  
        \draw[midarr={right=2pt}{}]
          (X1) -- (Y);
        \end{tikzpicture}
        \label{fig:causal_diagram}
	\end{minipage}}
 \subfloat[Noisy measurement in causal setting]{
    \begin{minipage}[c][1\width]{
	   0.24\textwidth}
	   \centering
	   \begin{tikzpicture}[
        midarr/.style 2 args={
          decoration={
            markings,
            mark=at position 1 with {\arrow[xshift=0pt]{triangle 45} \node[#1] {#2};},
          },
            postaction=decorate,
        },
        state/.append style={minimum size=22pt},
        node distance=0.37cm and 0.37cm
        ]
        \node[state]
          (e) {$\epsilon$};
        \node[state, above=of e]
          (Z) {$z$};
        \node[state, right=of Z]
          (X1) {$x_1$};
        \node[state, above=of X1]
          (X2) {$x_2$};
        \node[state,right=of X1]
          (Y) {$y$};
        \node[state,right=of e]
          (X1t) {$\widetilde{x_1}$};

        \draw[midarr={right=2pt}{}]
          (Z) -- (X1);  
        \draw[midarr={right=2pt}{}]
          (Z) -- (X2);  
        \draw[midarr={right=2pt}{}]
          (X1) -- (Y);
        \draw[midarr={right=2pt}{}]
          (X1) -- (X1t);
        \draw[midarr={right=2pt}{}]
          (e) -- (X1t);
        \end{tikzpicture}
        \label{fig:noise_causal_diagram}
	\end{minipage}
 }
  \subfloat[Anticausal setting]{
	\begin{minipage}[c][1\width]{
	   0.24\textwidth}
	   \centering
	   \begin{tikzpicture}[
        midarr/.style 2 args={
          decoration={
            markings,
            mark=at position 1 with {\arrow[xshift=0pt]{triangle 45} \node[#1] {#2};},
          },
            postaction=decorate,
        },
        state/.append style={minimum size=20pt},
        node distance=0.37cm and 0.37cm
        ]
        \node[state]
          (Z) {$z$};
        \node[state, right=of Z]
          (Y) {$y$};
        \node[state, above=of Y]
          (Q) {$q$};
        \node[state,right=of Y]
          (X1) {$x_1$};
        \node[state,right=of Q]
          (X2) {$x_2$};
        
        \draw[midarr={right=2pt}{}]
          (Z) -- (Y);  
        \draw[midarr={right=2pt}{}]
          (Z) -- (Q);  
        \draw[midarr={right=2pt}{}]
          (Y) -- (X1);
        \draw[midarr={right=2pt}{}]
          (Q) -- (X2);
        \end{tikzpicture}
        \label{fig:anticausal_diagram}
	\end{minipage}}
    \subfloat[Noisy measurements in anticausal setting]{
	\begin{minipage}[c][1\width]{
	   0.24\textwidth}
	   \centering
	   \begin{tikzpicture}[
        midarr/.style 2 args={
          decoration={
            markings,
            mark=at position 1 with {\arrow[xshift=0pt]{triangle 45} \node[#1] {#2};},
          },
            postaction=decorate,
        },
        state/.append style={minimum size=20pt},
        node distance=0.37cm and 0.37cm
        ]
        \node[state]
          (e) {$\epsilon$};
        \node[state, above=of e]
          (Z) {$z$};
        \node[state,right=of Z]
          (Y) {$y$};
        \node[state, above=of Y]
          (Q) {$q$};
        \node[state,right=of Y]
          (X1) {$x_1$};
        \node[state,right=of Q]
          (X2) {$x_2$};
        \node[state,right=of e]
          (X1t) {$\widetilde{x_1}$};
        
        \draw[midarr={right=2pt}{}]
          (Z) -- (Y);  
        \draw[midarr={right=2pt}{}]
          (Z) -- (Q);  
        \draw[midarr={right=2pt}{}]
          (Y) -- (X1);
        \draw[midarr={right=2pt}{}]
          (Q) -- (X2);
        \draw[midarr={right=2pt}{}]
          (e) -- (X1t);
        \draw[midarr={right=2pt}{}]
          (X1) -- (X1t);
        \end{tikzpicture}
        \label{fig:noisy_anticausal_diagram}
	\end{minipage}}
\caption{Toy causal models with one hidden confounder.
In \ref{fig:causal_diagram} and \ref{fig:anticausal_diagram}, 
the observed covariates are $x_1, x_2$. 
In \ref{fig:noise_causal_diagram} and \ref{fig:noisy_anticausal_diagram}, 
the observed covariates are $\widetilde{x_1}, x_2$.
In all cases, $y$ denotes the label.}
\label{fig:causal_graphs}
\vspace{-10px}
\end{figure}

To begin, we consider linear Gaussian models 
\citep[Figure \ref{fig:causal_graphs};][]{wright1934method},
with the following goals:
to 
(i) gain qualitative insights into when 
a predictor might rely on spurious signals
in the first place;
and (ii) provide a mechanism of action 
to explain the efficacy of CAD. 
First, we analyze the causal setting 
(features cause the label).
When the features share a common cause
and a predictor is well-specified (linear),
it will assign zero weight (in expectation)
to non-causal features.
However, when the causal features
are subject to observation noise (measurement error), 
the non-causal features are assigned non-zero weight.
Conversely, when we inject noise on non-causal features,
predictors rely more on causal features,
which we expect to result in better out-of-domain generalization.
In the causal framework, 
we observe that CAD might be usefully 
formalized as a process analogous
to intervening on the causal features,
thus d-separating the label
from the non-causal features \citep{pearl1985bayesian}.
Alternatively, we might conceptualize CAD 
with an anticausal model \citep{scholkopf2012causal}.
In this setup, the label of interest is 
one of several latent attributes
that directly causes some (but not all features).
In this interpretation, 
we imagine that we have intervened on the label
and the editor’s role is to simulate 
the counterfactual document 
that would flow from the alternative label, 
holding other attributes constant.
Note that this too d-separates the label
from the spurious correlate.
In both cases, any model 
trained on the resulting data
ought to rely only on the causal features.

Our toy abstraction
points to a useful diagnostic test. 
If indeed CAD involves interventions 
on spans that are (in some sense) 
analogous to the causal features in our toy model, 
then injecting noise on these words should 
increase model reliance on the non-causal features
and thus (in general) lead to 
deteriorating performance out-of-domain.
On the other hand, injecting noise 
on the non-causal features 
should lead the model to rely more 
on the causal features, 
leading to improved performance out of domain. 
Through a series of large-scale empirical experiments
addressing sentiment analysis 
and natural language inference (NLI) tasks,
we inject noise on the spans marked 
as causal vs non-causal.
We compare the effects of injecting noise on the spans
revised by the CAD editors, 
the spans selected through feature feedback \citep{zaidan2007using}, 
and to spans selected automatically
using feature attribution heuristics 
such as attention- and gradient-based saliency methods.
If indeed the hypotheses that
(i) identifying causal features requires human intervention;
and (ii) models relying on causal features 
generalize better out of domain; hold, 
we might expect that 
(compared to automatic attribution methods)
noising human-provided 
rationales would deteriorate out-of-domain performance,
while noising non-rationales should prove beneficial.

We show that an SVM sentiment analysis model
trained on the original $1.7k$ IMDb reviews 
from \citet{kaushik2020learning}
obtains $87.8\%$ accuracy on the IMDb test set and $79.9\%$ on 
Yelp reviews 
but when all \emph{rationales} 
are replaced with noise, 
the classifier experiences $\approx 11\%$ 
drop on in-sample accuracy and an even bigger drop 
of $\approx 28.7\%$ on Yelp.
However, as \emph{non-rationales} are replaced with noise,
in-domain accuracy goes down by $\approx 10\%$
but out-of-domain accuracy increases by $1.5\%$. 
Similarly, in NLI, the accuracy of a BERT classifier 
fine-tuned on a subsample of e-SNLI \citep{deyoung2020eraser} 
goes down by $\approx 20\%$ 
when \emph{rationales} are replaced with noise, 
whereas the out-of-domain accuracy 
goes down by $21.3$--$31.5\%$ on various datasets. 
If \emph{non-rationales} are replaced with noise, 
in-sample accuracy goes down by $6.2\%$
but out of domain accuracy drops by only $2.3$--$5.5\%$. 
Similar patterns are observed across both tasks,
on all datasets and models. 
However, when using attention masks,
the resulting changes in model performance 
do not appear to follow these trends.
In another test to probe whether human feedback
is indeed necessary to produce datasets 
with the observed quantitative results of CAD,
we experiment with 
style transfer methods for converting 
\emph{Positive} reviews into \emph{Negative} and vice versa.
Compared to an SVM classifier trained 
on style-transfer-augmented data, 
training on CAD leads to a gain 
of $5$--$16.4\%$ in accuracy on Amazon 
and $3.7$--$17.8\%$ on Yelp. 
Similarly, a BERT classifier fine-tuned on CAD 
outperforms the same classifier fine-tuned 
on style-transfer-augmented data 
by $4.9$--$21.5\%$ on Amazon and $1.9$--$9.5\%$ on Yelp.

\section{Related Work}
\label{sec:related_work}
NLP papers on spurious associations
have addressed social biases
\citep{dixon2018measuring,zhao2018gender,kiritchenko2018examining, dinan2019queens, may2019measuring},
spurious signals owing to annotation heuristics 
\citep{gururangan2018annotation, poliak2018hypothesis},
and artifacts from automatic data generation 
\citep{chen2016thorough, kaushik2018much},
Researchers have also demonstrated vulnerabilities
to synthetic transformations, 
such as distractor phrases \citep{jia2017adversarial,wallace2019universal}, 
document paraphrases \citep{iyyer2018adversarial,pfeiffer2019deep},
and synthetic but meaning-preserving modifications 
\citep{ribeiro2018semantically, glockner_acl18, shendarling}.

Researchers have proposed incorporating human feedback
solicited through a variety of mechanisms
including highlighting \emph{rationales},
spans of text indicative of the label 
\citep{zaidan2007using, zaidan2008modeling, poulis2017learning}.
To combat gender stereotypes,
\citet{lu2018gender,zmigrod-etal-2019-counterfactual,maudslay2019s} 
describe data augmentation approaches 
that programmatically alter text.
More recently, \citet{kaushik2020learning} employed crowd workers
to edit text to make an opposite label applicable.
Through their experiments they show
that classifiers trained on CAD 
generalize well out of domain. 
\citet{teney2020learning} show 
the benefits of CAD in computer vision and NLP,
and \citet{srivastava2020robustness} employ crowdworkers 
to augment their training data 
to capture potential unmeasured variables.
A growing body of work has also looked at 
reducing reliance on spurious correlations
by exploiting the stability of relationships 
between the target variable and its (graph) neighbors.
\citet{peters2016causal} propose \emph{invariant causal prediction}
to obtain a causal predictor from multiple datasets.
\citet{ghassami2017learning} discuss a similar approach 
but do not assume that the exogenous noise 
of the target variable stays fixed among environments. 
They also demonstrate the benefits of their approach 
(compared to \citet{peters2016causal})
in identifying all direct ancestors of the target variable.
\citet{arjovsky2019invariant} propose 
\emph{invariant risk minimization},
with the goal of learning a data representation 
such that the optimal predictor 
is shared across environments.

\section{Analysis of a Toy Model}
\label{sec:analysis}
We briefly review the OLS estimator
for the model $Y = X\beta + \mathbf{\epsilon}$,
where $Y \in \R{R}^{n}$ is the target,  
$X \in \R{R}^{n \times p}$ the design matrix,
$\beta \in \R{R}^{p}$ 
the coefficient vector we want to estimate, 
and $\epsilon \sim \C{N}(0, \sigma^2_\epsilon \B{I}_n)$ 
an iid noise term.
The OLS estimate ${\beta}^{ols}$
is given by
$\displaystyle \Cov(X,X) {\beta}^{ols} = \displaystyle \Cov(X,Y)$.
Representing $\textrm{Var}[X_i]$ as $\sigma^2_{x_i}$
and $\Cov(X_i,X_j)$ as $\sigma_{x_i, x_j}$, 
if we observe only two covariates $(p=2)$, then:
\begin{align}
\begin{split}
\beta_1^{ols} &= \frac{\sigma^2_{x_{2}}\sigma_{x_{1},y} - \sigma_{x_{1},x_{2}}\sigma_{x_{2},y}}{\sigma^2_{x_{1}}\sigma^2_{x_{2}} - {\sigma^2_{x_{1},x_{2}}}}
\end{split}
\begin{split}
\beta_2^{ols} &= \frac{\sigma^2_{x_{1}}\sigma_{x_{2},y} - \sigma_{x_{1},x_{2}}\sigma_{x_{1},y}}{\sigma^2_{x_{1}}\sigma^2_{x_{2}} - {\sigma^2_{x_{1},x_{2}}}}
\end{split}
\label{eq:ols-est}
\end{align}

Our analysis adopts the structural causal model 
(SCM) framework \citep{pearl2009causality}, 
formalizing causal relationships 
via Directed Acyclic Graphs (DAGs). 
Each edge of the form $A \rightarrow B \in \C{E}$
in a DAG $\C{G}=(\C{V},\C{E})$ indicates 
that the variable $A$ is (potentially)
a direct cause of variable $B$. 
All measured variables $X \in \C{V}$ in the model 
are deterministic functions of their corresponding parents 
$\textrm{Pa}(X) \subseteq \C{V}$ 
and a set of jointly independent noise terms.
For simplicity, we work with linear Gaussian SCMs 
in the presence of a single confounder
where each variable is a linear function of its parents 
and the noise terms are assumed to be additive and Gaussian.
We look at both causal and anticausal learning settings. 
In the former, we assume that a document 
causes the applicability of the label
(as in annotation, where the document truly causes the label).
In the latter interpretation,
we assume that the label is one latent variable (among many)
that causes features of the document
(as when a reviewer's ``actual sentiment'' influences what they write).
For simplicity, we assume that the latent variables
are correlated due to confounding but that 
each latent causes a distinct set of observed features.
Without loss of generality,
we assume that all variables have zero mean.
Both DAGs contain the four random variables 
$z, x_1, x_2, y$ 
and the anticausal DAG also contains
some additional latent variables $q$
(Figure \ref{fig:causal_graphs}). 
The derivations are standard 
and are included in Appendix \ref{sec:ols-step-by-step}.

\subsection{The Causal Setting}
We now focus on the causal setting
(Figure \ref{fig:causal_diagram}, \ref{fig:noise_causal_diagram})
Let the Gaussian SCM be defined as follows 
where the noise term for variable $x$ is defined as $u_{x}$:
\begin{align} \label{eq:causalSCM}
\begin{split}
    z &= u_z,  \\
    x_{1} &= b z + u_{x_1}, \\
    x_{2} &= c z + u_{x_2}, \\
    y &= a x_{1} + u_y,
\end{split} 
\begin{split}
    u_z &\sim \mathcal{N}(0, \sigma^2_{u_z}) \\
    u_{x_1} &\sim \mathcal{N}(0, \sigma^2_{u_{x1}}) \\
    u_{x_2} &\sim \mathcal{N}(0, \sigma^2_{u_{x2}}) \\
    u_y &\sim \mathcal{N}(0, \sigma^2_{u_y}).
\end{split} 
\end{align}
Applying OLS, we obtain $\beta_1^{ols} = a$ and $\beta_2^{ols} = 0$.
However, consider what happens if we only observe $x_1$ via a noisy proxy 
$\widetilde{x_1} \sim \mathcal{N}(x_1, \sigma^2_{u_{x_1}}+\sigma^2_{\epsilon_{x_1}})$
(Figure \ref{fig:noise_causal_diagram}).
Assuming, $\epsilon_{x_1} \ind  (x_1, x_2, y)$,
from Eq. \ref{eq:ols-est} we get the estimates 
$\widehat{\beta_1^{ols}}$ and $\widehat{\beta_2^{ols}}$
(Eq. \ref{eq:ols-est-causal-noise})
in the presence of observation noise on $x_1$.
\begin{align}
\label{eq:ols-est-causal-noise}
\begin{split}
\widehat{\beta_1^{ols}} &= \frac{
                            a(\sigma^2_{u_{z}}(b^2\sigma^2_{u_{x2}}
                            +c^2\sigma^2_{u_{x1}})
                            +\sigma^2_{u_{x1}}\sigma^2_{u_{x2}})
                        }
                        {
     \sigma^2_{u_{z}}(b^2\sigma^2_{u_{x2}}+c^2\sigma^2_{u_{x1}})
                +\sigma^2_{u_{x1}}\sigma^2_{u_{x2}}
         +\sigma^2_{\epsilon_{x1}}(c^2\sigma^2_{u_z}+\sigma^2_{u_{x2}})
                        }\\
\widehat{\beta_2^{ols}} &= \frac{
                        acb\sigma^2_{\epsilon_{x1}}\sigma^2_{u_z}
                        }
                        {
                        \sigma^2_{u_{z}}(b^2\sigma^2_{u_{x2}}+c^2\sigma^2_{u_{x1}})+\sigma^2_{u_{x1}}\sigma^2_{u_{x2}}
                        +\sigma^2_{\epsilon_{x1}}(c^2\sigma^2_{u_z}+\sigma^2_{u_{x2}})
                        }\\
\end{split}
\end{align}
As we can see, $\widehat{\beta_1^{ols}} \propto \frac{1}{\sigma^2_{\epsilon_{x1}}}$.
This shows us that as $\sigma^2_{\epsilon_{x1}}$ increases, 
$|\widehat{\beta_1^{ols}}|$ (the magnitude of the coefficient for $x_1$) decreases 
and $|\widehat{\beta_2^{ols}}|$ (the magnitude of the coefficient for $x_2$) increases. 
The asymptotic OLS estimates in the presence 
of infinite observational noise 
is 
$\lim_{\sigma^2_{\epsilon_{x1}} \rightarrow \infty} 
\widehat{\beta_1^{ols}} = 0$,
whereas $\widehat{\beta_2^{ols}}$ 
converges to a finite non-zero value.  
On the other hand, observing a noisy version of $x_{2}$
will not affect our OLS estimates 
if there is no measurement error on $x_1$.

These simple graphs provide qualitative insights 
into when we should expect a model 
to rely on spurious patterns.
In the causal setting, 
under perfect measurement, 
the causal variable 
d-separates the non-causal variable from the label 
(Figure \ref{fig:causal_diagram}). 
However, under observation noise,
a predictor will rely on the non-causal variable 
(Eq. \ref{eq:ols-est-causal-noise}).
Moreover, when the causal feature is noisily observed,
additional observation noise on non-causal features
yields models that are more reliant on causal features.
We argue that while review text is not noisily observed per se,
learning with imperfect feature representations 
acquired by training deep networks on finite samples
has an effect that is analogous 
to learning with observation noise. 

\paragraph{Connection to Counterfactually Augmented Data}
In the causal setting,
intervening on the causal feature,
d-separates the label $y$ 
from the non-causal feature $x_2$,
and thus models trained 
on samples from the interventional distribution
will rely solely on the causal feature, 
even when it is noisily observed.
We argue that in a qualitative sense,
the process of generating CAD
resembles such an intervention,
however instead of intervening randomly,
we
ensure that for each example,
we produce two sets of values of $x_1$,
one such that the label is applicable
and one such that it is not applicable.
One is given in the dataset,
and the other is produced via the revision.

\subsection{An Anticausal Interpretation}
Alternatively, rather than 
thinking of features causing the applicable label,
we might think of the ``causal feature" 
as a direct effect of the label (not a cause).
In this case, so long as the relationship
is truly not deterministic, 
even absent noisy observation,
conditioning on the causal feature
does not d-separate the label 
from the non-causal feature 
and thus models should be expected
to assign weight to both 
causal and non-causal variables. 

As in the causal setting,
as we increase observation noise on the causal variable,
the weight assigned to the non-causal variable should increase. 
Conversely, as in the causal setting 
with observation noise on $x_1$,
as observation noise on the non-causal feature $x_2$ increases,
we expect the learned predictor 
to rely more on the causal feature. 
We derive the OLS coefficients 
(including under the presence of observational noise,
Fig. \ref{fig:noisy_anticausal_diagram}) 
in this setting in Appendix \ref{subsec:anticausal}.

\paragraph{Connection to Counterfactually Augmented Data}
In this interpretation,
we think of CAD as a process by which
we (the designers of the experiment)
intervene on the label itself
and the human editors, 
play the role of a simulator 
that we imagine to be capable 
of generating a counterfactual example,
holding all other latent variables constant.
In the sentiment case,
we could think of the editors  
as providing us with the review 
that would have existed
had the sentiment been flipped,
holding all other aspects of the review constant. 
Note that by intervening on the label,
we d-separate it from the spurious correlate $x_2$
(Figure \ref{fig:anticausal_diagram}).

\subsection{Insights and Testable Hypotheses}

In both the causal and anticausal models,
the mechanism underlying the causal relationship 
that binds $x_1$ to $y$ 
(regardless of direction)
is that binding language 
to a semantic concept (such as sentiment),
which we expect to be more stable across settings
than the more capricious relationships
among the background variables,
e.g., those linking genre and production quality.

In that spirit, if spans edited to generate 
\emph{counterfactually revised data} (CRD) 
are analogous to the causal
(or anticausal) variables, 
in the causal (or anticausal) graphs, 
then we might expect that 
noising those spans (e.g. by random word replacement)
should lead to models that rely more on 
non-causal features and perform worse 
on out of domain data. 
On the other hand, we expect that 
noising unedited spans 
should have the opposite behavior,
leading to degraded in-domain performance,
but comparatively better out-of-domain performance. 
In the remainder of the paper, 
we investigate these hypotheses, 
finding evidence that qualitatively 
confirms the predictions of our theory.

We freely acknowledge the speculative nature 
of this analysis and concede 
that the mapping between the messy 
unstructured data we wish to model
and the neatly disentangled portrait
captured by our linear Gaussian models
leaves a gap to be closed through further
iterations of theoretical refinement and scientific experiment. 
Ultimately, our argument is not that 
this simple analysis fully accounts 
for counterfactually augmented data
but instead that it is a useful abstraction
for formalizing two (very different)
perspectives on how to conceive of CAD,
and for suggesting interesting hypotheses 
amenable to empirical verification.

\section{Empirical Results}
\label{sec:results}
If spans marked as rationales by humans 
via editing or highlighting 
are analogous to causal features, 
then noising those spans should lead to models 
that rely more on non-causal features 
and thus perform worse on out-of-domain data,
and noising the unmarked spans 
(analagous to non-causal features)
should have the opposite behavior. 
In this section, we test these hypotheses 
empirically on real-world datasets. 
Additionally, we investigate 
whether the feedback from human workers
is yielding anything qualitatively different 
from what might be seen with spans marked 
by automated feature attribution methods 
such as attention and saliency.
Along similar, lines we ask whether CAD in the first place 
offers qualitative advantages over
what might be achieved via 
automatic sentiment-flipping methods
through experiments with text style transfer algorithms.

We conduct experiments on sentiment analysis
\citep{zaidan2007using, kaushik2020learning} 
and NLI \citep{deyoung2020eraser}.
All datasets are accompanied with human feedback 
(tokens deemed relevant to the label's applicability)
which we refer to as \emph{rationales}. 
For the first set of experiments,
we rely on four models: 
Support Vector Machines (SVMs), 
Bidirectional Long Short-Term Memory Networks (BiLSTMs) 
with Self-Attention \citep{graves2005framewise}, 
BERT \citep{devlin2019bert},
and Longformer \citep{Beltagy2020Longformer}.
For the second set of experiments, 
we rely on four state-of-the-art style transfer models 
representative of different methodologies, 
each representative of a different approach
to automatically generate 
new examples with flipped labels 
\citep{hu2017toward, li2018delete, sudhakar2019transforming, madaan2020politeness}.
To evaluate classifier performance 
on the resulting augmented data, 
we consider SVMs, Naive Bayes (NB),
BiLSTMs with Self Attention, and BERT. 
We relegate implementation details 
to Appendix~\ref{sec:model_details}.

For sentiment analysis, we use 
SVM, BiLSTM with Self Attention, 
BERT, and Longformer models.
In each document, we replace a fraction
of \emph{rationale} (or \emph{non-rationale}) tokens 
with random tokens sampled from the vocabulary, 
and train our models, 
repeating the process $5$ times. 
We perform similar experiments for NLI using BERT. 
As an individual premise-hypothesis pair 
is often not as long as a movie review, 
many pairs only have one or two words 
marked as \emph{rationales}.
To observe the effects from gradually injecting 
noise on \emph{rationales} or \emph{non-rationales},
we select only those premise-hypothesis pairs 
that have a minimum $10$ tokens 
marked as \emph{rationales}. 
Since no neutral pairs exist 
with $10$ or more rationale tokens, 
we consider only a binary classification setting
(entailment-contradiction),
and downsample the majority class 
to ensure a $50$:$50$ label split.

\begin{figure*}[t!]
    \begin{subfigure}[h]{\textwidth}
         \centering
         \includegraphics[width=\linewidth]{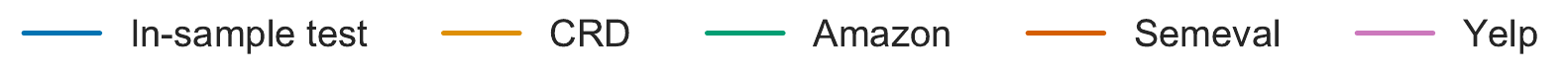}
    \end{subfigure}\\
    \begin{subfigure}[h]{\textwidth}
         \centering
         \includegraphics[width=0.48\linewidth]{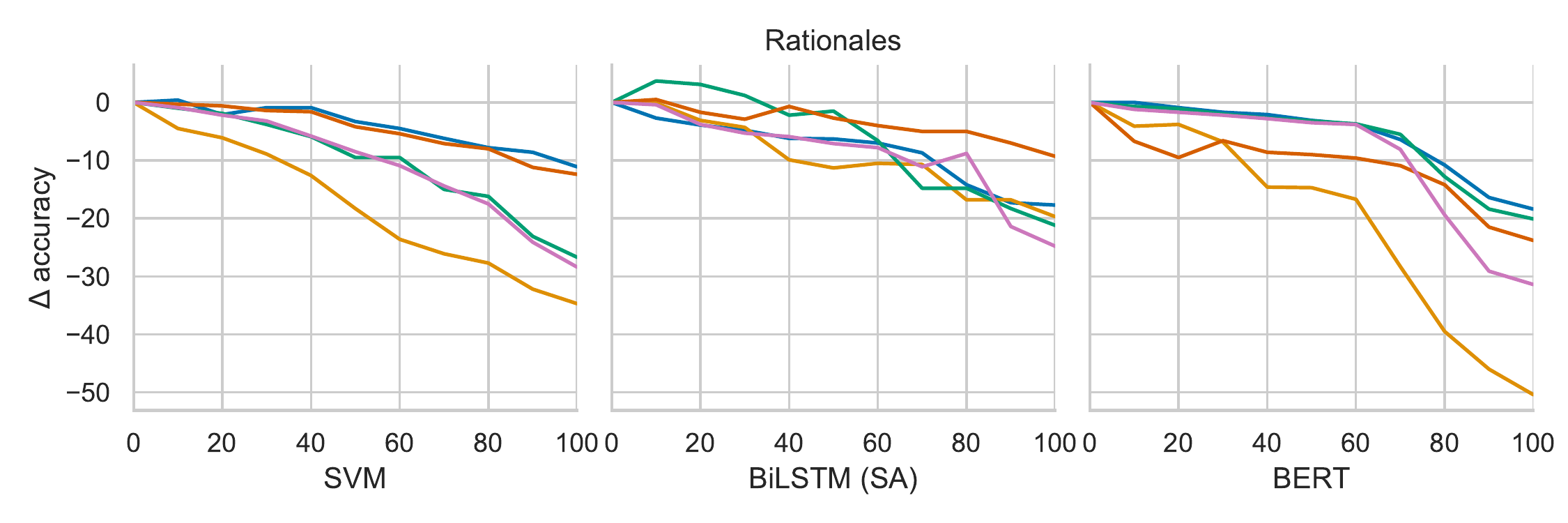}
        \hfill
          \includegraphics[width=0.48\linewidth]{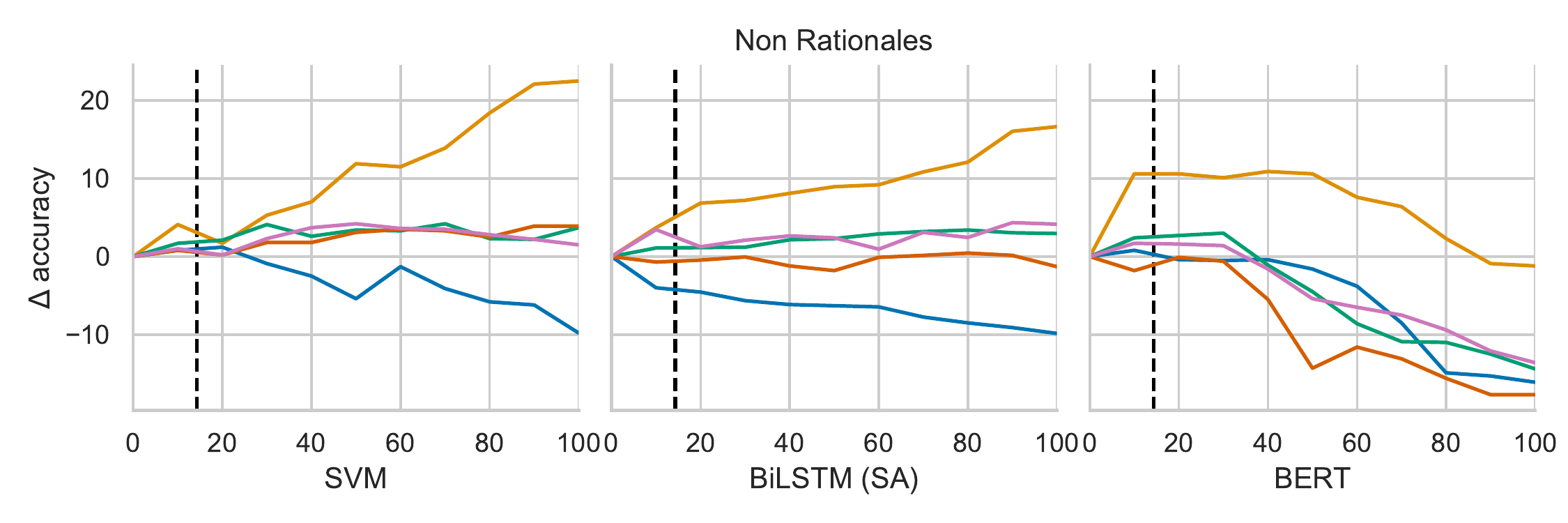}
          \caption{Noising spans marked by humans}
          \label{fig:iclr_human}
      \end{subfigure}\\
    \begin{subfigure}[h]{\textwidth}
         \centering
         \includegraphics[width=0.48\linewidth]{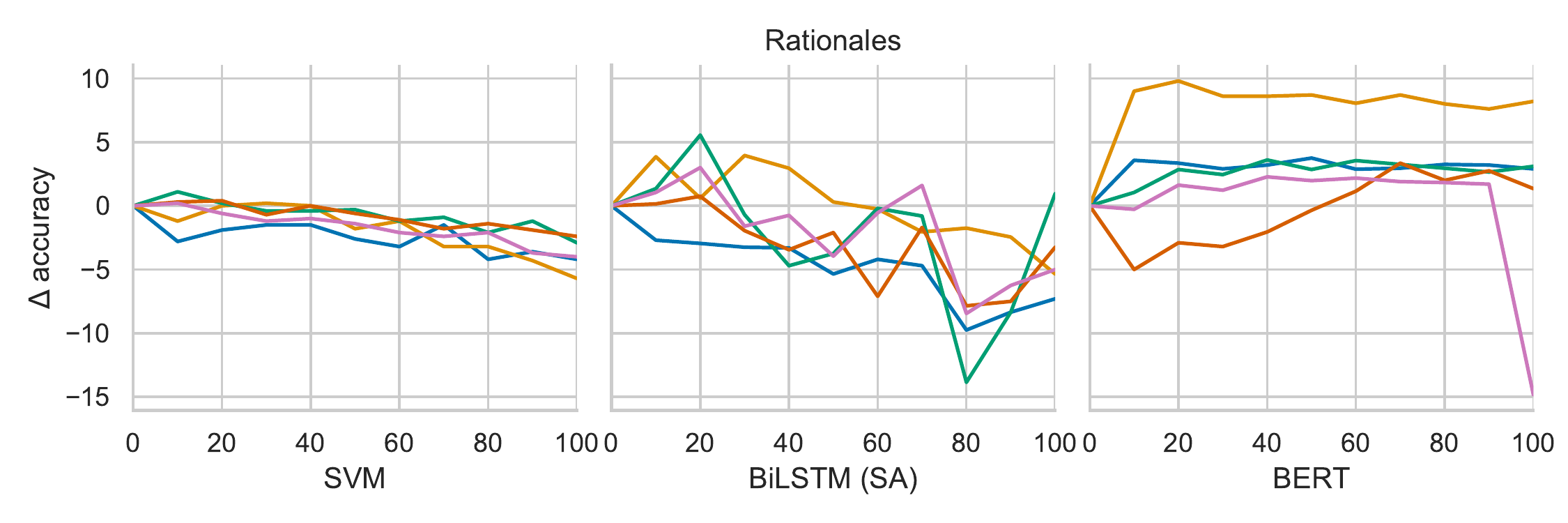}
        \hfill
          \includegraphics[width=0.48\linewidth]{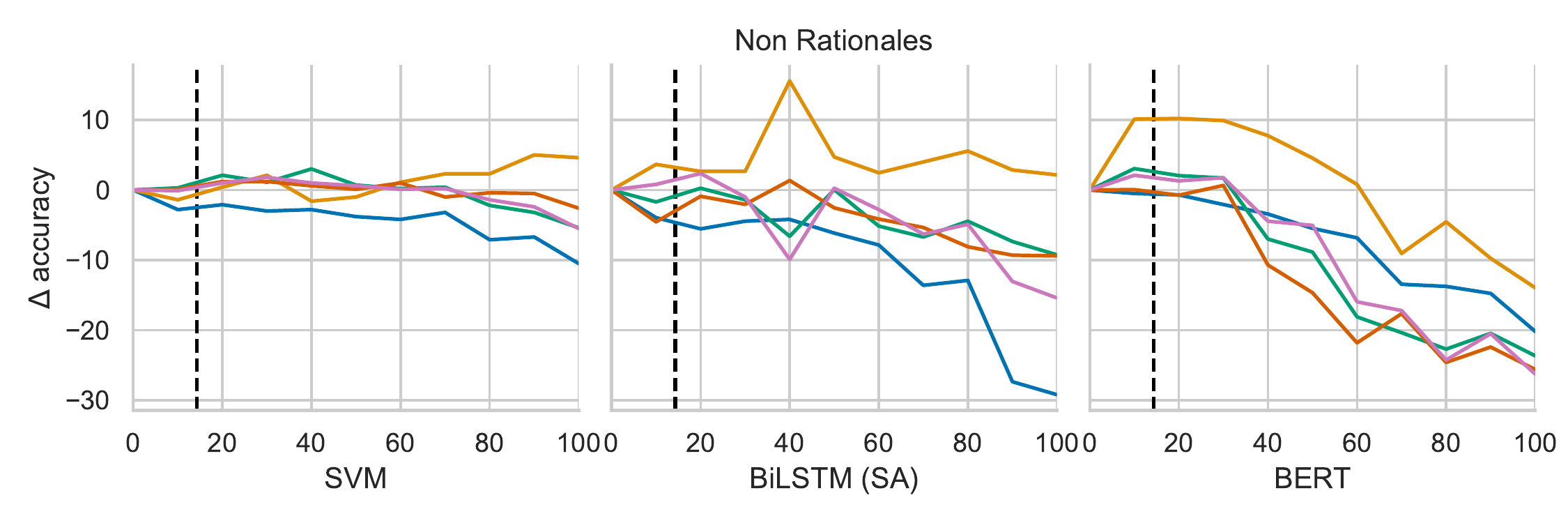}
          \caption{Noising spans marked by Attention}
          \label{fig:iclr_attention}
      \end{subfigure}\\
    \begin{subfigure}[h]{\textwidth}
         \centering
         \includegraphics[width=0.48\linewidth]{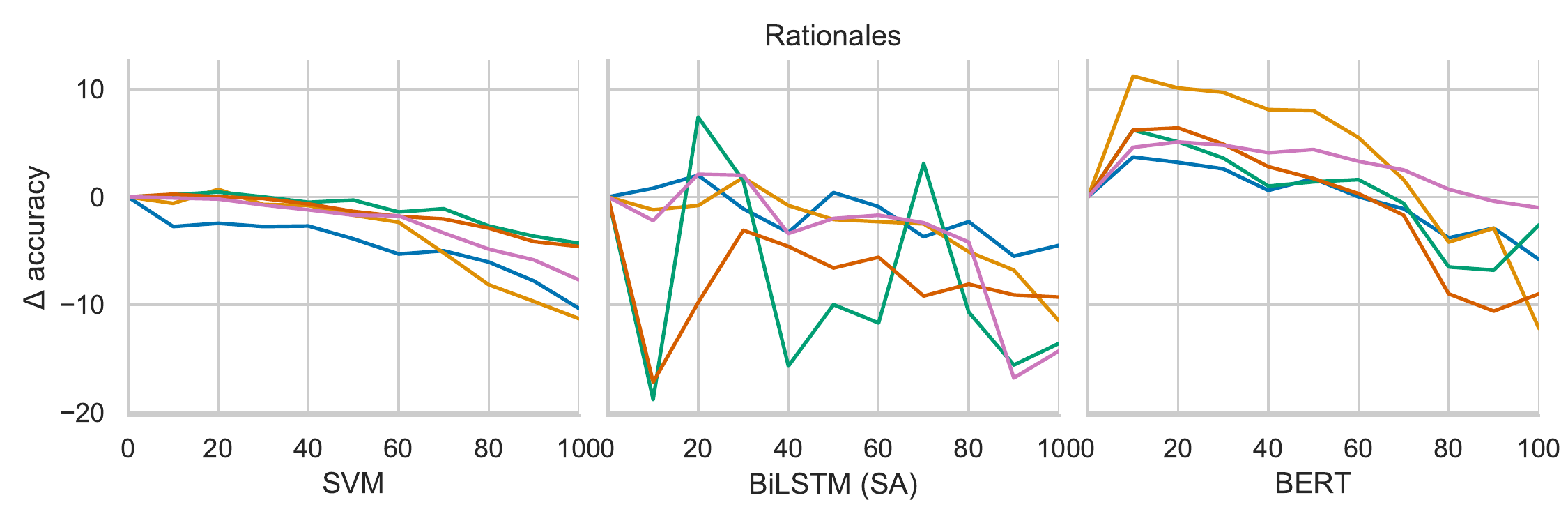}
        \hfill
          \includegraphics[width=0.48\linewidth]{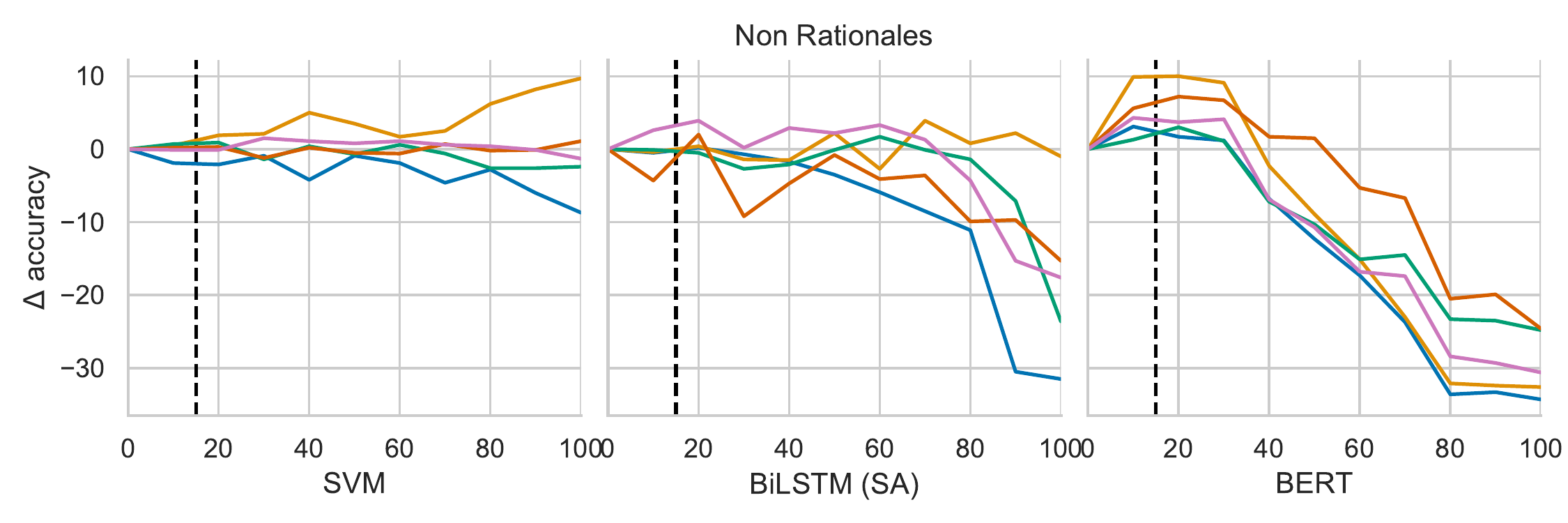}
          \caption{Noising spans marked via gradient based feature attribution}
          \label{fig:iclr_saliency}
      \end{subfigure}
\caption{
Change in classifier accuracy as noise is injected on \emph{rationales/non-rationales} for IMDb reviews from \citet{kaushik2020learning}.
\label{fig:iclr_noise}}
\end{figure*}
\begin{figure*}[t!]
    \begin{subfigure}[t!]{\textwidth}
         \centering
         \includegraphics[width=0.48\linewidth]{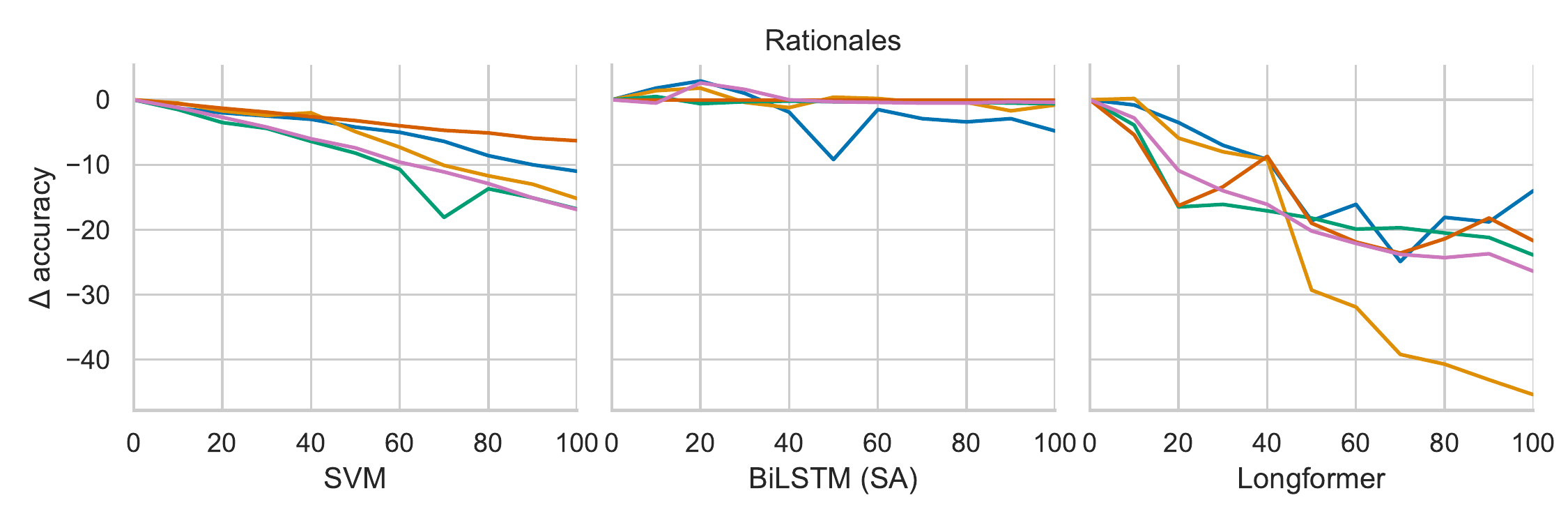}
         \hfill
          \includegraphics[width=0.48\linewidth]{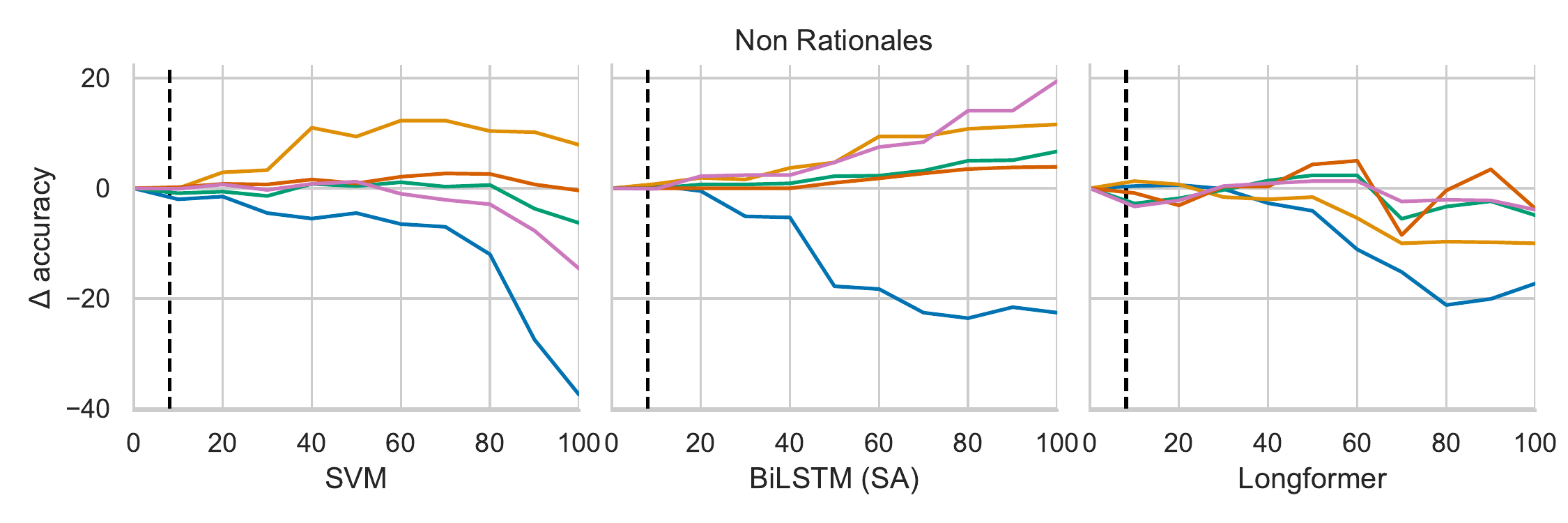}
          \caption{Noising spans marked by humans}
          \label{fig:zaidan_human}
      \end{subfigure}\\
    \begin{subfigure}[t!]{\textwidth}
         \centering
         \includegraphics[width=0.48\linewidth]{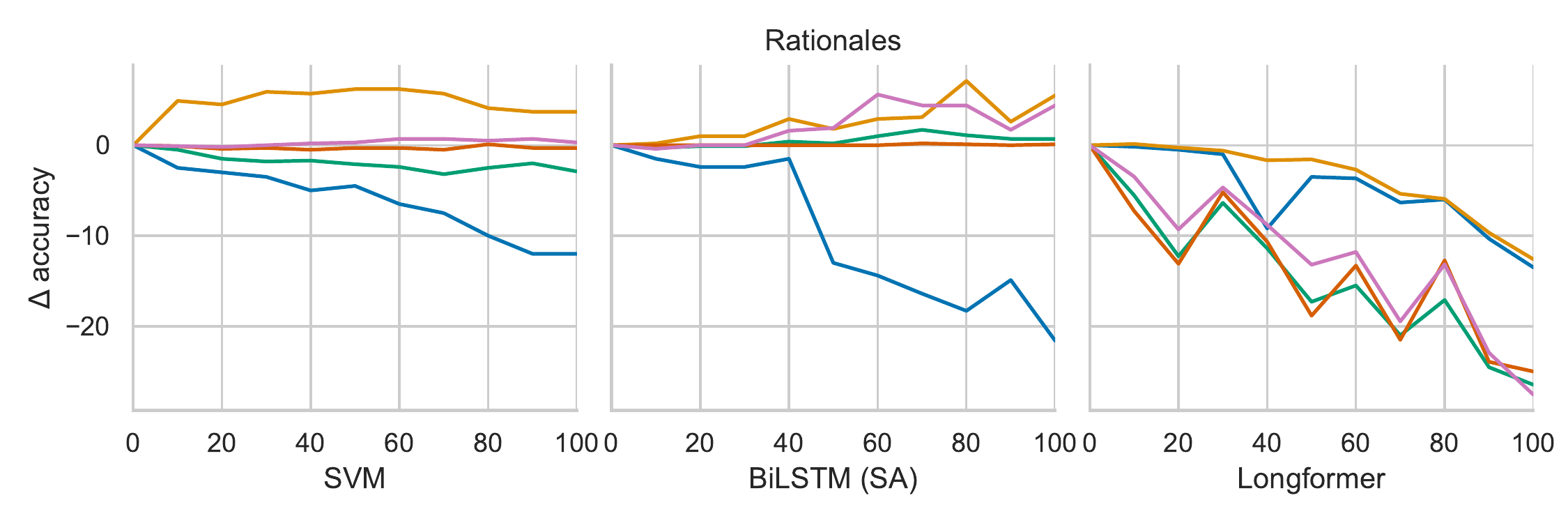}
         \hfill
          \includegraphics[width=0.48\linewidth]{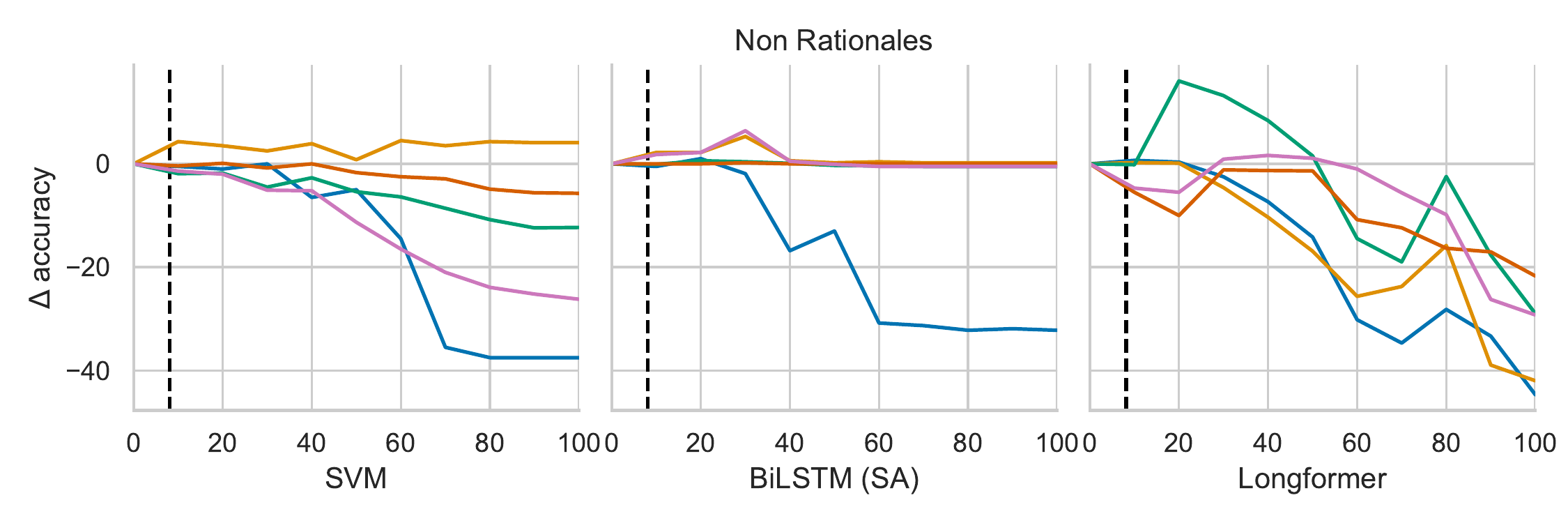}
          \caption{Noising spans marked by Attention}
          \label{fig:zaidan_attention}
      \end{subfigure}\\
    \begin{subfigure}[t!]{\textwidth}
         \centering
         \includegraphics[width=0.48\linewidth]{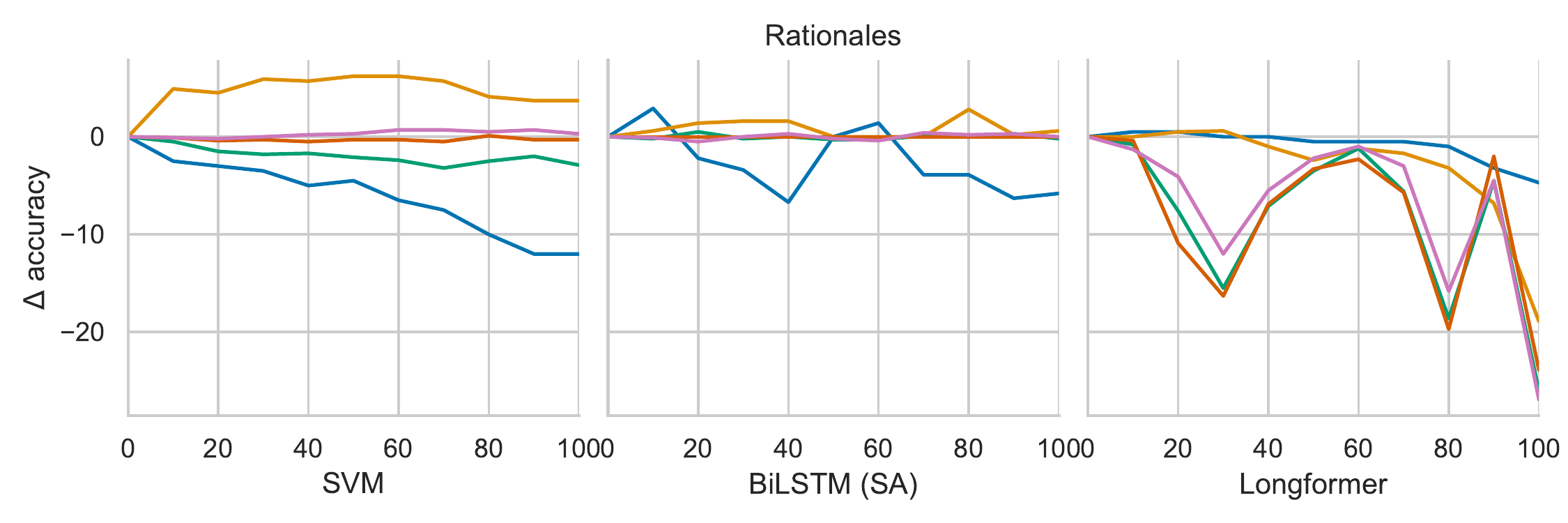}
         \hfill
          \includegraphics[width=0.48\linewidth]{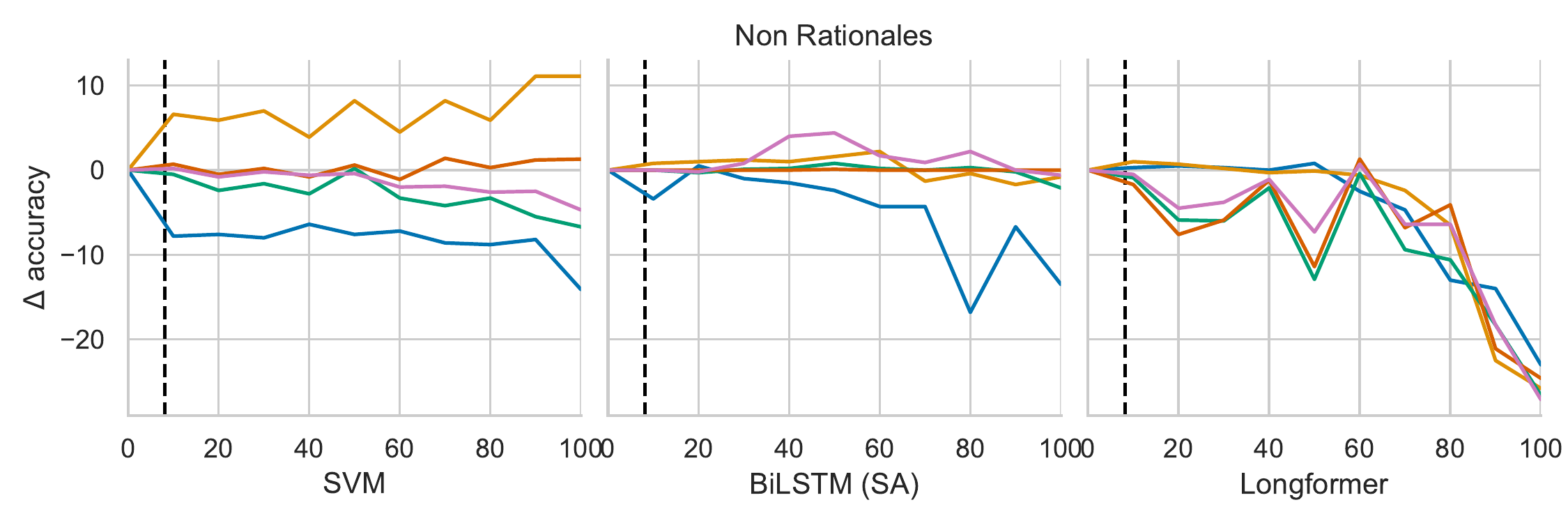}
          \caption{Noising spans marked via gradient based feature attribution}
          \label{fig:zaidan_saliency}
      \end{subfigure}
\caption{
Change in classifier accuracy as noise is injected on \emph{rationales/non-rationales} for IMDb reviews from \citet{zaidan2007using}. In both Figures \ref{fig:iclr_noise} and \ref{fig:zaidan_noise}, the vertical dashed line indicates the fraction of median length of \emph{non-rationales} equal to the median length of \emph{rationales}. 
\label{fig:zaidan_noise}}
\end{figure*}

Figures \ref{fig:iclr_noise} and \ref{fig:zaidan_noise} 
show the difference in mean accuracy over $5$ runs. 
For all classifiers, as the noise in \emph{rationales} increases, 
in-sample accuracy stays relatively stable 
compared to out-of-domain accuracy. 
An SVM classifier trained 
on the original $1.7k$ IMDb reviews 
from \citet{kaushik2020learning}
obtains $87.8\%$ accuracy on the IMDb test set 
and $79.9\%$ on Yelp reviews.\footnote{The 
out-of-domain evaluation sets in \citet{kaushik2020learning} 
do not have $50$:$50$ label split. 
We enforce this split to observe
when a classifier approaches random baseline performance. All datasets can be found at \href{https://github.com/acmi-lab/counterfactually-augmented-data}{https://github.com/acmi-lab/counterfactually-augmented-data}} 
As a greater fraction of \emph{rationales} are replaced 
with random words from the vocabulary, 
the classifier experiences a drop of $\approx 11\%$ 
by the time all \emph{rationale} tokens are replaced with noise. 
However, it experiences an $28.7\%$ 
drop in accuracy on Yelp reviews.
Similarly, on the same datasets, 
a fine-tuned BERT classifier 
sees its in-sample accuracy drop by $18.4\%$, 
and by $31.4\%$ on Yelp as \emph{rationale} tokens 
replaced by noise go from $0$ to $100\%$.
However, as more \emph{non-rationales} are replaced with noise, 
in-sample accuracy for SVM goes down by $\approx 10\%$ 
but increases by $1.5\%$ on Yelp. 
For BERT, in-sample accuracy decreases 
by only $16.1\%$ and only $13.6\%$ on Yelp
(Also see 
Appendix Table \ref{tab:iclr_human_noise}, 
and Appendix Figure \ref{fig:iclr_human_appendix}).

\begin{table*}[!t]
  \begin{center}
  \caption{Accuracy of BERT trained on SNLI \citep{deyoung2020eraser} as noise is injected on human identified \emph{rationales/non-rationales}. RP and RH are Revised Premise and Revised Hypothesis test sets in \citet{kaushik2020learning}. MNLI-M and MNLI-MM are MNLI \citep{williams2018broad} dev sets.
  \label{tab:nli_results}}
  \begin{tabular}{ l c c c c c c c c c c c}
    \toprule
    & \multicolumn{11}{c}{Percent noise added to train data rationales} \\
    Dataset & $0$ & $10$ & $20$ & $30$ & $40$ & $50$ & $60$ & $70$ & $80$ & $90$ & $100$\\
    \midrule
    In-sample test & $91.6$ & $90.7$ & $90.0$ & $88.9$ & $87.3$ & $86.2$ & $84.4$ & $80.2$ & $78.0$ & $72.2$ & $71.9$\\
    RP & $72.7$ & $70.7$ & $69.1$ & $67.1$ & $65.7$ & $62.4$ & $61.8$ & $57.7$ & $55.6$ & $53.8$ & $51.4$\\
    RH & $84.7$ & $80.8$ & $80.4$ & $79.5$ & $77.2$ & $75.7$ & $73.3$ & $67.7$ & $64.0$ & $57.9$ & $53.2$\\
    MNLI-M & $75.6$ & $74.7$ & $73.9$ & $72.0$ & $70.6$ & $69.1$ & $64.7$ & $59.1$ & $55.8$ & $54.4$ & $53.3$\\
    MNLI-MM & $77.9$ & $76.7$ & $75.6$ & $73.9$ & $72.3$ & $70.8$ & $65.6$ & $58.4$ & $55.1$ & $53.6$ & $52.5$\\
    \midrule
    & \multicolumn{11}{c}{Percent noise added to train data non-rationales} \\
    Dataset  & $0$ & $10$ & $20$ & $30$ & $40$ & $50$ & $60$ & $70$ & $80$ & $90$ & $100$\\
    \midrule
    In-sample test & $91.6$ & $91.4$ & $91.3$ & $90.9$ & $90.8$ & $89.9$ & $89.0$ & $88.7$ & $87.8$ & $86.7$ & $85.4$\\
    RP & $72.7$ & $73.5$ & $73.2$ & $72.1$ & $71.5$ & $70.7$ & $70.6$ & $70.6$ & $70.6$ & $70.6$ & $70.4$\\
    RH & $84.7$ & $83.6$ & $82.6$ & $81.9$ & $81.3$ & $81.1$ & $80.5$ & $79.8$ & $79.4$ & $79.4$ & $79.2$\\
    MNLI-M & $75.6$ & $74.9$ & $74.4$ & $72.6$ & $72.4$ & $71.8$ & $71.3$ & $71.3$ & $70.9$ & $70.9$ & $70.8$\\
    MNLI-MM & $77.9$ & $76.2$ & $75.8$ & $75.0$ & $74.6$ & $74.3$ & $73.9$ & $73.7$ & $73.3$ & $73.0$ & $72.8$\\
    \bottomrule
  \end{tabular}
  \end{center}
\end{table*}

\begin{table*}[!th]
 \begin{center}
  \caption{Out-of-domain accuracy of models trained on original only, CAD, and original and \emph{sentiment-flipped} reviews}\label{tab:sentiment_out}
  \begin{tabular}{ l c c c c}
    \toprule
    Training data & SVM & NB & BiLSTM (SA) & BERT \\
    \midrule
    \multicolumn{5}{c}{Accuracy on Amazon Reviews}\\
    \midrule
    CAD ($3.4k$) & $\textbf{79.3}$ & $\textbf{78.6}$ & $\textbf{71.4}$ & $\textbf{83.3}$ \\
    Orig. \& \citet{hu2017toward} & $66.4$ & $71.8$& $62.6$ & $78.4$\\
    Orig. \& \citet{li2018delete} & $62.9$ & $65.4$& $57.6$ & $61.8$\\
    Orig. \& \citet{sudhakar2019transforming} & $64.0$ & $69.3$& $54.7$ & $77.2$\\
    Orig. \& \citet{madaan2020politeness} & $74.3$ & $73.0$& $63.8$ & $71.3$\\
    Orig. ($3.4k$) & $74.5$ & $74.3$ & $68.9$ & $80.0$ \\
    \midrule
    \multicolumn{5}{c}{Accuracy on Semeval 2017 (Twitter)}\\
    \midrule
    CAD ($3.4k$) & $\textbf{66.8}$ & $\textbf{72.4}$ & $\textbf{58.2}$ & $\textbf{82.8}$ \\
    Orig. \& \citet{hu2017toward} & $60.9$ & $63.4$& $56.6$ & $79.2$\\
    Orig. \& \citet{li2018delete} & $57.6$ & $60.8$& $54.7$ & $62.7$\\
    Orig. \& \citet{sudhakar2019transforming} & $59.4$ & $62.6$& $54.9$ & $72.5$\\
    Orig. \& \citet{madaan2020politeness} & $62.8$ & $63.6$& $54.6$ & $79.3$\\
    Orig. ($3.4k$) & $63.1$ & $63.7$ & $50.7$ & $72.6$ \\
    \midrule
    \multicolumn{5}{c}{Accuracy on Yelp Reviews}\\
    \midrule
    CAD ($3.4k$) & $\textbf{85.6}$ & $\textbf{86.3}$ & $\textbf{73.7}$ & $\textbf{86.6}$ \\
    Orig. \& \citet{hu2017toward} & $77.4$ & $80.4$& $68.8$ & $84.7$\\
    Orig. \& \citet{li2018delete} & $67.8$ & $73.6$& $63.1$ & $77.1$\\
    Orig. \& \citet{sudhakar2019transforming} & $69.4$ & $75.1$& $66.2$ & $84.5$\\
    Orig. \& \citet{madaan2020politeness} & $81.3$ & $82.1$& $68.6$ & $78.8$\\
    Orig. ($3.4k$) & $81.9$ & $82.3$ & $72.0$ & $84.3$ \\
\bottomrule
  \end{tabular}
  \end{center}
\end{table*}

We obtain similar results using \emph{rationales} 
identified via feature feedback.
An SVM classifier trained on reviews from \citet{zaidan2007using} 
sees in-sample accuracy drop by $11\%$, 
and accuracy on Yelp drop by $16.9\%$ 
as noise is inserted on \emph{rationales} 
but goes down by $17.3\%$ and $14.6\%$, respectively 
when noise is inserted in \emph{non-rationales}. 
For Longformer, in-sample accuracy drops by $14\%$ 
and accuracy on Yelp goes down by $26.4\%$ 
compared to a drop of $17.3\%$ and gain of $3.9\%$, 
respectively, when noise is inserted in \emph{non-rationales}.
Similar patterns are observed across datasets and models 
(see Figure \ref{fig:zaidan_human}, Appendix Table \ref{tab:zaidan_human_noise}, and Appendix Figure \ref{fig:zaidan_human_appendix}).\footnote{While 
similar trends are observed for both feature feedback and CAD, 
it is less clear how to incorporate feature feedback 
for training effectively with deep neural networks 
and pre-trained transformer architectures, 
whereas training (or fine-tuning) models on CAD is straightforward.}

For NLI, the in-sample accuracy of BERT 
fine-tuned on an SNLI subsample drops by $\approx 20\%$ 
when \emph{rationales} are replaced with noise, 
and out-of-domain accuracy goes down by $21.3$--$31.5\%$ 
on various datasets (Table \ref{tab:nli_results}). 
Whereas, if \emph{non-rationales} are replaced with noise, 
in-sample accuracy goes down by $6.2\%$
but out-of-domain accuracy drops by only $2.3$--$5.5\%$.
These results support our hypothesis 
that spans marked by humans as causing a label 
are analogous to causal variables.

Interestingly, in our NLI experiments, 
for various models the drops in both in-sample 
and out-of-domain accuracy are greater in magnitude 
when noise is injected in \emph{rationales} 
versus when it is injected in \emph{non-rationales}. 
This is opposite to what we observe in sentiment analysis.
We conjecture that these results are due to the fact that
in our experiment design for NLI, 
we only keep those premise-hypothesis pairs 
that contain at least 10 tokens marked as \emph{rationales} 
so we can observe the difference in accuracy 
as the amount of noise increases. 
A consequence of this selection is that many pairs selected 
have many more tokens marked as 
\emph{rationales} than \emph{non-rationales}, 
whereas, in sentiment analysis this is the opposite.
Hence, in NLI when some percentage of \emph{rationales} are replaced 
by noise, this corresponds to many more edited tokens 
than when a corresponding percentage of \emph{non-rationales} are noised.

To compare human feedback
to automatic feature attribution methods 
such as attention \citep{bahdanau2015neural} 
and gradient based saliency methods \citep{li2016visualizing}, 
we conduct the same set of experiments 
assuming tokens attended to (or not) 
by an attention based classifier (BiLSTM with Self-Attention) 
or identified as highly influential 
by a gradient based feature attribution method (salience scores)
as new \emph{rationales} (or \emph{non-rationales}).
In this case, unlike our findings with human feedback, 
we observe markedly different behavior 
than predicted by our analysis of the toy causal model
(See Figures \ref{fig:iclr_attention}, \ref{fig:iclr_saliency},
\ref{fig:zaidan_attention}, and \ref{fig:zaidan_saliency};
and Appendix Tables \ref{tab:iclr_attention_noise}, 
\ref{tab:iclr_saliency_noise}, 
\ref{tab:zaidan_attention_noise}, 
and \ref{tab:zaidan_saliency_noise}).

While we might not expect spurious signals
to be as reliable out of domain,
that does not mean that they will always fail.
For example, while the associations between genre and sentiment
learned from a dataset of book reviews
might not hold in a dataset of kitchen appliances,
but nevertheless hold in a dataset of audiobook reviews. 
In such settings, even though noising non-causal features
would lead to models relying more on causal features, 
this may not result in better out-of-domain performance.

We also look at whether we really need 
to go through the process of collecting CAD 
(or human-annotated rationales) at all 
or if automated methods for generating ``counterfactuals'' 
might obtain similar gains in out-of-domain performance, 
as the former could be an expensive process.
We experiment with state-of-the-art style transfer methods 
to convert \emph{Positive} reviews 
into \emph{Negative} and vice versa. 
Ideally, we would expect these methods 
to preserve a document's ``content'' 
while modifying the attributes 
that relate to sentiment 
(if they obtain perfect disentanglement in the feature space).
Sentiment classifiers trained on original and
\emph{sentiment-flipped} reviews generated 
using style transfer methods 
often give better out-of-domain performance 
compared to training only on original data 
of same size (Table \ref{tab:sentiment_out}). 
However, models trained on 
CAD perform even better across all datasets, 
hinting at the value of human feedback.

\section{Conclusion}
\label{sec:conclusion}
While prior work offers promising clues 
to the benefits of CAD generated 
through human-in-the-loop mechanisms,
previous work lacked formal frameworks 
for thinking about the technique,
or comparisons to plausible alternatives. 
In this paper, through simple analysis 
on toy linear Gaussian models 
followed by a large-scale empirical investigation 
on sentiment analysis and NLI tasks, 
we formalize CAD and take some initial steps
towards understanding its practical efficacy.
Our analysis suggests that data corrupted 
by adding noise to rationale spans
(analogous to adding noise to causal features)
will degrade out-of-domain performance,
while noise added to non-causal features 
may make models more robust out-of-domain.
Our empirical study focuses on sentiment analysis and NLI 
and our findings remain consistent across datasets and models. 
Furthermore, the two tasks are subjectively very different 
as sentiment analysis requires a strong consideration 
of expressions of opinion than stated facts,
whereas NLI is 
the 
opposite.
We also 
show
that models trained on the augmentation 
of original data and revised data 
generated by style transfer methods 
had better out-of-domain generalization 
in some cases compared to models 
trained on original data alone,
but performed worse than models trained on CAD. 
In future work, we will look at how 
these findings generalize to other domains, 
including computer vision, and investigate 
the surprisingly low susceptibility 
of pre-trained transformers to spurious associations.

\section*{Acknowledgements}
The authors are grateful to NVIDIA for providing GPUs to conduct the experiments, Salesforce Research and Facebook AI for their financial support, and Sanket Mehta, Sina Fazelpour and Tejas Khot for our discussions and their valuable feedback.

\bibliography{iclr2021_conference}
\bibliographystyle{iclr2021_conference}

\newpage
\appendix

\section{OLS Estimation Under Noisy Measurement}
\label{sec:ols-step-by-step}
\subsection{Causal setting}

Let the Gaussian SCM be defined as follows where the noise term for variable $x$ is defined as $u_{x}$:
\begin{align} \label{eq:appdx_causalSCM}
\begin{split}
    z &= u_z,  \\
    x_{1} &= b z + u_{x_1}, \\
    x_{2} &= c z + u_{x_2}, \\
    y &= a x_{1} + u_y,
\end{split} 
\begin{split}
    u_z &\sim \mathcal{N}(0, \sigma^2_{u_z}) \\
    u_{x_1} &\sim \mathcal{N}(0, \sigma^2_{u_{x1}}) \\
    u_{x_2} &\sim \mathcal{N}(0, \sigma^2_{u_{x2}}) \\
    u_y &\sim \mathcal{N}(0, \sigma^2_{u_y}).
\end{split} 
\end{align}
\begin{align}
\begin{split}
\sigma^2_{x_1} &= b^2\sigma^2_{u_z} + \sigma^2_{u_{x1}}\\
\sigma^2_{x_2} &= c^2\sigma^2_{u_z} + \sigma^2_{u_{x2}}\\
\sigma_{x_1,x_2} &= bc\sigma^2_{u_z}\\
\sigma_{x_1,y} &= ab^2\sigma^2_{u_{z}} + a\sigma^2_{u_{x1}}\\
\sigma_{x_2,y} &= abc\sigma^2_{u_{z}}
\end{split}
\end{align}
Then if we were to solve the linear regression problem $y = x_1 \beta_1 + x_2 \beta_2 + \beta_0$, then using Eq. \ref{eq:ols-est}  we obtain the following values for $\beta_{0}^{ols}$, $\beta_{1}^{ols}$ and $\beta_{2}^{ols}$:
\begin{align}
\beta_1^{ols} = \frac{\sigma^2_{x_{2}}\sigma_{x_{1},y} - \sigma_{x_{1},x_{2}}\sigma_{x_{2},y}}{\sigma^2_{x_{1}}\sigma^2_{x_{2}} - {\sigma^2_{x_{1},x_{2}}}}
 &= \frac{( c^2\sigma^2_{u_z} + \sigma^2_{u_{x2}}) (ab^2\sigma^2_{u_{z}} + a\sigma^2_{u_{x1}})   - (bc\sigma^2_{u_z})(abc\sigma^2_{u_{z}})}{(b^2\sigma^2_{u_z} + \sigma^2_{u_{x1}})( c^2\sigma^2_{u_z} + \sigma^2_{u_{x2}}) - b^2c^2\sigma^4_{u_z}} \\ \nonumber
 &= a \frac{(b^2\sigma^2_{u_z} + \sigma^2_{u_{x1}})( c^2\sigma^2_{u_z} + \sigma^2_{u_{x2}}) - b^2c^2\sigma^4_{u_z}}{(b^2\sigma^2_{u_z} + \sigma^2_{u_{x1}})( c^2\sigma^2_{u_z} + \sigma^2_{u_{x2}}) - b^2c^2\sigma^4_{u_z}} = a
\end{align}
\begin{align}
\beta_2^{ols} = \frac{\sigma^2_{x_{1}}\sigma_{x_{2},y} - \sigma_{x_{1},x_{2}}\sigma_{x_{1},y}}{\sigma^2_{x_{1}}\sigma^2_{x_{2}} - {\sigma^2_{x_{1},x_{2}}}} &= \frac{ (b^2\sigma^2_{u_z} + \sigma^2_{u_{x1}})(abc\sigma^2_{u_{z}}) - ({bc\sigma^2_{u_z}})(ab^2\sigma^2_{u_{z}} + a\sigma^2_{u_{x1}}) }{(b^2\sigma^2_{u_z} + \sigma^2_{u_{x1}})( c^2\sigma^2_{u_z} + \sigma^2_{u_{x2}}) - b^2c^2\sigma^4_{u_z}} = 0
\end{align}
However, if the setting is slightly different, and we observe a noisy version of $x_1$, given by $\widetilde{x_1}$:
\begin{align}
\widetilde{x_1} &= x_1 + \epsilon_{x1}, \quad \epsilon_{x_1} \sim \mathcal{N}(0, \sigma^2_{\epsilon_{x1}})
\end{align}
Since $\epsilon_{x_1} \ind  (x_1, x_2, y)$,
\begin{align}
\sigma^2_{\widetilde{x_1}} &= \textrm{Var}[x_1+\epsilon_{x_1}] = b^2\sigma^2_{u_z} + \sigma^2_{u_{x1}} + \sigma^2_{\epsilon_{x1}} \\
\sigma_{\widetilde{x_{1}},Y} &= \sigma_{x_{1},Y} = \R{E}[(bz + u_{x_1})(ax_1  + u_{y})]  = ab^2\sigma^2_{u_{z}} + a\sigma^2_{u_{x1}}\\
\sigma_{\widetilde{x_{1}},x_{2}} &= \sigma_{X_{1},X_{2}} = bc\sigma^2_{u_z}
\end{align}
Plugging these values into Eq. \ref{eq:ols-est} we get the OLS estimates $\widehat{\beta_1^{ols}}$ and $\widehat{\beta_2^{ols}}$ in the presence of observation noise on $X_1$:
\begin{align}
\begin{split}
\widehat{\beta_1^{ols}} &=  \frac{\sigma^2_{x_{2}}\sigma_{\widetilde{x_1},y} - \sigma_{\widetilde{x_{1}},x_{2}}\sigma_{x_{2},y}}{\sigma^2_{\widetilde{x_1}}\sigma^2_{x_2} - {\sigma^2_{\widetilde{x_1},x_2}}} 
                    =  \frac{
                        \begin{matrix}
                           (c^2\sigma^2_{u_z} + \sigma^2_{u_{x2}})(ab^2\sigma^2_{u_{z}} + a\sigma^2_{u_{x1}}) - ( bc\sigma^2_{u_z})( abc\sigma^2_{u_{z}})
                        \end{matrix}
                        }
                        {
                        \begin{matrix}
                          ( b^2\sigma^2_{u_z} + \sigma^2_{u_{x1}} + \sigma^2_{\epsilon_{x1}})(c^2\sigma^2_{u_z} + \sigma^2_{u_{x2}}) -
                          b^2c^2\sigma^4_{u_z}
                        \end{matrix}
                        }\\
                    &=  \frac{
                        \begin{matrix}
                            a(\sigma^2_{u_{z}}(b^2\sigma^2_{u_{x2}}
                            +c^2\sigma^2_{u_{x1}})
                            +\sigma^2_{u_{x1}}\sigma^2_{u_{x2}})
                        \end{matrix}
                        }
                        {
                        \begin{matrix}
                            \sigma^2_{u_{z}}(b^2\sigma^2_{u_{x2}}+c^2\sigma^2_{u_{x1}})
                            +\sigma^2_{u_{x1}}\sigma^2_{u_{x2}}
                            +\sigma^2_{\epsilon_{x1}}(c^2\sigma^2_{u_z}+\sigma^2_{u_{x2}})
                        \end{matrix}
                        }\\
&= \frac{\beta_1^{ols}}{1+\lambda_{c}}\\
\lambda_{c} &= \frac{\sigma^2_{\epsilon_{x1}}(c^2\sigma^2_{u_z}+\sigma^2_{u_{x2}})}{\sigma^2_{u_{z}}(b^2\sigma^2_{u_{x2}}+c^2\sigma^2_{u_{x1}})+\sigma^2_{u_{x1}}\sigma^2_{u_{x2}}}\\
\widehat{\beta_2^{ols}} &=  \frac{\sigma^2_{\widetilde{x_{1}}}\sigma_{x_{2},y} - \sigma_{\widetilde{x_{1}},x_{2}}\sigma_{\widetilde{x_{1}},y}}{\sigma^2_{\widetilde{x_{1}}}\sigma^2_{x_{2}} - {\sigma^2_{\widetilde{x_{1}},x_{2}}}} =  \frac{
                        (b^2\sigma^2_{u_z} + \sigma^2_{u_{x1}} + \sigma^2_{\epsilon_{x1}}) abc\sigma^2_{u_{z}} - (bc\sigma^2_{u_z})(ab^2\sigma^2_{u_{z}} + a\sigma^2_{u_{x1}})
                        }
                        {
                        \begin{matrix}
                        \sigma^2_{u_{z}}(b^2\sigma^2_{u_{x2}}+c^2\sigma^2_{u_{x1}})+\sigma^2_{u_{x1}}\sigma^2_{u_{x2}}
                        +\sigma^2_{\epsilon_{x1}}(c^2\sigma^2_{u_z}+\sigma^2_{u_{x2}})
                        \end{matrix}
                        } \\
                    &=\frac{
                        acb\sigma^2_{\epsilon_{x1}}\sigma^2_{u_z}
                        }
                        {
                        \begin{matrix}
                        \sigma^2_{u_{z}}(b^2\sigma^2_{u_{x2}}+c^2\sigma^2_{u_{x1}})+\sigma^2_{u_{x1}}\sigma^2_{u_{x2}}
                        +\sigma^2_{\epsilon_{x1}}(c^2\sigma^2_{u_z}+\sigma^2_{u_{x2}})
                        \end{matrix}
                        }
\end{split}
\end{align}
As we can see $\lambda_c > 0$ and  $\lambda_c \propto \sigma^2_{\epsilon_{x1}}$. This shows us that as $\sigma^2_{\epsilon_{x1}}$ increases, $|\widehat{\beta_1^{ols}}|$ (magnitude of the coefficient for $X_1$) decreases and $|\widehat{\beta_2^{ols}}|$ (magnitude of the coefficient for $X_2$) increases. $\lim\limits_{\sigma^2_{\epsilon_{x1}} \rightarrow \infty}\widehat{\beta_1^{ols}}  = 0$, and
$\lim\limits_{\sigma^2_{\epsilon_{x1}} \rightarrow \infty}\widehat{\beta_2^{ols}}  = \frac{acb\sigma^2_{u_z}}{c^2\sigma^2_{u_z}+\sigma^2_{u_{x_2}}}$.

\subsection{Anticausal setting} \label{subsec:anticausal}

Once again we assume that each variable $V$ is a linear function of its parents $\textrm{Pa}(V)$. The noise terms are assumed to be Gaussian and are jointly independent. 
\begin{align} \label{eq:appdx_anticausalSCM}
\begin{split}
    z &= u_z, \\
    q &= a z + u_q,\\
    y &= b z + u_y,\\
    x_{2} &= c q + u_{x_{2}}, \\
    x_{1} &= d y + u_{x_{1}},
\end{split}
\begin{split}
    u_z &\sim \mathcal{N}(0, \sigma^2_{u_z}) \\
    u_q &\sim \mathcal{N}(0, \sigma^2_{u_q}) \\
    u_y &\sim \mathcal{N}(0, \sigma^2_{u_y}) \\
    u_{x_{1}} &\sim \mathcal{N}(0, \sigma^2_{u_{x1}}) \\
    u_{x_{2}} &\sim \mathcal{N}(0, \sigma^2_{u_{x2}})
\end{split}
\end{align}
\begin{align}
\begin{split}
\sigma^2_{x_1} &= d^2b^2\sigma^2_{u_z} + d^2\sigma^2_{u_{y}} + \sigma^2_{u_{x1}}\\
\sigma^2_{x_2} &= c^2a^2\sigma^2_{u_z} + c^2\sigma^2_{u_{q}} + \sigma^2_{u_{x2}}\\
\sigma_{x_{1},x_{2}} &= abcd\sigma^2_{u_z}\\
\sigma_{x_{1},y} &= db^2\sigma^2_{u_{z}} + d\sigma^2_{u_{y}}\\
\sigma_{x_{2},y} &= abc\sigma^2_{u_{z}}
\end{split}
\end{align}

If we were to solve the linear regression problem $y = x_1 \beta_1 + x_2 \beta_2 + \beta_0$, then using Eq. \ref{eq:ols-est}  we get the OLS estimates $\beta_1^{ols}$ and $\beta_2^{ols}$:
\begin{align}
\begin{split}
\beta_1^{ols} &= \frac{\sigma^2_{x_{2}}\sigma_{{x_1},y} - \sigma_{{x_{1}},x_{2}}\sigma_{x_{2},y}}{\sigma^2_{{x_1}}\sigma^2_{x_2} - {\sigma_{{x_1},x_2}}^2}  \\
        &=\frac{(c^2a^2\sigma^2_{u_z} + c^2\sigma^2_{u_{q}} + \sigma^2_{u_{x2}})(db^2\sigma^2_{u_{z}} + d\sigma^2_{u_{y}}) - (abcd\sigma^2_{u_z})(abc\sigma^2_{u_{z}})}{(d^2b^2\sigma^2_{u_z} + d^2\sigma^2_{u_{y}} + \sigma^2_{u_{x1}})( c^2a^2\sigma^2_{u_z} + c^2\sigma^2_{u_{q}} + \sigma^2_{u_{x2}}) - (a^2b^2c^2d^2{\sigma^2_{u_z}}^2)} \\
            &=\frac{
                d(a^2c^2\sigma^2_{u_z} \sigma^2_{u_{y}} +(c^2\sigma^2_{u_q} + \sigma^2_{u_{x2}})(b^2\sigma^2_{u_z}+\sigma^2_{u_y}))
                }
                {
                \begin{matrix}
                    (d^2b^2\sigma^2_{u_z}+\sigma^2_{u_{x1}}+d^2\sigma^2_{u_y})
                    (\sigma^2_{u_{x2}}+c^2\sigma^2_{u_q})
                    +(\sigma^2_{u_{x1}}+d^2\sigma^2_{u_y})c^2a^2\sigma^2_{u_z}
                \end{matrix}
                }\\
                \\
\beta_2^{ols} &= \frac{\sigma^2_{{x_{1}}}\sigma_{x_{2},y} - \sigma_{{x_{1}},x_{2}}\sigma_{{x_{1}},y}}{\sigma^2_{{x_{1}}}\sigma^2_{x_{2}} - {\sigma_{{x_{1}},x_{2}}}^2} \\
                &= \frac{ (d^2b^2\sigma^2_{u_z} + d^2\sigma^2_{u_{y}} + \sigma^2_{u_{x1}})(abc\sigma^2_{u_{z}}) - (abcd\sigma^2_{u_z})( db^2\sigma^2_{u_{z}} + d\sigma^2_{u_{y}})
                }{(d^2b^2\sigma^2_{u_z} + d^2\sigma^2_{u_{y}} + \sigma^2_{u_{x1}})( c^2a^2\sigma^2_{u_z} + c^2\sigma^2_{u_{q}} + \sigma^2_{u_{x2}}) - (a^2b^2c^2d^2{\sigma^2_{u_z}}^2)}\\
                &=\frac{
                abc\sigma^2_{u_z}\sigma^2_{u_{x1}}
                }
                {
                \begin{matrix}
                (d^2b^2\sigma^2_{u_z}+\sigma^2_{u_{x1}}+d^2\sigma^2_{u_y})(\sigma^2_{u_{x2}}+c^2\sigma^2_{u_q})
                +(\sigma^2_{u_{x1}}+d^2\sigma^2_{u_y})c^2a^2\sigma^2_{u_z}
                \end{matrix}
                } \label{eq:appdx_anticaus-ols}
\end{split}
\end{align}
However, if the setting is slightly different, and we observe a noisy version of $x_1$, given by $\widetilde{x_1}$:
\begin{align}
\widetilde{x_1} = x_1 + \epsilon_{x_1}, \quad\quad
\epsilon_{x_1} \sim \mathcal{N}(0, \sigma^2_{\epsilon_{x1}})
\end{align}
Since $\epsilon_{x_1} \ind  x_2, y$, in order to obtain expressions for the OLS estimates $\widehat{\beta_1^{ols}}, \widehat{\beta_2^{ols}}$ in the presence of observation noise, in Eq. \ref{eq:appdx_anticaus-ols} we only need to replace $\sigma^2_{u_{x1}}$ with $\sigma^2_{u_{\widetilde{x1}}}$, which is given by:
\begin{align}
    \sigma^2_{u_{\widetilde{x1}}} = \sigma^2_{u_{x1}} + \sigma^2_{\epsilon_{x1}}
\end{align}

\begin{align}
\widehat{\beta_1^{ols}}  &=\frac{
                d(a^2c^2\sigma^2_{u_z} \sigma^2_{u_{y}} +(c^2\sigma^2_{u_q} + \sigma^2_{u_{x2}})(b^2\sigma^2_{u_z}+\sigma^2_{u_y}))
                }
                {
                \begin{matrix}
                    (d^2b^2\sigma^2_{u_z}+\sigma^2_{u_{\widetilde{x1}}}+d^2\sigma^2_{u_y})
                    (\sigma^2_{u_{x2}}+c^2\sigma^2_{u_q})
                    +(\sigma^2_{u_{\widetilde{x1}}}+d^2\sigma^2_{u_y})c^2a^2\sigma^2_{u_z}
                \end{matrix}
                }\nonumber \\
            &=\frac{
                d(a^2c^2\sigma^2_{u_z} \sigma^2_{u_{y}} +(c^2\sigma^2_{u_q} + \sigma^2_{u_{x2}})(b^2\sigma^2_{u_z}+\sigma^2_{u_y}))
                }
                {
                \begin{matrix}
                    (d^2b^2\sigma^2_{u_z}+(\sigma^2_{u_{x1}} + \sigma^2_{\epsilon_{x1}})+d^2\sigma^2_{u_y})
                    (\sigma^2_{u_{x2}}+c^2\sigma^2_{u_q})
                    +((\sigma^2_{u_{x1}} + \sigma^2_{\epsilon_{x1}})+d^2\sigma^2_{u_y})c^2a^2\sigma^2_{u_z}
                \end{matrix}
                } 
\end{align}
\begin{align}
\widehat{\beta_2^{ols}}      &= \frac{
                abc\sigma^2_{u_z}\sigma^2_{u_{\widetilde{x_1}}}
                }
                {
                \begin{matrix}
                (d^2b^2\sigma^2_{u_z}+\sigma^2_{u_{\widetilde{x_1}}}+d^2\sigma^2_{u_y})(\sigma^2_{u_{x2}}+c^2\sigma^2_{u_q})
                +(\sigma^2_{u_{\widetilde{x_1}}}+d^2\sigma^2_{u_y})c^2a^2\sigma^2_{u_z}
                \end{matrix}
                }  \nonumber \\  
            &=\frac{
                abc\sigma^2_{u_z}(\sigma^2_{u_{x1}} + \sigma^2_{\epsilon_{x1}})
                }
                {
                \begin{matrix}
                (d^2b^2\sigma^2_{u_z}+(\sigma^2_{u_{x1}} + \sigma^2_{\epsilon_{x1}})+d^2\sigma^2_{u_y})(\sigma^2_{u_{x2}}+c^2\sigma^2_{u_q})
                +((\sigma^2_{u_{x1}} + \sigma^2_{\epsilon_{x1}})+d^2\sigma^2_{u_y})c^2a^2\sigma^2_{u_z}
                \end{matrix}
                } 
\end{align}

\begin{align}
   \widehat{\beta_1^{ols}} &=\frac{\beta_1^{ols}}{1+\lambda^{x_1}_{ac}} \quad\quad\quad\quad \widehat{\beta_2^{ols}} = \frac{\beta_2^{ols}}{1+\lambda^{x_1}_{ac}}\left[1+\frac{\sigma^2_{\epsilon_{x_1}}}{\sigma^2_{u_{x_1}}}\right] 
\end{align}

\begin{align}
\lambda^{x_1}_{ac} &= \frac{\sigma^2_{\epsilon_{x_1}}(c^2a^2\sigma^2_{u_z} + c^2\sigma^2_{u_q} + \sigma^2_{u_{x2}})}{(d^2b^2\sigma^2_{u_z}+\sigma^2_{u_{x_1}}+d^2\sigma^2_{u_y})(\sigma^2_{u_{x2}}+c^2\sigma^2_{u_q})+(\sigma^2_{u_{x1}}+d^2\sigma^2_{u_y})c^2a^2\sigma^2_{u_z}}
\end{align}
where $\lambda^{x_1}_{ac} > 0$ and  $\lambda^{x_1}_{ac} \propto \sigma^2_{\epsilon_{x1}}$. Thus, as $\sigma^2_{\epsilon_{x1}}$ increases, $|\widehat{\beta_1^{ols}}|$ decreases. The asymptotic OLS estimates in the presence of infinite observational noise can be seen to be: $\lim\limits_{\sigma^2_{\epsilon_{x1}} \rightarrow \infty} \widehat{\beta_1^{ols}} = 0$
, where as $\lim\limits_{\sigma^2_{\epsilon_{x1}} \rightarrow \infty} \widehat{\beta_2^{ols}} =  \beta_2^{ols}\frac{((d^2b^2\sigma^2_{u_z}+\sigma^2_{u_{x1}}+d^2\sigma^2_{u_y})(\sigma^2_{u_{x2}}+c^2\sigma^2_{u_q})+(\sigma^2_{u_{x1}}+d^2\sigma^2_{u_y})c^2a^2\sigma^2_{u_z})}{(\sigma^2_{u_{x1}}(c^2a^2\sigma^2_{u_z} + c^2\sigma^2_{u_q} + \sigma^2_{u_{x2}}))}$.

Similarly, if we observe a noisy version of $X_2$, given by $\widetilde{X_2}$:
\begin{align}
\widetilde{x_2} = x_2 + \epsilon_{x_2}, \quad\quad
\epsilon_{x_2} \sim \mathcal{N}(0, \sigma^2_{\epsilon_{x2}})
\end{align}
Since $\epsilon_{x_2} \ind  x_1, y$, in order to obtain expressions for the OLS estimates $\widehat{\beta_1^{ols}}, \widehat{\beta_2^{ols}}$ in the presence of observation noise on non-causal features, in Eq. \ref{eq:appdx_anticaus-ols} we only need to replace $\sigma^2_{u_{x2}}$ with $\sigma^2_{u_{\widetilde{x2}}}$, which is given by:
\begin{align}
    \sigma^2_{u_{\widetilde{x2}}} = \sigma^2_{u_{x2}} + \sigma^2_{\epsilon_{x2}}
\end{align}

\begin{align}
  \widehat{\beta_1^{ols}}    &=\frac{
                d(a^2c^2\sigma^2_{u_z} \sigma^2_{u_{y}} +(c^2\sigma^2_{u_q} + \sigma^2_{u_{\widetilde{x_2}}})(b^2\sigma^2_{u_z}+\sigma^2_{u_y}))
                }
                {
                \begin{matrix}
                    (d^2b^2\sigma^2_{u_z}+\sigma^2_{u_{x1}}+d^2\sigma^2_{u_y})
                    (\sigma^2_{u_{\widetilde{x_2}}}+c^2\sigma^2_{u_q})
                    +(\sigma^2_{u_{x1}}+d^2\sigma^2_{u_y})c^2a^2\sigma^2_{u_z}
                \end{matrix}
                } \nonumber\\ 
        &=\frac{
                d(a^2c^2\sigma^2_{u_z} \sigma^2_{u_{y}} +(c^2\sigma^2_{u_q} + (\sigma^2_{u_{x2}} + \sigma^2_{\epsilon_{x2}}))(b^2\sigma^2_{u_z}+\sigma^2_{u_y}))
                }
                {
                \begin{matrix}
                    (d^2b^2\sigma^2_{u_z}+\sigma^2_{u_{x1}}+d^2\sigma^2_{u_y})
                    ((\sigma^2_{u_{x2}} + \sigma^2_{\epsilon_{x2}})+c^2\sigma^2_{u_q})
                    +(\sigma^2_{u_{x1}}+d^2\sigma^2_{u_y})c^2a^2\sigma^2_{u_z}
                \end{matrix}
                }
\end{align}

\begin{align}
  \widehat{\beta_2^{ols}} &=\frac{
                abc\sigma^2_{u_z}\sigma^2_{u_{x1}}
                }
                {
                \begin{matrix}
                (d^2b^2\sigma^2_{u_z}+\sigma^2_{u_{x1}}+d^2\sigma^2_{u_y})( \sigma^2_{u_{\widetilde{x_2}}}+c^2\sigma^2_{u_q})
                +(\sigma^2_{u_{x1}}+d^2\sigma^2_{u_y})c^2a^2\sigma^2_{u_z}
                \end{matrix}
                }  \\
                &= \frac{
                abc\sigma^2_{u_z}\sigma^2_{u_{x1}}
                }
                {
                \begin{matrix}
                (d^2b^2\sigma^2_{u_z}+\sigma^2_{u_{x1}}+d^2\sigma^2_{u_y})( (\sigma^2_{u_{{x_2}}} + \sigma^2_{\epsilon_{x2}})+c^2\sigma^2_{u_q})
                +(\sigma^2_{u_{x1}}+d^2\sigma^2_{u_y})c^2a^2\sigma^2_{u_z}
                \end{matrix}
                } 
\end{align}

\begin{align}
\begin{split}
\widehat{\beta_1^{ols}} = \frac{\beta_1^{ols}}{1+\lambda^{x_2}_{ac}}\left[1+\frac{\sigma^2_{\epsilon_{x2}}(b^2\sigma^2_{u_z}+\sigma^2_{u_y})}{a^2c^2\sigma^2_{u_z}\sigma^2_{u_y} +(c^2\sigma^2_{u_q}+\sigma^2_{u_{x2}})(b^2\sigma^2_{u_z}+\sigma^2_{u_y})}\right]
\quad 
\widehat{\beta_2^{ols}} = \frac{\beta_2^{ols}}{1+\lambda^{x_2}_{ac}}
\\
\lambda^{x_2}_{ac} = \frac{\sigma^2_{\epsilon_{x2}}(d^{2}b^{2}\sigma^2_{u_z}+\sigma^2_{u_{x1}}+d^2\sigma^2_{u_y})}{(d^2b^2\sigma^2_{u_z}+\sigma^2_{u_{x1}}+d^2\sigma^2_{u_y})(\sigma^2_{u_{x2}}+c^2\sigma^2_{u_q})+(\sigma^2_{u_{x1}}+d^2\sigma^2_{u_y})c^2a^2\sigma^2_{u_z}}
\end{split}
\end{align}
where $\lambda^{x_2}_{ac} > 0$ and  $\lambda^{x_2}_{ac} \propto \sigma^2_{\epsilon_{x2}}$. Thus, as $\sigma^2_{\epsilon_{x2}}$ increases, $|\widehat{\beta_1^{ols}}|$ increases. The asymptotic OLS estimates in the presence of infinite observational noise can be seen to be: $\lim\limits_{\sigma^2_{\epsilon_{x2}} \rightarrow \infty} 
\widehat{\beta_2^{ols}} = 0$
, where as $\lim\limits_{\sigma^2_{\epsilon_{x2}} \rightarrow \infty} \widehat{\beta_1^{ols}} =  \beta_1^{ols} \frac{
(b^2\sigma^2_{u_z}+\sigma^2_{u_y})((d^2b^2\sigma^2_{u_z}+\sigma^2_{u_{x1}}+d^2\sigma^2_{u_y})(\sigma^2_{u_{x2}}+c^2\sigma^2_{u_q})+(\sigma^2_{u_{x1}}+d^2\sigma^2_{u_y})c^2a^2\sigma^2_{u_z}) }   {
(a^2c^2\sigma^2_{u_z}\sigma^2_{u_y} +(c^2\sigma^2_{u_q}+\sigma^2_{u_{x2}})(b^2\sigma^2_{u_z}+\sigma^2_{u_y}))(d^{2}b^{2}\sigma^2_{u_z}+\sigma^2_{u_{x1}}+d^2\sigma^2_{u_y})}$.

\clearpage

\section{Model Implementation Details for Section \ref{sec:results}}
\label{sec:model_details}
\paragraph{Standard Methods} 
We use \texttt{scikit-learn} \citep{scikit-learn} 
implementations of SVMs and Na\"ive Bayes for sentiment analysis. 
We train these models on TF-IDF bag of words feature representations 
of the reviews \citep{jones1972statistical}. 
We identify parameters for both classifiers 
using grid search conducted over the validation set. 

\paragraph{BiLSTM} We restrict the vocabulary to the most frequent $20k$ tokens, 
replacing out-of-vocabulary tokens by \texttt{UNK}. 
We fix the maximum input length at $330$ tokens 
when training on reviews from \citet{kaushik2020learning} 
and $2678$ when doing so on \citet{zaidan2007using}, and pad smaller reviews.
Each token is represented by a randomly-initialized $300$-dimensional embedding.
Our model consists of a bidirectional LSTM (hidden dimension $128$) 
with recurrent dropout (probability $0.5$) 
and self attention following the embedding layer.
We use the self attention implementation 
discussed in \citet{lin2016structured} 
with hyperparameter values $d=64$ and $r=64$. 
To generate output, 
we feed this (fixed-length) representation through
a fully-connected hidden layer
(hidden dimension $32$), 
and then a fully-connected output layer 
with softmax activation. 
We train all models for a maximum of $20$ epochs 
using Adam \citep{kingma2014adam}, 
with a learning rate of $\expnumber{1}{-4}$ and a batch size of $16$.  
We apply early stopping when validation loss does not decrease for $5$ epochs.

\paragraph{Pretrained Transformers} 
We use off-the-shelf uncased BERT Base and Longformer Base 
models \citep{wolf2019huggingface}, fine-tuning for each task.
We used BERT for experiments on the smaller IMDb dataset 
used by \citet{kaushik2020learning} 
(with a maximum review length of 330 tokens) 
and Longformer for the dataset 
presented by \citet{zaidan2007using}
(with maximum review length of 2678). 
To account for BERT's sub-word tokenization,
we set the maximum token length is set 
at $350$ for sentiment analysis and $50$ for NLI. 
In case of Longformer, that is $3072$.\footnote{Longformer is better suited to work on longer texts compared to BERT. Maximum length of a review in \citeauthor{zaidan2007using} is $2678$ tokens whereas in \citeauthor{kaushik2020learning} is only $330$ tokens.}
We fine-tune BERT up to $20$ epochs 
with same early stopping criteria as for BiLSTM,
using the BERT Adam optimizer with a batch size of $16$ 
(to fit on a $16$GB Tesla V-$100$ GPU).
We found learning rates of $\expnumber{5}{-5}$ and $\expnumber{1}{-5}$
to work best for sentiment analysis and NLI respectively.
We fine-tune Longformer for $10$ epochs with early stopping, 
using a batch size of 8 (to fit on $64$GB of GPU memory). 

\paragraph{Style Transfer Methods} For \citet{hu2017toward},\footnote{\href{https://github.com/asyml/texar/tree/master/examples/text_style_transfer}{https://github.com/asyml/texar/tree/master/examples/text\_style\_transfer}} \citet{sudhakar2019transforming},\footnote{\href{https://github.com/agaralabs/transformer-drg-style-transfer}{https://github.com/agaralabs/transformer-drg-style-transfer}} and \citet{madaan2020politeness},\footnote{\href{https://github.com/tag-and-generate/}{https://github.com/tag-and-generate/}} we found the default hyperparameters used by the authors to work best on our task. In case of \citet{li2018delete},\footnote{\href{https://github.com/lijuncen/Sentiment-and-Style-Transfer}{https://github.com/lijuncen/Sentiment-and-Style-Transfer}} we followed the training schedule presented in the paper. However, since the paper does not present results on IMDb reviews, we experimented with multiple values of the \emph{salience ratio}, and used a salience ratio of $5.5$ for our downstream task based on transfer accuracy and \textsc{bleu} scores achieved on the validation set.
For all style transfer methods,
we experimented with multiple sequence lengths, 
and found that models worked best 
on sentence level (versus review-level) data, 
with sequence length of $30$, 
truncating longer sentences in the process. 
For each review, we passed individual sentences 
through each model and reconstructed whole reviews 
by joining the resulting \emph{sentiment-flipped} sentences.

\clearpage
\section{Full Results Corresponding to Noise Injection}
\begin{table*}[ht]
  \begin{center}
  \caption{Accuracy of various sentiment analysis classifiers trained on $1.7k$ original reviews from \citet{kaushik2020learning} as noise is injected on \emph{rationales/non-rationales} identified via human feedback.
  \label{tab:iclr_human_noise}}
  \begin{tabular}{ l c c c c c c c c c c c}
    \toprule
    Dataset & \multicolumn{11}{c}{Percent noise in rationales} \\
    \midrule
    \multicolumn{12}{c}{SVM}\\
    \midrule
     & $0$ & $10$ & $20$ & $30$ & $40$ & $50$ & $60$ & $70$ & $80$ & $90$ & $100$\\
    In-sample test & $87.8$ & $88.2$ & $85.7$ & $86.9$ & $86.9$ & $84.5$ & $83.3$ & $81.6$ & $80$ & $79.2$ & $76.7$\\
    CRD & $51.8$ & $47.3$ & $45.7$ & $42.9$ & $39.2$ & $33.5$ & $28.2$ & $25.7$ & $24.1$ & $19.6$ & $17.1$\\
    Amazon & $73.2$ & $72.2$ & $71.3$ & $69.4$ & $67.3$ & $63.7$ & $63.7$ & $58.2$ & $57$ & $50.1$ & $46.5$\\
    Semeval & $62.5$ & $62.2$ & $61.9$ & $61.1$ & $60.9$ & $58.3$ & $57.1$ & $55.4$ & $54.5$ & $51.3$ & $50.1$\\
    Yelp & $79.9$ & $79$ & $77.7$ & $76.7$ & $74.1$ & $71.4$ & $69$ & $65.5$ & $62.4$ & $55.8$ & $51.5$\\
    \midrule
    \multicolumn{12}{c}{BiLSTM with Self Attention}\\
    \midrule
    In-sample test & $81.5$ & $78.8$ & $77.6$ & $76.7$ & $75.3$ & $75.2$ & $74.5$ & $72.8$ & $67.3$ & $64.2$ & $63.8$\\
    CRD & $49.4$ & $49.3$ & $46.3$ & $45.1$ & $39.5$ & $38.1$ & $38.9$ & $38.7$ & $32.6$ & $32.6$ & $29.7$\\
    Amazon & $65.4$ & $69.1$ & $68.5$ & $66.6$ & $63.2$ & $63.9$ & $58.8$ & $50.6$ & $50.6$ & $47.1$ & $44.2$\\
    Semeval & $59.3$ & $59.8$ & $57.6$ & $56.4$ & $58.6$ & $56.6$ & $55.3$ & $54.3$ & $54.3$ & $52.3$ & $50$\\
    Yelp & $71.2$ & $70.8$ & $67.4$ & $65.9$ & $65.3$ & $64.1$ & $63.4$ & $60.1$ & $62.4$ & $49.8$ & $46.4$\\
    \midrule
    \multicolumn{12}{c}{BERT}\\
    \midrule
    In-sample test & $87.4$ & $87.4$ & $86.5$ & $85.7$ & $85.3$ & $84.3$ & $83.6$ & $81$ & $76.6$ & $71$ & $69$\\
    CRD & $82.2$ & $78.1$ & $78.4$ & $75.4$ & $67.6$ & $67.5$ & $65.5$ & $53.9$ & $42.7$ & $36.2$ & $31.8$\\
    Amazon & $76.2$ & $75.5$ & $75.1$ & $74.2$ & $73.5$ & $73$ & $72.5$ & $70.7$ & $63.4$ & $57.8$ & $56.1$\\
    Semeval & $76.4$ & $69.7$ & $66.9$ & $69.8$ & $67.8$ & $67.4$ & $66.8$ & $65.5$ & $62.2$ & $54.9$ & $52.6$\\
    Yelp & $83.7$ & $82.5$ & $82$ & $81.5$ & $80.9$ & $80.2$ & $79.9$ & $75.6$ & $64.3$ & $54.6$ & $52.3$\\
    \midrule
    Dataset & \multicolumn{11}{c}{Percent noise in non-rationales} \\
    \midrule
    \multicolumn{12}{c}{SVM}\\
    \midrule
    In-sample test & $87.8$ & $88.6$ & $89$ & $86.9$ & $85.3$ & $82.4$ & $86.5$ & $83.7$ & $82$ & $81.6$ & $78$\\
    CRD & $51.8$ & $55.9$ & $53.5$ & $57.1$ & $58.8$ & $63.7$ & $63.3$ & $65.7$ & $70.2$ & $73.9$ & $74.3$\\
    Amazon & $73.2$ & $74.9$ & $75.3$ & $77.3$ & $75.8$ & $76.6$ & $76.5$ & $77.4$ & $75.5$ & $75.4$ & $76.9$\\
    Semeval & $62.5$ & $63.3$ & $62.7$ & $64.3$ & $64.3$ & $65.6$ & $66$ & $65.8$ & $65$ & $66.4$ & $66.4$\\
    Yelp & $79.9$ & $80.9$ & $80.1$ & $82.2$ & $83.6$ & $84.1$ & $83.5$ & $83.4$ & $82.7$ & $82.1$ & $81.4$\\
    \midrule
    \multicolumn{12}{c}{BiLSTM with Self Attention}\\
    \midrule
    In-sample test & $81.5$ & $77.5$ & $77$ & $75.9$ & $75.4$ & $75.2$ & $75.1$ & $73.8$ & $73$ & $72.4$ & $71.7$\\
    CRD & $49.4$ & $53.1$ & $56.25$ & $56.6$ & $57.5$ & $58.4$ & $58.6$ & $60.3$ & $61.5$ & $65.5$ & $66.1$\\
    Amazon & $65.4$ & $66.5$ & $66.6$ & $66.6$ & $67.6$ & $67.7$ & $68.3$ & $68.6$ & $68.8$ & $68.5$ & $68.4$\\
    Semeval & $59.3$ & $58.6$ & $58.9$ & $59.3$ & $58.1$ & $57.5$ & $59.2$ & $59.5$ & $59.8$ & $59.5$ & $58$\\
    Yelp & $71.2$ & $74.7$ & $72.5$ & $73.3$ & $73.9$ & $73.6$ & $72.2$ & $74.3$ & $73.7$ & $75.6$ & $75.4$\\
    \midrule
    \multicolumn{12}{c}{BERT}\\
    \midrule
    In-sample test & $87.4$ & $88.2$ & $87$ & $86.9$ & $87$ & $85.8$ & $83.6$ & $78.9$ & $72.5$ & $72.1$ & $71.3$\\
    CRD & $82.2$ & $92.8$ & $92.8$ & $92.3$ & $93.1$ & $92.8$ & $89.8$ & $88.6$ & $84.5$ & $81.3$ & $81$\\
    Amazon & $76.2$ & $78.6$ & $78.9$ & $79.2$ & $75.1$ & $71.7$ & $67.6$ & $65.3$ & $65.2$ & $63.7$ & $61.8$\\
    Semeval & $76.4$ & $74.6$ & $76.3$ & $75.8$ & $70.9$ & $62.1$ & $64.8$ & $63.3$ & $60.8$ & $58.7$ & $58.7$\\
    Yelp & $83.7$ & $85.4$ & $85.3$ & $85.1$ & $82.1$ & $78.3$ & $77.2$ & $76.2$ & $74.3$ & $71.6$ & $70.1$\\
\bottomrule
  \end{tabular}
  \end{center}
\end{table*}

\begin{table*}[ht]
  \begin{center}
  \caption{Accuracy of various sentiment analysis classifiers trained on $1.7k$ original reviews from \citet{kaushik2020learning} as noise is injected on \emph{rationales/non-rationales} identified via Attention masks.
  \label{tab:iclr_attention_noise}}
  \begin{tabular}{ l c c c c c c c c c c c}
    \toprule
    Dataset & \multicolumn{11}{c}{Percent noise in rationales} \\
    \midrule
    \multicolumn{12}{c}{SVM}\\
    \midrule
     & $0$ & $10$ & $20$ & $30$ & $40$ & $50$ & $60$ & $70$ & $80$ & $90$ & $100$\\
    In-sample test & $87.8$ & $85$ & $85.9$ & $86.3$ & $86.3$ & $85.2$ & $84.6$ & $86.3$ & $83.6$ & $84.2$ & $83.6$\\
    CRD & $51.8$ & $50.6$ & $51.8$ & $52$ & $51.8$ & $50$ & $50.6$ & $48.6$ & $48.6$ & $47.5$ & $46.1$\\
    Amazon & $73.2$ & $74.3$ & $73.4$ & $72.8$ & $72.8$ & $72.9$ & $72$ & $72.3$ & $71.1$ & $72$ & $70.3$\\
    Semeval & $62.5$ & $62.8$ & $62.9$ & $61.8$ & $62.5$ & $61.9$ & $61.4$ & $60.7$ & $61.1$ & $60.6$ & $60.1$\\
    Yelp & $79.9$ & $80.1$ & $79.3$ & $78.7$ & $78.9$ & $78.5$ & $77.8$ & $77.5$ & $77.8$ & $76.2$ & $75.9$\\
    \midrule
    \multicolumn{12}{c}{BiLSTM with Self Attention}\\
    \midrule
    In-sample test & $81.5$ & $78.8$ & $78.6$ & $78.3$ & $78.2$ & $76.2$ & $77.3$ & $76.8$ & $71.8$ & $73.2$ & $74.2$\\
    CRD & $49.4$ & $53.3$ & $50$ & $53.4$ & $52.4$ & $49.7$ & $49.2$ & $47.4$ & $47.7$ & $47$ & $44.1$\\
    Amazon & $65.4$ & $66.8$ & $71$ & $64.7$ & $60.7$ & $61.7$ & $65.2$ & $64.6$ & $51.6$ & $57.1$ & $66.4$\\
    Semeval & $59.3$ & $59.5$ & $60.1$ & $57.4$ & $55.9$ & $57.2$ & $52.2$ & $57.6$ & $51.5$ & $51.8$ & $56.1$\\
    Yelp & $71.2$ & $72.3$ & $74.2$ & $69.6$ & $70.5$ & $67.3$ & $70.7$ & $72.8$ & $62.8$ & $65$ & $66.2$\\
    \midrule
    \multicolumn{12}{c}{BERT}\\
    \midrule
    In-sample test & $87.4$ & $93$ & $90.8$ & $90.3$ & $90.6$ & $91.2$ & $90.3$ & $90.4$ & $90.7$ & $90.6$ & $90.3$\\
    CRD & $82.2$ & $91.2$ & $92$ & $90.8$ & $90.8$ & $90.9$ & $90.3$ & $90.9$ & $90.2$ & $89.8$ & $90.4$\\
    Amazon & $76.2$ & $77.3$ & $79.1$ & $78.7$ & $79.8$ & $79.1$ & $79.8$ & $79.5$ & $79.2$ & $78.9$ & $79.3$\\
    Semeval & $76.4$ & $71.4$ & $73.5$ & $73.2$ & $74.4$ & $76.1$ & $77.6$ & $79.8$ & $78.4$ & $79.2$ & $77.8$\\
    Yelp & $83.7$ & $83.5$ & $85.4$ & $84.9$ & $86$ & $85.7$ & $85.9$ & $85.6$ & $85.5$ & $85.4$ & $68.9$\\
    \midrule
    Dataset & \multicolumn{11}{c}{Percent noise in non-rationales} \\
    \midrule
    \multicolumn{12}{c}{SVM}\\
    \midrule
    In-sample test & $87.8$ & $85$ & $85.7$ & $84.8$ & $85$ & $84$ & $83.6$ & $84.6$ & $80.7$ & $81.1$ & $77.3$\\
    CRD & $51.8$ & $50.4$ & $52.2$ & $53.9$ & $50.2$ & $50.8$ & $52.9$ & $54.1$ & $54.1$ & $56.8$ & $56.4$\\
    Amazon & $73.2$ & $73.5$ & $75.3$ & $74.3$ & $76.2$ & $73.9$ & $73.4$ & $73.6$ & $71$ & $70$ & $67.8$\\
    Semeval & $62.5$ & $62.6$ & $63.7$ & $63.7$ & $63.1$ & $62.6$ & $63.5$ & $61.5$ & $62.1$ & $62$ & $59.9$\\
    Yelp & $79.9$ & $79.8$ & $80.9$ & $81.7$ & $80.9$ & $80.5$ & $80$ & $80.1$ & $78.5$ & $77.5$ & $74.4$\\
    \midrule
    \multicolumn{12}{c}{BiLSTM with Self Attention}\\
    \midrule
    In-sample test & $81.5$ & $77.6$ & $76$ & $77.1$ & $77.3$ & $75.4$ & $73.7$ & $67.9$ & $68.6$ & $54.2$ & $52.3$\\
    CRD & $49.4$ & $53.1$ & $52.1$ & $52.1$ & $65$ & $54.1$ & $51.9$ & $53.4$ & $55$ & $52.3$ & $51.6$\\
    Amazon & $65.4$ & $63.7$ & $65.7$ & $64$ & $58.8$ & $65.5$ & $60.3$ & $58.7$ & $61$ & $58.1$ & $56.2$\\
    Semeval & $59.3$ & $54.8$ & $58.4$ & $57.3$ & $60.7$ & $56.8$ & $55.2$ & $54$ & $51.2$ & $50$ & $49.9$\\
    Yelp & $71.2$ & $72$ & $73.6$ & $70.2$ & $61.3$ & $71.5$ & $68.4$ & $64.9$ & $66.3$ & $58.2$ & $55.8$\\
    \midrule
    \multicolumn{12}{c}{BERT}\\
    \midrule
    In-sample test & $87.4$ & $86.9$ & $86.7$ & $85.3$ & $84$ & $81.9$ & $80.6$ & $74$ & $74$ & $73$ & $67.2$\\
    CRD & $82.2$ & $92.3$ & $92.4$ & $92.1$ & $90$ & $86.8$ & $83$ & $73.2$ & $77.7$ & $72.5$ & $68.5$\\
    Amazon & $76.2$ & $79.5$ & $78.5$ & $77.9$ & $69.2$ & $67.4$ & $58.1$ & $55.9$ & $53.5$ & $55.8$ & $52.6$\\
    Semeval & $76.4$ & $76.5$ & $75.7$ & $77.1$ & $65.7$ & $61.8$ & $54.6$ & $58.8$ & $51.8$ & $54$ & $50.8$\\
    Yelp & $83.7$ & $85.8$ & $85$ & $85.5$ & $79.3$ & $78.7$ & $67.8$ & $66.5$ & $59.5$ & $63.2$ & $57.5$\\
    \bottomrule
  \end{tabular}
  \end{center}
\end{table*}

\begin{table*}[ht]
  \begin{center}
  \caption{Accuracy of various sentiment analysis classifiers trained on $1.7k$ original reviews from \citet{kaushik2020learning} as noise is injected on \emph{rationales/non-rationales} identified via Allen NLP Saliency Interpreter.
  \label{tab:iclr_saliency_noise}}
  \begin{tabular}{ l c c c c c c c c c c c}
    \toprule
    Dataset & \multicolumn{11}{c}{Percent noise in rationales} \\
    \midrule
    \multicolumn{12}{c}{SVM}\\
    \midrule
     & $0$ & $10$ & $20$ & $30$ & $40$ & $50$ & $60$ & $70$ & $80$ & $90$ & $100$\\
    In-sample test & $87.8$ & $85.1$ & $85.4$ & $85.1$ & $85.1$ & $83.9$ & $82.5$ & $82.8$ & $81.8$ & $80$ & $77.5$\\
    CRD & $51.8$ & $51.2$ & $52.5$ & $51.1$ & $51$ & $50.1$ & $49.5$ & $46.6$ & $43.7$ & $42.1$ & $40.5$\\
    Amazon & $73.2$ & $73.4$ & $73.65$ & $73.2$ & $72.7$ & $72.9$ & $71.8$ & $72.1$ & $70.5$ & $69.6$ & $68.9$\\
    Semeval & $62.5$ & $62.8$ & $62.5$ & $62.4$ & $61.9$ & $61.2$ & $60.7$ & $60.5$ & $59.6$ & $58.4$ & $57.9$\\
    Yelp & $79.9$ & $79.8$ & $79.7$ & $79.1$ & $78.7$ & $78.2$ & $78.1$ & $76.6$ & $75.1$ & $74.1$ & $72.2$\\
    \midrule
    \multicolumn{12}{c}{BiLSTM with Self Attention}\\
    \midrule
    In-sample test & $81.5$ & $82.3$ & $83.5$ & $80.4$ & $78.2$ & $81.9$ & $80.6$ & $77.8$ & $79.2$ & $76$ & $77$\\
    CRD & $49.4$ & $48.2$ & $48.6$ & $51.2$ & $48.6$ & $47.3$ & $47.1$ & $46.9$ & $44.3$ & $42.6$ & $37.9$\\
    Amazon & $65.4$ & $46.6$ & $72.8$ & $66.9$ & $49.7$ & $55.4$ & $53.7$ & $68.5$ & $54.7$ & $49.8$ & $51.8$\\
    Semeval & $59.3$ & $42.1$ & $49.5$ & $56.2$ & $54.7$ & $52.7$ & $53.7$ & $50.1$ & $51.2$ & $50.2$ & $50$\\
    Yelp & $71.2$ & $69$ & $73.3$ & $73.2$ & $67.8$ & $69.2$ & $69.5$ & $68.8$ & $67$ & $54.4$ & $56.9$\\
    \midrule
    \multicolumn{12}{c}{BERT}\\
    \midrule
    In-sample test & $87.4$ & $91.1$ & $90.6$ & $90$ & $88$ & $89.1$ & $87.4$ & $86.3$ & $83.6$ & $84.5$ & $81.6$\\
    CRD & $82.2$ & $93.4$ & $92.3$ & $91.9$ & $90.3$ & $90.2$ & $87.7$ & $83.8$ & $78$ & $79.3$ & $70$\\
    Amazon & $76.2$ & $82.4$ & $81.3$ & $79.8$ & $77.2$ & $77.6$ & $77.8$ & $75.6$ & $69.7$ & $69.4$ & $73.6$\\
    Semeval & $76.4$ & $82.6$ & $82.8$ & $81.3$ & $79.2$ & $78.1$ & $76.7$ & $74.7$ & $67.4$ & $65.8$ & $67.4$\\
    Yelp & $83.7$ & $88.3$ & $88.8$ & $88.5$ & $87.8$ & $88.1$ & $87$ & $86.2$ & $84.4$ & $83.3$ & $82.7$\\
    \midrule
    Dataset & \multicolumn{11}{c}{Percent noise in non-rationales} \\
    \midrule
    \multicolumn{12}{c}{SVM}\\
    \midrule
    In-sample test & $87.8$ & $85.9$ & $85.7$ & $86.9$ & $83.6$ & $86.9$ & $85.9$ & $83.2$ & $85$ & $81.8$ & $79.1$\\
    CRD & $51.8$ & $52.3$ & $53.7$ & $53.9$ & $56.8$ & $55.3$ & $53.5$ & $54.3$ & $58$ & $60$ & $61.5$\\
    Amazon & $73.2$ & $73.9$ & $74.1$ & $71.8$ & $73.6$ & $72.5$ & $73.8$ & $72.6$ & $70.6$ & $70.6$ & $70.8$\\
    Semeval & $62.5$ & $62.7$ & $62.8$ & $61.3$ & $62.7$ & $62$ & $61.9$ & $63.2$ & $62.3$ & $62.4$ & $63.6$\\
    Yelp & $79.9$ & $79.8$ & $79.8$ & $81.4$ & $81$ & $80.7$ & $81$ & $80.5$ & $80.3$ & $79.8$ & $78.6$\\
    \midrule
    \multicolumn{12}{c}{BiLSTM with Self Attention}\\
    \midrule
    In-sample test & $81.5$ & $81$ & $81.7$ & $80.8$ & $79.8$ & $78$ & $75.6$ & $73$ & $70.4$ & $51$ & $50$\\
    CRD & $49.4$ & $49$ & $49.8$ & $48$ & $47.9$ & $51.6$ & $46.7$ & $53.3$ & $50.2$ & $51.6$ & $48.4$\\
    Amazon & $65.4$ & $65.3$ & $64.9$ & $62.7$ & $63.3$ & $65.3$ & $67.1$ & $65.3$ & $64$ & $58.3$ & $41.8$\\
    Semeval & $59.3$ & $55$ & $61.3$ & $50.1$ & $54.6$ & $58.5$ & $55.2$ & $55.7$ & $49.4$ & $49.6$ & $44$\\
    Yelp & $71.2$ & $73.8$ & $75.1$ & $71.4$ & $74.1$ & $73.4$ & $74.5$ & $72.5$ & $66.9$ & $55.9$ & $53.6$\\
    \midrule
    \multicolumn{12}{c}{BERT}\\
    \midrule
    In-sample test & $87.4$ & $90.5$ & $89.1$ & $88.6$ & $80.6$ & $75.1$ & $70.1$ & $63.7$ & $53.8$ & $54.1$ & $53.1$\\
    CRD & $82.2$ & $92.1$ & $92.2$ & $91.3$ & $79.9$ & $73.3$ & $67.1$ & $59.2$ & $50.1$ & $49.8$ & $49.6$\\
    Amazon & $76.2$ & $77.5$ & $79.2$ & $77.3$ & $69$ & $65.9$ & $61.1$ & $61.7$ & $52.9$ & $52.7$ & $51.4$\\
    Semeval & $76.4$ & $82$ & $83.6$ & $83.1$ & $78.1$ & $77.9$ & $71.1$ & $69.7$ & $55.9$ & $56.5$ & $51.8$\\
    Yelp & $83.7$ & $88$ & $87.4$ & $87.8$ & $76.8$ & $73$ & $66.9$ & $66.3$ & $55.3$ & $54.4$ & $53.1$\\
    \bottomrule
  \end{tabular}
  \end{center}
\end{table*}

\begin{table*}[ht]
  \begin{center}
  \caption{Accuracy of various sentiment analysis classifiers trained on reviews from \citet{zaidan2007using} as noise is injected on \emph{rationales/non-rationales} identified via human feedback.
  \label{tab:zaidan_human_noise}}
  \begin{tabular}{ l c c c c c c c c c c c}
    \toprule
    Dataset & \multicolumn{11}{c}{Percent noise in rationales} \\
    \midrule
    \multicolumn{12}{c}{SVM}\\
    \midrule
     & $0$ & $10$ & $20$ & $30$ & $40$ & $50$ & $60$ & $70$ & $80$ & $90$ & $100$\\
    In-sample test & $87.5$ & $86.2$ & $85.5$ & $85$ & $84.5$ & $83.3$ & $82.5$ & $81.1$ & $78.9$ & $77.5$ & $76.5$\\
    CRD & $46.1$ & $45.6$ & $44.4$ & $43.7$ & $44.1$ & $41.2$ & $38.8$ & $36$ & $34.4$ & $33.1$ & $30.9$\\
    Amazon & $68.6$ & $67.1$ & $65.1$ & $64.2$ & $62.2$ & $60.4$ & $57.9$ & $50.5$ & $54.9$ & $53.5$ & $51.8$\\
    Semeval & $56.7$ & $56.1$ & $55.4$ & $54.8$ & $54.1$ & $53.5$ & $52.7$ & $52$ & $51.6$ & $50.8$ & $50.4$\\
    Yelp & $76.2$ & $75$ & $73.5$ & $72$ & $70.2$ & $68.8$ & $66.6$ & $65.1$ & $63.3$ & $61.1$ & $59.3$\\
    \midrule
    \multicolumn{12}{c}{BiLSTM with Self Attention}\\
    \midrule
    In-sample test & $80.3$ & $82.1$ & $83.2$ & $81.3$ & $78.4$ & $71.1$ & $78.8$ & $77.4$ & $76.9$ & $77.4$ & $75.5$\\
    CRD & $49.2$ & $50.6$ & $51$ & $48.8$ & $48$ & $49.6$ & $49.4$ & $48.8$ & $48.8$ & $47.5$ & $48.4$\\
    Amazon & $50$ & $50.5$ & $49.4$ & $49.7$ & $49.8$ & $49.7$ & $49.7$ & $49.7$ & $49.6$ & $49.5$ & $49.4$\\
    Semeval & $50$ & $50$ & $50$ & $50$ & $50$ & $50$ & $50$ & $50$ & $50$ & $50$ & $50$\\
    Yelp & $50.5$ & $50$ & $53.1$ & $52.1$ & $50.5$ & $50.2$ & $50.1$ & $50$ & $50$ & $50.2$ & $50.1$\\
    \midrule
    \multicolumn{12}{c}{Longformer}\\
    \midrule
    In-sample test & $97.5$ & $96.7$ & $94$ & $90.5$ & $88.3$ & $78.9$ & $81.4$ & $72.6$ & $79.4$ & $78.7$ & $83.5$\\
    CRD & $93.4$ & $93.6$ & $87.5$ & $85.4$ & $84.2$ & $64.1$ & $61.5$ & $54.2$ & $52.7$ & $50.3$ & $48$\\
    Amazon & $81.8$ & $77.9$ & $65.3$ & $65.7$ & $64.7$ & $63.6$ & $61.9$ & $62.1$ & $61.3$ & $60.6$ & $57.9$\\
    Semeval & $80.3$ & $74.9$ & $64$ & $66.9$ & $71.6$ & $61.3$ & $58.4$ & $56.7$ & $58.9$ & $62.1$ & $58.6$\\
    Yelp & $88.6$ & $85.8$ & $77.7$ & $74.6$ & $72.5$ & $68.4$ & $66.5$ & $64.8$ & $64.3$ & $64.9$ & $62.2$\\
    \midrule
    Dataset & \multicolumn{11}{c}{Percent noise in non-rationales} \\
    \midrule
    \multicolumn{12}{c}{SVM}\\
    \midrule
    In-sample test & $87.5$ & $85.5$ & $86$ & $83$ & $82$ & $83$ & $81$ & $80.5$ & $75.5$ & $60$ & $50$\\
    CRD & $46.1$ & $46.1$ & $49$ & $49.4$ & $57.1$ & $55.5$ & $58.4$ & $58.4$ & $56.5$ & $56.3$ & $54$\\
    Amazon & $68.6$ & $67.7$ & $68$ & $67.2$ & $69.4$ & $69$ & $69.7$ & $68.9$ & $69.2$ & $64.9$ & $62.3$\\
    Semeval & $56.7$ & $56.9$ & $57.5$ & $57.4$ & $58.3$ & $57.6$ & $58.8$ & $59.4$ & $59.3$ & $57.4$ & $56.3$\\
    Yelp & $76.2$ & $76.1$ & $76.9$ & $75.9$ & $77$ & $77.4$ & $75.2$ & $74.1$ & $73.3$ & $68.5$ & $61.6$\\
    \midrule
    \multicolumn{12}{c}{BiLSTM with Self Attention}\\
    \midrule
    In-sample test & $80.3$ & $80.8$ & $79.8$ & $75.2$ & $75$ & $62.5$ & $62$ & $57.7$ & $56.7$ & $58.7$ & $57.7$\\
    CRD & $49.2$ & $50$ & $51.1$ & $50.8$ & $52.9$ & $53.9$ & $58.6$ & $58.6$ & $60$ & $60.4$ & $60.8$\\
    Amazon & $50$ & $50$ & $50.7$ & $50.7$ & $50.9$ & $52.2$ & $52.3$ & $53.2$ & $55$ & $55.1$ & $56.7$\\
    Semeval & $50$ & $50$ & $50$ & $50$ & $50$ & $51$ & $51.8$ & $52.7$ & $53.5$ & $53.8$ & $53.9$\\
    Yelp & $50.5$ & $50.4$ & $52.7$ & $52.9$ & $52.9$ & $55.2$ & $58$ & $58.9$ & $64.6$ & $64.6$ & $70$\\
    \midrule
    \multicolumn{12}{c}{Longformer}\\
    \midrule
    In-sample test & $97.5$ & $97.9$ & $98.1$ & $97.4$ & $94.8$ & $93.4$ & $86.4$ & $82.3$ & $76.3$ & $77.4$ & $80.2$\\
    CRD & $93.4$ & $94.7$ & $94.1$ & $91.8$ & $91.4$ & $91.8$ & $88$ & $83.4$ & $83.7$ & $83.6$ & $83.4$\\
    Amazon & $81.8$ & $79$ & $80$ & $81.5$ & $83.2$ & $84.2$ & $84.1$ & $76.3$ & $78.5$ & $79.4$ & $76.9$\\
    Semeval & $80.3$ & $79.4$ & $77.2$ & $80.6$ & $80.6$ & $84.6$ & $85.3$ & $71.8$ & $79.9$ & $83.7$ & $76.6$\\
    Yelp & $88.6$ & $85.3$ & $86.4$ & $89$ & $89.5$ & $89.9$ & $89.9$ & $86.2$ & $86.5$ & $86.4$ & $84.7$\\
\bottomrule
  \end{tabular}
  \end{center}
\end{table*}

\begin{table*}[ht]
  \begin{center}
  \caption{Accuracy of various sentiment analysis classifiers trained on reviews from \citet{zaidan2007using} as noise is injected on \emph{rationales/non-rationales} identified via Attention masks.
  \label{tab:zaidan_attention_noise}}
  \begin{tabular}{ l c c c c c c c c c c c}
    \toprule
    Dataset & \multicolumn{11}{c}{Percent noise in rationales} \\
    \midrule
    \multicolumn{12}{c}{SVM}\\
    \midrule
     & $0$ & $10$ & $20$ & $30$ & $40$ & $50$ & $60$ & $70$ & $80$ & $90$ & $100$\\
    In-sample test & $87.5$ & $85$ & $84.5$ & $84$ & $82.5$ & $83$ & $81$ & $80$ & $77.5$ & $75.5$ & $75.5$\\
    CRD & $46.1$ & $51$ & $50.6$ & $52$ & $51.8$ & $52.3$ & $52.3$ & $51.8$ & $50.2$ & $49.8$ & $49.8$\\
    Amazon & $68.6$ & $68.1$ & $67.1$ & $66.8$ & $66.9$ & $66.5$ & $66.2$ & $65.4$ & $66.1$ & $66.6$ & $65.7$\\
    Semeval & $56.7$ & $56.6$ & $56.3$ & $56.4$ & $56.2$ & $56.4$ & $56.4$ & $56.2$ & $56.8$ & $56.4$ & $56.4$\\
    Yelp & $76.2$ & $76.1$ & $76$ & $76.2$ & $76.4$ & $76.5$ & $76.9$ & $76.9$ & $76.7$ & $76.9$ & $76.5$\\
    \midrule
    \multicolumn{12}{c}{BiLSTM with Self Attention}\\
    \midrule
    In-sample test & $80.3$ & $78.8$ & $77.9$ & $77.9$ & $78.8$ & $67.3$ & $65.9$ & $63.9$ & $62$ & $65.4$ & $58.7$\\
    CRD & $49.2$ & $49.4$ & $50.2$ & $50.2$ & $52.1$ & $51$ & $52.1$ & $52.3$ & $56.3$ & $51.8$ & $54.7$\\
    Amazon & $50$ & $49.7$ & $49.9$ & $49.9$ & $50.4$ & $50.2$ & $51$ & $51.7$ & $51.1$ & $50.7$ & $50.7$\\
    Semeval & $50$ & $50$ & $50$ & $50$ & $50$ & $50$ & $50$ & $50.2$ & $50.1$ & $50$ & $50.1$\\
    Yelp & $50.5$ & $50.1$ & $50.5$ & $50.5$ & $52.1$ & $52.4$ & $56.1$ & $54.9$ & $54.9$ & $52.2$ & $54.9$\\
    \midrule
    \multicolumn{12}{c}{Longformer}\\
    \midrule
    In-sample test & $97.5$ & $97.3$ & $97$ & $96.5$ & $88.3$ & $94$ & $93.8$ & $91.2$ & $91.5$ & $87.2$ & $84$\\
    CRD & $93.4$ & $93.5$ & $93.1$ & $92.8$ & $91.7$ & $91.8$ & $90.7$ & $88$ & $87.5$ & $83.7$ & $80.8$\\
    Amazon & $81.8$ & $76.3$ & $69.5$ & $75.4$ & $70.4$ & $64.5$ & $66.3$ & $60.8$ & $64.7$ & $57.3$ & $55.3$\\
    Semeval & $80.3$ & $73$ & $67.2$ & $75.1$ & $69.6$ & $61.5$ & $67$ & $58.8$ & $67.6$ & $56.4$ & $55.3$\\
    Yelp & $88.6$ & $85.1$ & $79.3$ & $83.9$ & $79.8$ & $75.4$ & $76.8$ & $69.1$ & $75.4$ & $65.7$ & $61$\\
    \midrule
    Dataset & \multicolumn{11}{c}{Percent noise in non-rationales} \\
    \midrule
    \multicolumn{12}{c}{SVM}\\
    \midrule
    In-sample test & $87.5$ & $87$ & $86.5$ & $87.5$ & $81$ & $82.5$ & $73$ & $52$ & $50$ & $50$ & $50$\\
    CRD & $46.1$ & $50.4$ & $49.6$ & $48.6$ & $50$ & $46.9$ & $50.6$ & $49.6$ & $50.4$ & $50.2$ & $50.2$\\
    Amazon & $68.6$ & $66.7$ & $66.8$ & $64.1$ & $65.9$ & $63.2$ & $62.2$ & $60$ & $57.8$ & $56.2$ & $56.3$\\
    Semeval & $56.7$ & $56.3$ & $56.8$ & $55.9$ & $56.7$ & $55$ & $54.2$ & $53.8$ & $51.8$ & $51.1$ & $51$\\
    Yelp & $76.2$ & $74.8$ & $74.2$ & $71.1$ & $71$ & $64.9$ & $59.7$ & $55.2$ & $52.3$ & $51$ & $50$\\
    \midrule
    \multicolumn{12}{c}{BiLSTM with Self Attention}\\
    \midrule
    In-sample test & $80.3$ & $79.8$ & $81.3$ & $78.4$ & $63.5$ & $67.3$ & $49.5$ & $49$ & $48.1$ & $48.4$ & $48.1$\\
    CRD & $49.2$ & $51.4$ & $51.4$ & $54.5$ & $49.8$ & $49.4$ & $49.6$ & $49.4$ & $49.4$ & $49.4$ & $49.4$\\
    Amazon & $50$ & $49.9$ & $50.6$ & $50.4$ & $50.1$ & $49.7$ & $49.6$ & $49.5$ & $49.5$ & $49.5$ & $49.5$\\
    Semeval & $50$ & $50$ & $50$ & $50.2$ & $50$ & $50$ & $50$ & $50$ & $50$ & $50$ & $50$\\
    Yelp & $50.5$ & $52.3$ & $52.7$ & $56.9$ & $51$ & $50.4$ & $50$ & $50$ & $50$ & $50$ & $50$\\
    \midrule
    \multicolumn{12}{c}{Longformer}\\
    \midrule
    In-sample test & $97.5$ & $98.2$ & $97.8$ & $95$ & $90.2$ & $83.3$ & $67.3$ & $62.8$ & $69.3$ & $64.2$ & $52.8$\\
    CRD & $93.4$ & $93.6$ & $93.5$ & $88.8$ & $83.1$ & $76.5$ & $67.8$ & $69.7$ & $77.6$ & $54.5$ & $51.4$\\
    Amazon & $81.8$ & $81.6$ & $97.8$ & $95$ & $90.2$ & $83.3$ & $67.3$ & $62.8$ & $79.3$ & $64.2$ & $52.8$\\
    Semeval & $80.3$ & $74.8$ & $70.3$ & $79.1$ & $79$ & $78.9$ & $69.5$ & $67.9$ & $64$ & $63.3$ & $58.6$\\
    Yelp & $88.6$ & $83.9$ & $83.1$ & $89.5$ & $90.2$ & $89.7$ & $87.6$ & $83$ & $78.8$ & $62.4$ & $59.4$\\
    \bottomrule
  \end{tabular}
  \end{center}
\end{table*}

\begin{table*}[ht]
  \begin{center}
  \caption{Accuracy of various sentiment analysis classifiers trained on reviews from \citet{zaidan2007using} as noise is injected on \emph{rationales/non-rationales} identified via Allen NLP Saliency interpreter.
  \label{tab:zaidan_saliency_noise}}
  \begin{tabular}{ l c c c c c c c c c c c}
    \toprule
    Dataset & \multicolumn{11}{c}{Percent rationales tokens replaced by noise} \\
    \midrule
    \multicolumn{12}{c}{SVM}\\
    \midrule
     & $0$ & $10$ & $20$ & $30$ & $40$ & $50$ & $60$ & $70$ & $80$ & $90$ & $100$\\
    In-sample test & $87.5$ & $85$ & $84.5$ & $84$ & $82.5$ & $83$ & $81$ & $80$ & $77.5$ & $75.5$ & $75.5$\\
    CRD & $46.1$ & $51$ & $50.6$ & $52$ & $51.8$ & $52.3$ & $52.3$ & $51.8$ & $50.2$ & $49.8$ & $49.8$\\
    Amazon & $68.6$ & $68.1$ & $67.1$ & $66.8$ & $66.9$ & $66.5$ & $66.2$ & $65.4$ & $66.1$ & $66.6$ & $65.7$\\
    Semeval & $56.7$ & $56.6$ & $56.3$ & $56.4$ & $56.2$ & $56.4$ & $56.4$ & $56.2$ & $56.8$ & $56.4$ & $56.4$\\
    Yelp & $76.2$ & $76.1$ & $76$ & $76.2$ & $76.4$ & $76.5$ & $76.9$ & $76.9$ & $76.7$ & $76.9$ & $76.5$\\
    \midrule
    \multicolumn{12}{c}{BiLSTM with Self Attention}\\
    \midrule
    In-sample test & $80.3$ & $83.2$ & $78.1$ & $76.9$ & $73.6$ & $80.3$ & $81.7$ & $76.4$ & $76.4$ & $74$ & $74.5$\\
    CRD & $49.2$ & $49.8$ & $50.6$ & $50.8$ & $50.8$ & $49.2$ & $49.2$ & $49.2$ & $52$ & $49.4$ & $49.8$\\
    Amazon & $50$ & $49.8$ & $50.5$ & $49.8$ & $50$ & $49.7$ & $49.7$ & $50.1$ & $50$ & $50.3$ & $49.8$\\
    Semeval & $50$ & $50$ & $50$ & $50$ & $50$ & $50$ & $50$ & $50$ & $50$ & $50$ & $50$\\
    Yelp & $50.5$ & $50.4$ & $50$ & $50.5$ & $50.8$ & $50.3$ & $50.1$ & $50.9$ & $50.7$ & $50.8$ & $50.5$\\
    \midrule
    \multicolumn{12}{c}{Longformer}\\
    \midrule
    In-sample test & $97.5$ & $98$ & $98$ & $97.5$ & $97.5$ & $97$ & $97$ & $97$ & $96.5$ & $94.3$ & $92.8$\\
    CRD & $93.4$ & $93.4$ & $93.9$ & $94$ & $92.4$ & $91$ & $92.2$ & $91.7$ & $90.2$ & $86.6$ & $74.5$\\
    Amazon & $81.8$ & $81$ & $74.2$ & $66.3$ & $74.7$ & $78.3$ & $80.6$ & $76.2$ & $63.2$ & $77.3$ & $55.8$\\
    Semeval & $80.3$ & $79.9$ & $69.4$ & $64$ & $73.4$ & $77$ & $78$ & $74.6$ & $60.6$ & $78.3$ & $56.4$\\
    Yelp & $88.6$ & $87.3$ & $84.5$ & $76.6$ & $83.1$ & $86.4$ & $87.6$ & $85.6$ & $72.8$ & $84.1$ & $61.7$\\
    \midrule
    Dataset & \multicolumn{11}{c}{Percent noise in non-rationales} \\
    \midrule
    \multicolumn{12}{c}{SVM}\\
    \midrule
    In-sample test & $87.5$ & $79.7$ & $79.9$ & $79.5$ & $81.1$ & $79.9$ & $80.3$ & $78.9$ & $78.7$ & $79.3$ & $73.4$\\
    CRD & $46.1$ & $52.7$ & $52$ & $53.1$ & $50$ & $54.3$ & $50.6$ & $54.3$ & $52$ & $57.2$ & $57.2$\\
    Amazon & $68.6$ & $68.1$ & $66.2$ & $67$ & $65.8$ & $68.8$ & $65.3$ & $64.4$ & $65.3$ & $63.1$ & $61.9$\\
    Semeval & $56.7$ & $57.4$ & $56.2$ & $56.9$ & $55.9$ & $57.3$ & $55.6$ & $58.1$ & $57$ & $57.9$ & $58$\\
    Yelp & $76.2$ & $76.4$ & $75.4$ & $76$ & $75.6$ & $75.8$ & $74.2$ & $74.3$ & $73.6$ & $73.7$ & $71.5$\\
    \midrule
    \multicolumn{12}{c}{BiLSTM with Self Attention}\\
    \midrule
    In-sample test & $80.3$ & $76.9$ & $80.8$ & $79.3$ & $78.8$ & $77.9$ & $76$ & $76$ & $63.5$ & $73.6$ & $66.8$\\
    CRD & $49.2$ & $50$ & $50.2$ & $50.4$ & $50.2$ & $50.8$ & $51.4$ & $47.9$ & $48.8$ & $47.5$ & $48.4$\\
    Amazon & $50$ & $50$ & $49.7$ & $50.1$ & $50.2$ & $50.8$ & $50.2$ & $50$ & $50.3$ & $49.8$ & $47.9$\\
    Semeval & $50$ & $50$ & $50$ & $50$ & $50$ & $50.1$ & $50$ & $50$ & $50$ & $50$ & $50$\\
    Yelp & $50.5$ & $50.5$ & $50.3$ & $51.3$ & $54.5$ & $54.9$ & $52.2$ & $51.4$ & $52.7$ & $50.5$ & $54.9$\\
    \midrule
    \multicolumn{12}{c}{Longformer}\\
    \midrule
    In-sample test & $97.5$ & $97.8$ & $98$ & $97.8$ & $97.5$ & $98.3$ & $95$ & $92.8$ & $84.5$ & $83.5$ & $74.5$\\
    CRD & $93.4$ & $94.4$ & $94.1$ & $93.6$ & $93.1$ & $93.3$ & $92.8$ & $91$ & $86.9$ & $70.9$ & $67.6$\\
    Amazon & $81.8$ & $80.9$ & $75.9$ & $75.8$ & $79.7$ & $68.9$ & $81.4$ & $72.4$ & $71.2$ & $63.5$ & $55.2$\\
    Semeval & $80.3$ & $78.6$ & $72.7$ & $74.4$ & $79.1$ & $68.9$ & $81.6$ & $73.5$ & $76.2$ & $59.2$ & $55.7$\\
    Yelp & $88.6$ & $88.1$ & $84.1$ & $84.8$ & $87.5$ & $81.3$ & $89.3$ & $82.2$ & $82.2$ & $70.3$ & $61.5$\\
    \bottomrule
  \end{tabular}
  \end{center}
\end{table*}

\begin{figure*}[t!]
    \begin{subfigure}[h]{\textwidth}
         \centering
         \includegraphics[width=0.78\linewidth]{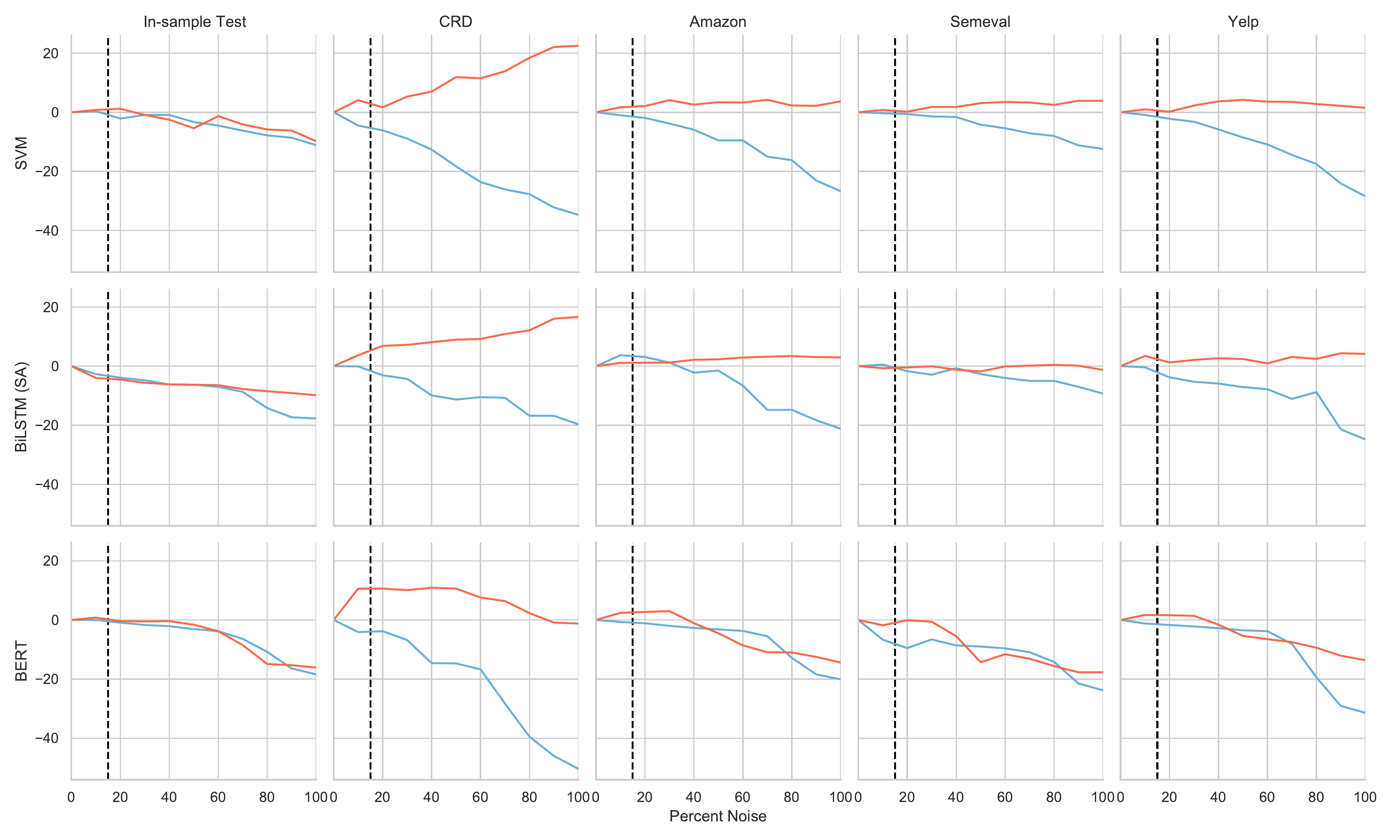}
          \caption{Noising spans marked by humans}
          \label{fig:iclr_human_appendix}
    \end{subfigure}\\
    \begin{subfigure}[h]{\textwidth}
         \centering
         \includegraphics[width=0.78\linewidth]{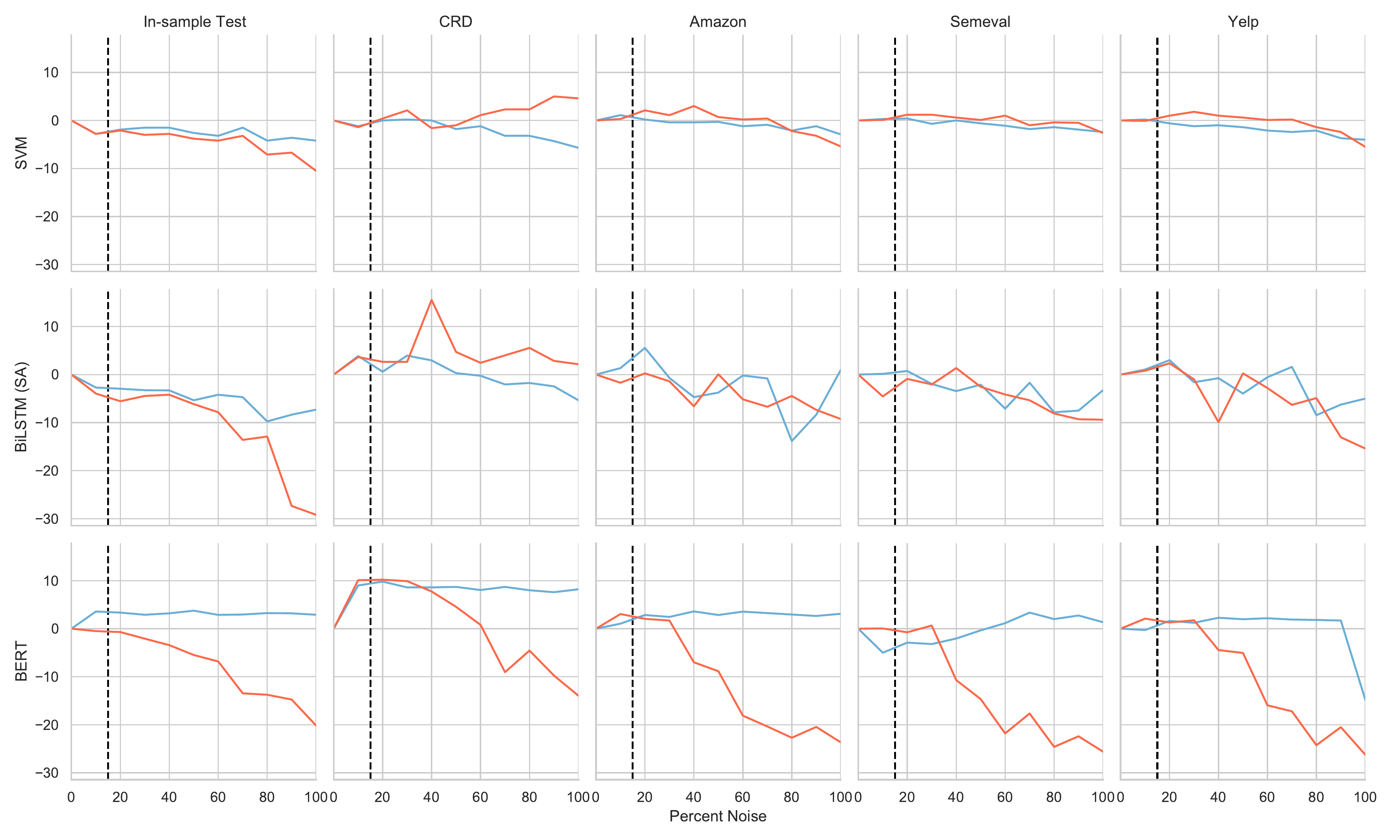}
          \caption{Noising spans marked by Attention}
          \label{fig:iclr_attention_appendix}
    \end{subfigure}\\
    \begin{subfigure}[h]{\textwidth}
         \centering
         \includegraphics[width=0.78\linewidth]{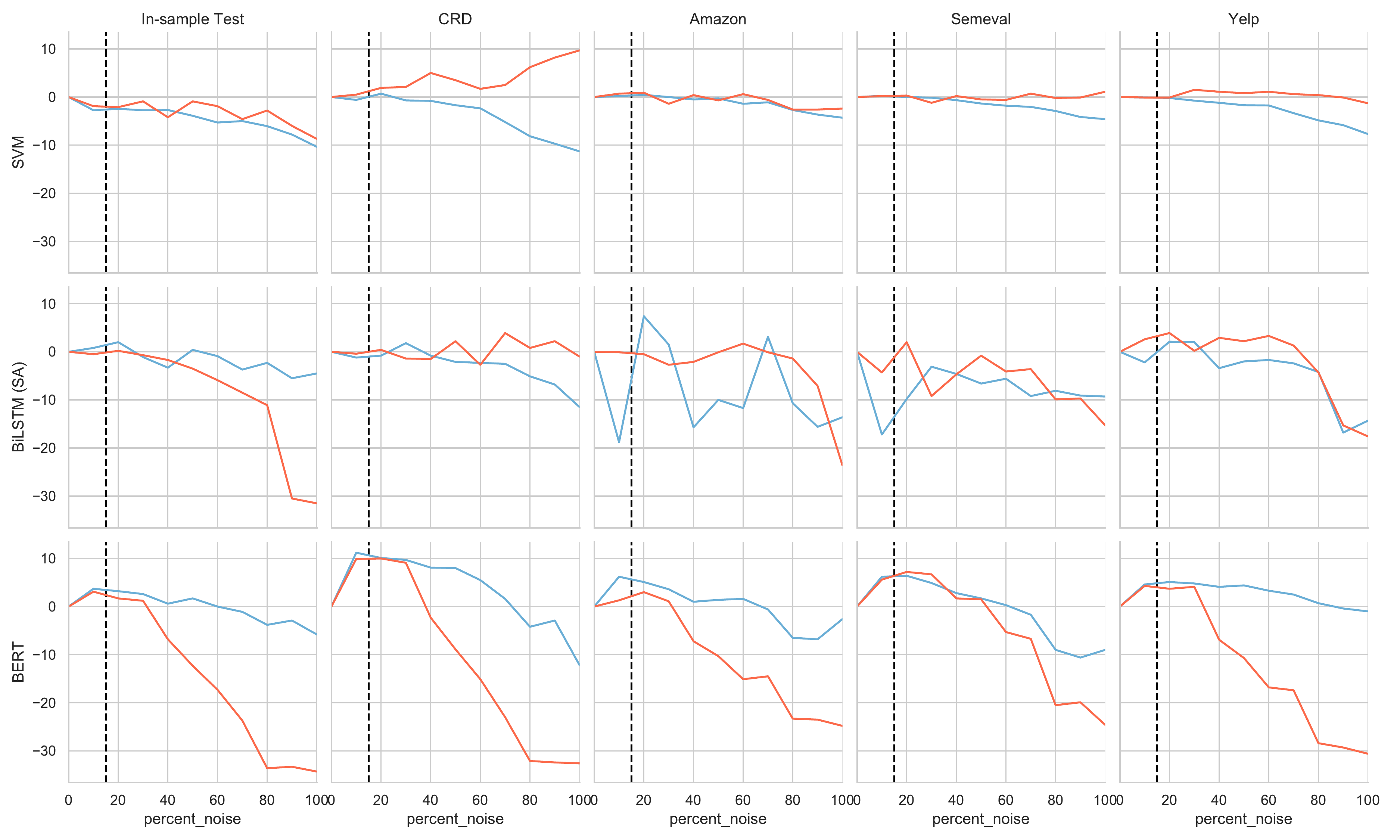}
          \caption{Noising spans marked by AllenNLP Saliency Interpreter}
          \label{fig:iclr_saliency_appendix}
          \vspace{-1mm}
    \end{subfigure}
\caption{
Change in classifier accuracy as noise is injected on \emph{rationales} (in \textcolor{blue}{blue}) or \emph{non-rationales} (in \textcolor{red}{red}) for IMDb reviews from \citet{kaushik2020learning}. The vertical dashed line indicates the fraction of median length of \emph{non-rationales} equal to the median length of \emph{rationales}.
\label{fig:iclr_noise_appendix}}
\end{figure*}

\newpage

\begin{figure*}[t!]
    \begin{subfigure}[h]{\textwidth}
         \centering
         \includegraphics[width=0.78\linewidth]{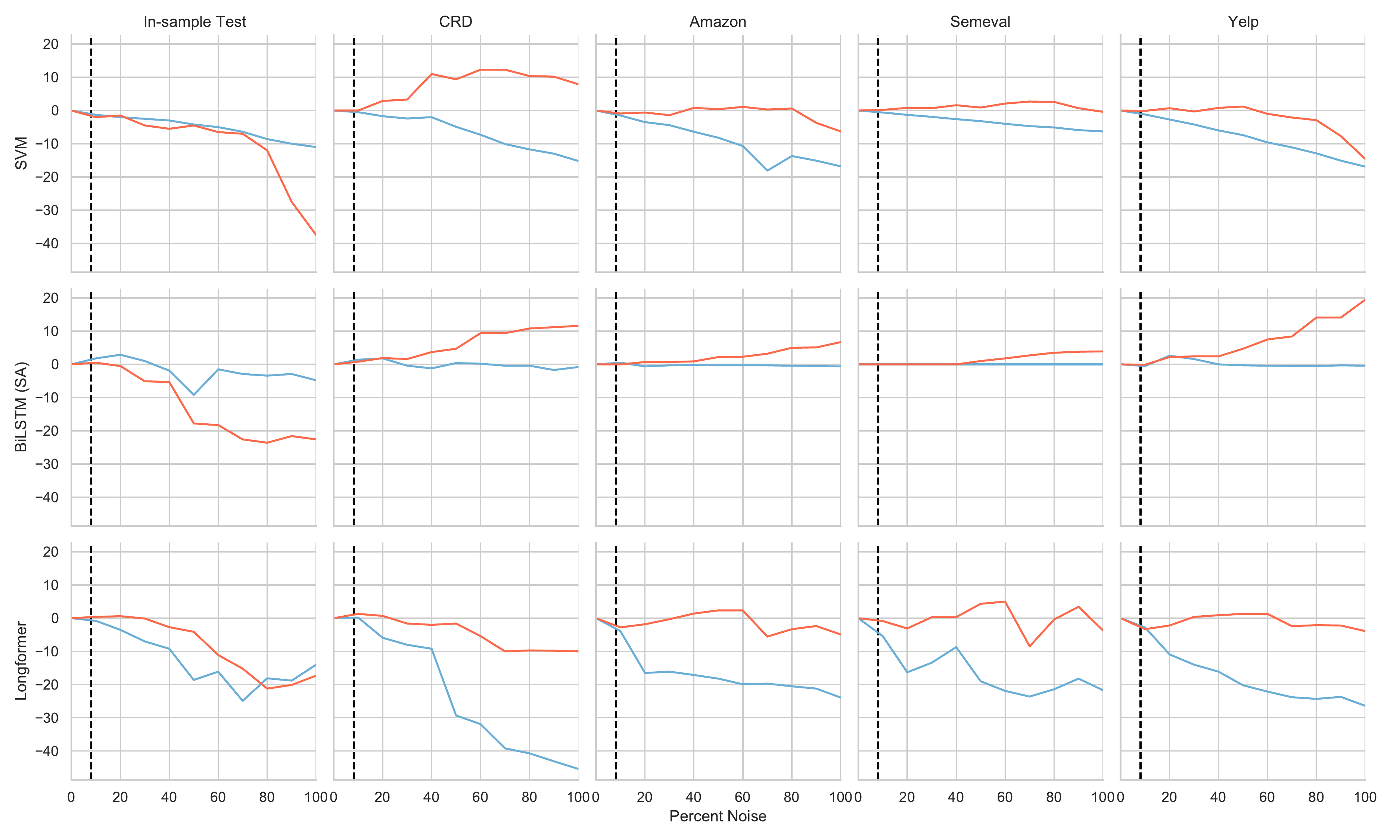}
          \caption{Noising spans marked by humans}
          \label{fig:zaidan_human_appendix}
    \end{subfigure}\\
    \begin{subfigure}[h]{\textwidth}
         \centering
         \includegraphics[width=0.78\linewidth]{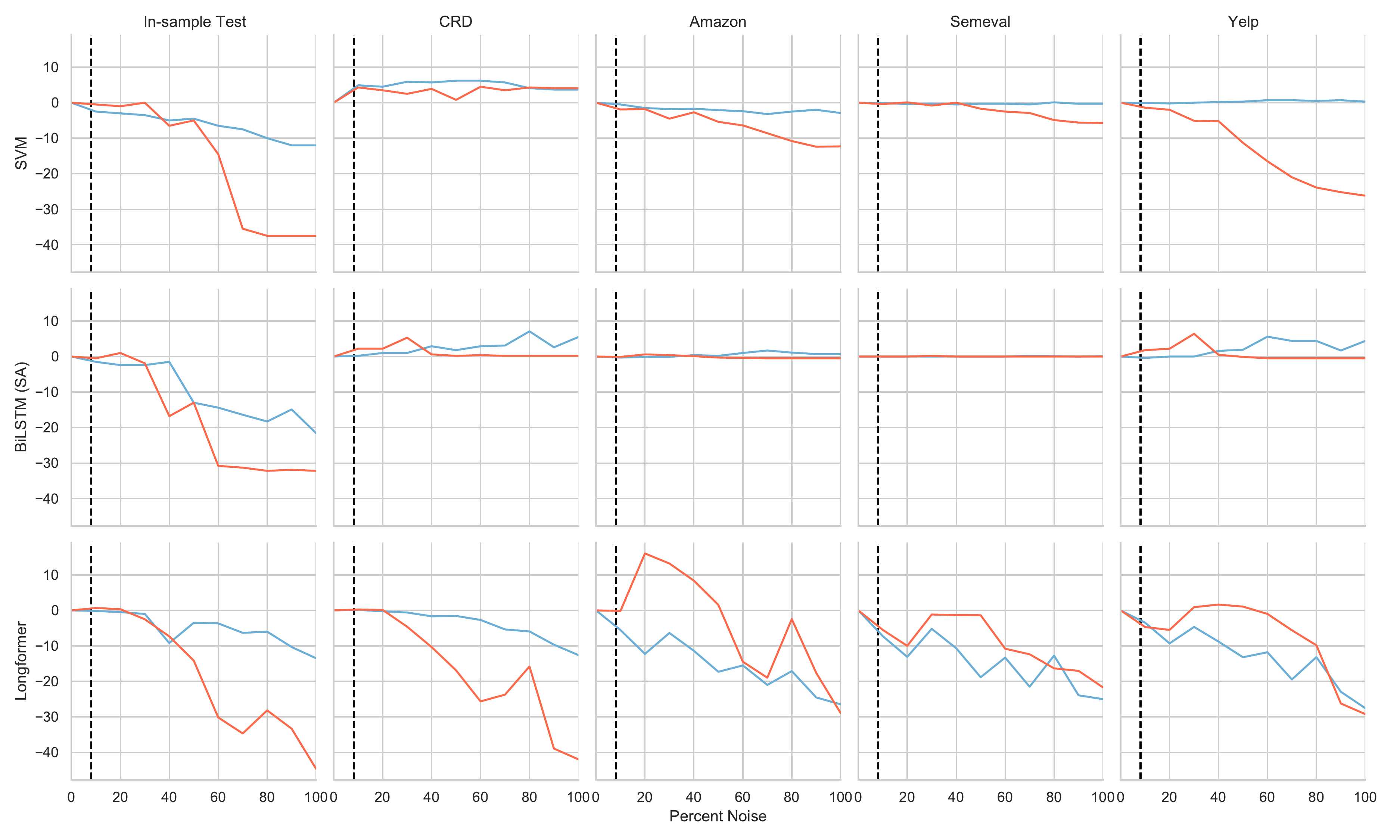}
          \caption{Noising spans marked by Attention}
          \label{fig:zaidan_attention_appendix}
    \end{subfigure}\\
    \begin{subfigure}[h]{\textwidth}
         \centering
         \includegraphics[width=0.78\linewidth]{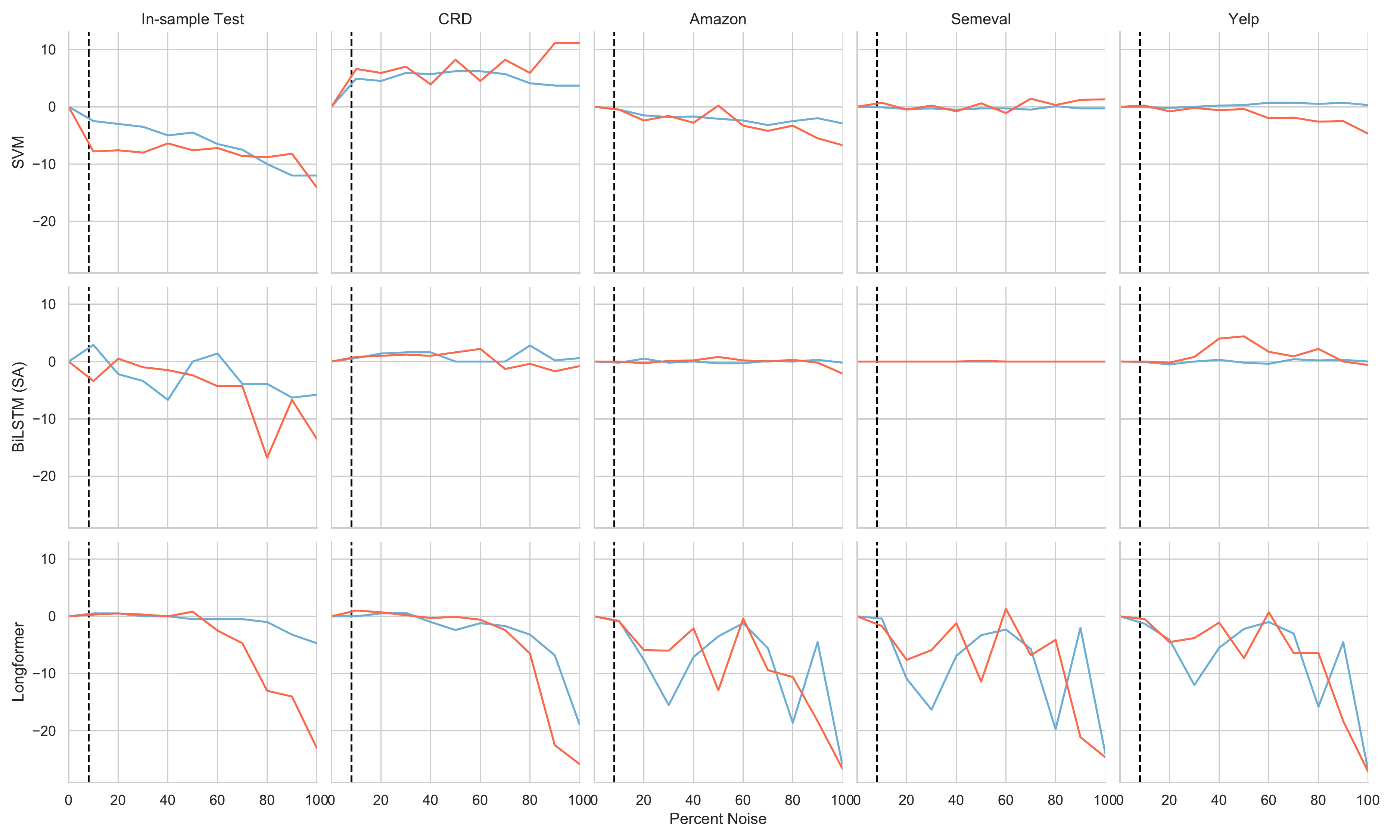}
          \caption{Noising spans marked by AllenNLP Saliency Interpreter}
          \label{fig:zaidan_saliency_appendix}
          \vspace{-1mm}
    \end{subfigure}
\caption{
Change in classifier accuracy as noise is injected on \emph{rationales} (in \textcolor{blue}{blue}) or \emph{non-rationales} (in \textcolor{red}{red}) for IMDb reviews from \citet{zaidan2007using}. The vertical dashed line indicates the fraction of median length of \emph{non-rationales} equal to the median length of \emph{rationales}.
\label{fig:zaidan_noise_appendix}}
\end{figure*}
\clearpage

\begin{figure*}[t!]
    \begin{subfigure}[b]{\textwidth}
         \centering
         \includegraphics[width=0.51\linewidth]{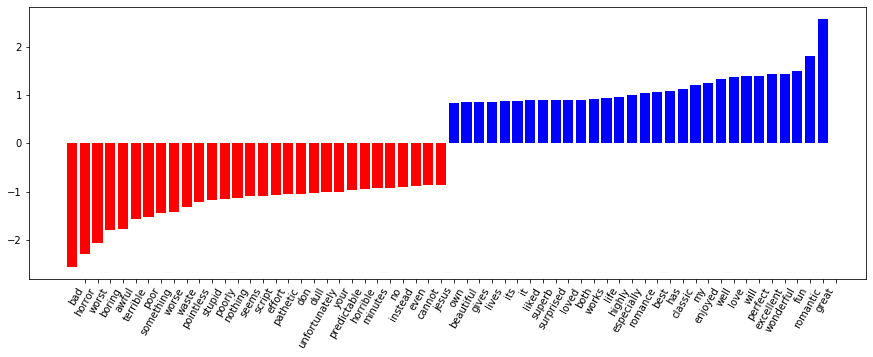}
         \caption{Trained on the original $1.7k$ IMDb reviews in \citet{kaushik2020learning}}
     \end{subfigure}
    \begin{subfigure}[b]{.49\textwidth}
         \centering
         \includegraphics[width=\linewidth]{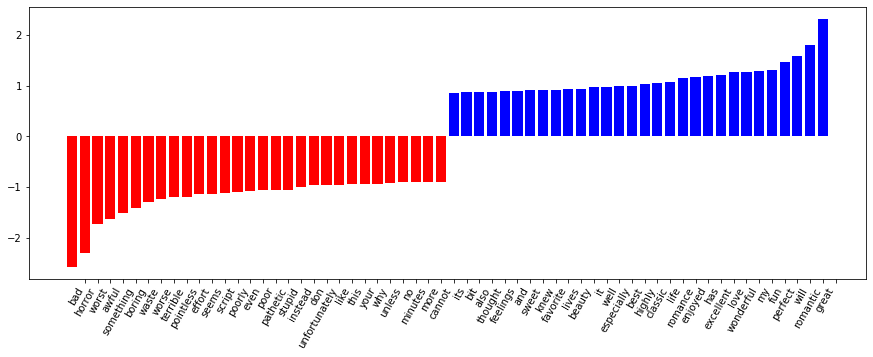}
         \caption{20\% noise in rationales}
     \end{subfigure}
      \begin{subfigure}[b]{.49\textwidth}
          \centering
          \includegraphics[width=\linewidth]{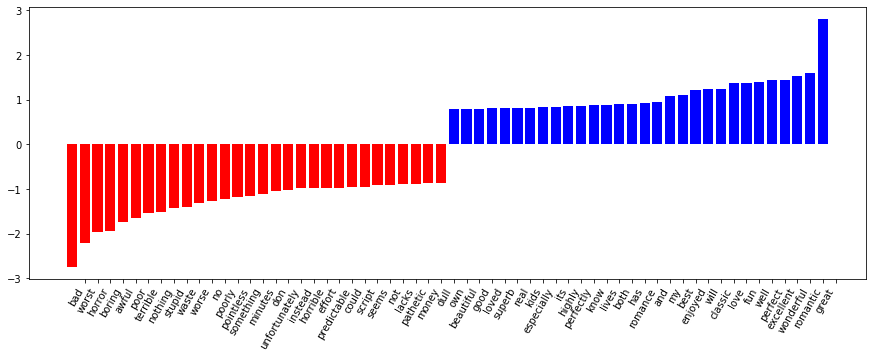}
          \caption{20\% noise in non-rationales}
      \end{subfigure} \\
     \begin{subfigure}[b]{.49\textwidth}
         \centering
         \includegraphics[width=\textwidth]{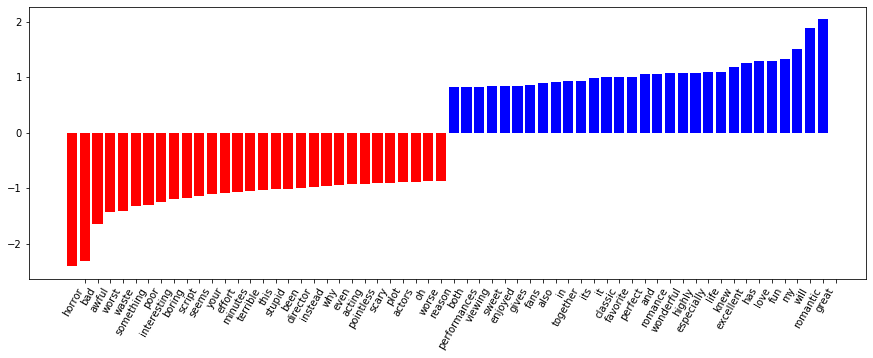}
         \caption{40\% noise in rationales}
     \end{subfigure}
     \begin{subfigure}[b]{.49\textwidth}
         \centering
         \includegraphics[width=\textwidth]{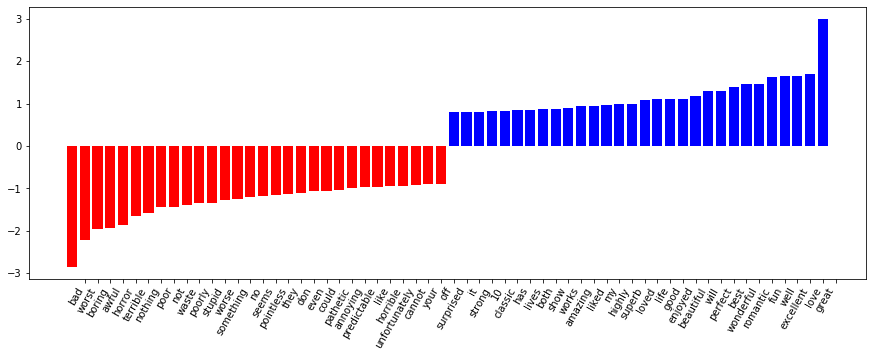}
         \caption{40\% noise in non-rationales}
     \end{subfigure}\\
     \begin{subfigure}[b]{.49\textwidth}
         \centering
         \includegraphics[width=\textwidth]{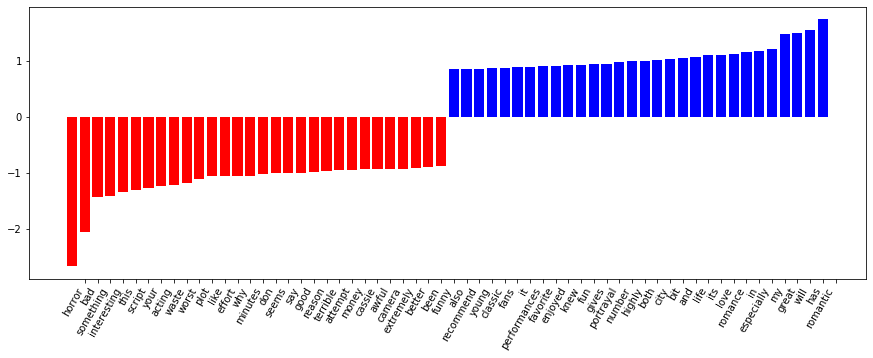}
         \caption{60\% noise in rationales}
     \end{subfigure}
     \begin{subfigure}[b]{.49\textwidth}
         \centering
         \includegraphics[width=\textwidth]{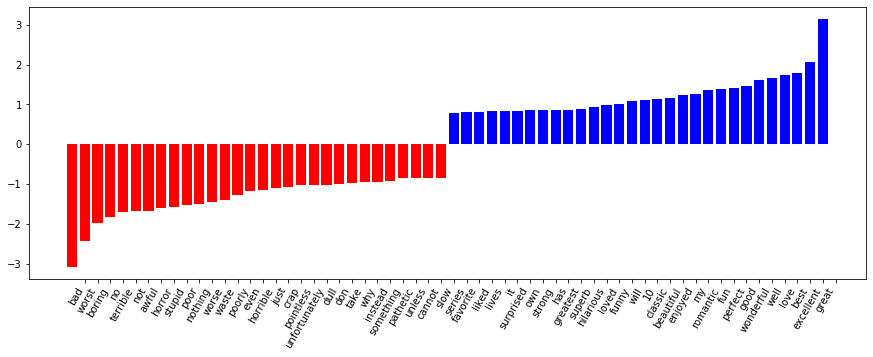}
         \caption{60\% noise in non-rationales}
     \end{subfigure}\\
     \begin{subfigure}[b]{.49\textwidth}
         \centering
         \includegraphics[width=\textwidth]{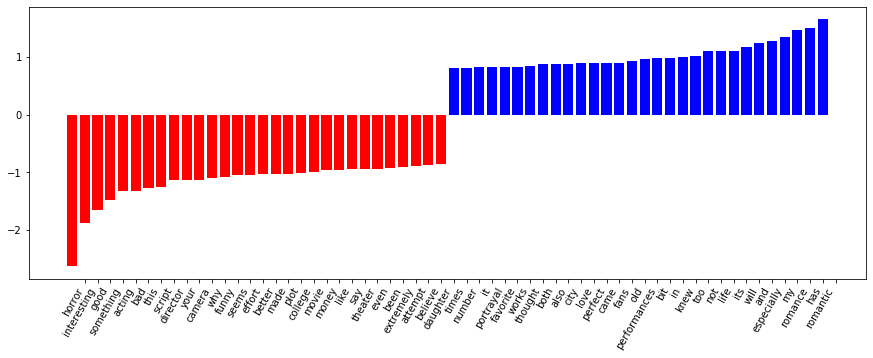}
         \caption{80\% noise in rationales}
     \end{subfigure}
     \begin{subfigure}[b]{.49\textwidth}
         \centering
         \includegraphics[width=\textwidth]{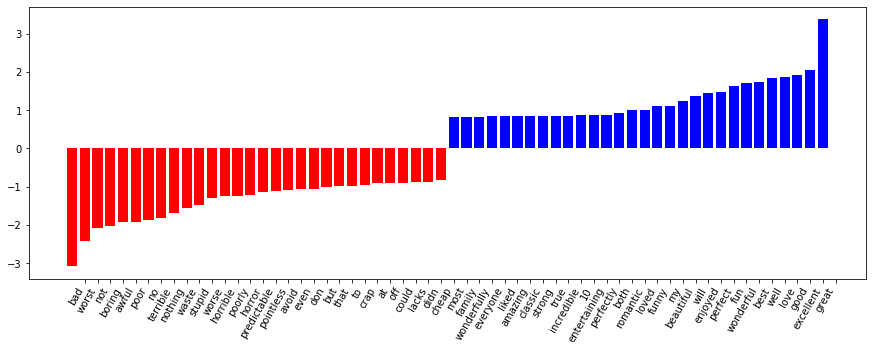}
         \caption{80\% noise in non-rationales}
     \end{subfigure}\\
     \begin{subfigure}[b]{.49\textwidth}
         \centering
         \includegraphics[width=\textwidth]{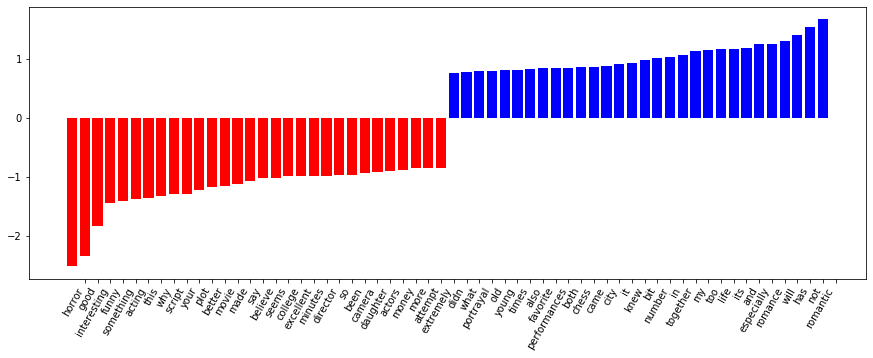}
         \caption{100\% noise in rationales}
     \end{subfigure}
     \begin{subfigure}[b]{.49\textwidth}
         \centering
         \includegraphics[width=\textwidth]{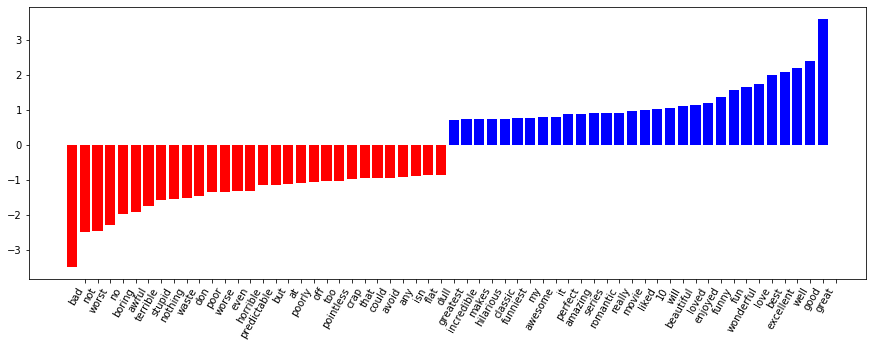}
         \caption{100\% noise in non-rationales}
     \end{subfigure}\\
\caption{
Most important features learned by an SVM classifier trained on TF-IDF bag of words. Rationales are identified by humans.
\label{iclr_human}}
\end{figure*}

\begin{figure*}[t!]
    \begin{subfigure}[b]{\textwidth}
         \centering
         \includegraphics[width=0.51\linewidth]{figures/ICLR/orig.png}
         \caption{Trained on the original $1.7k$ IMDb reviews in \citet{kaushik2020learning}}
     \end{subfigure}
    \begin{subfigure}[b]{.49\textwidth}
         \centering
         \includegraphics[width=\linewidth]{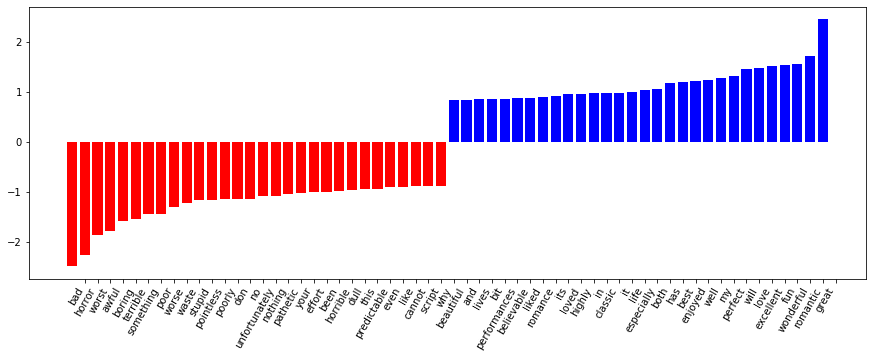}
         \caption{20\% noise in rationales}
     \end{subfigure}
      \begin{subfigure}[b]{.49\textwidth}
          \centering
          \includegraphics[width=\linewidth]{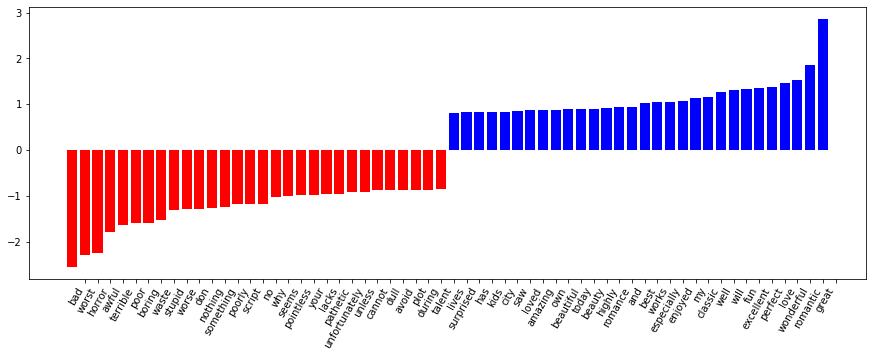}
          \caption{20\% noise in non-rationales}
      \end{subfigure} \\
     \begin{subfigure}[b]{.49\textwidth}
         \centering
         \includegraphics[width=\textwidth]{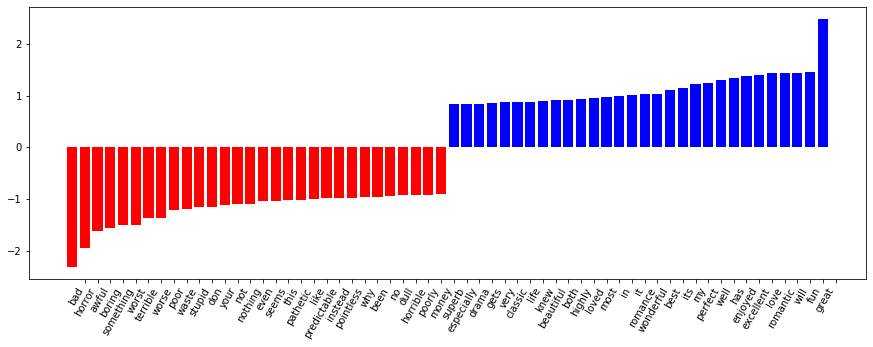}
         \caption{40\% noise in rationales}
     \end{subfigure}
     \begin{subfigure}[b]{.49\textwidth}
         \centering
         \includegraphics[width=\textwidth]{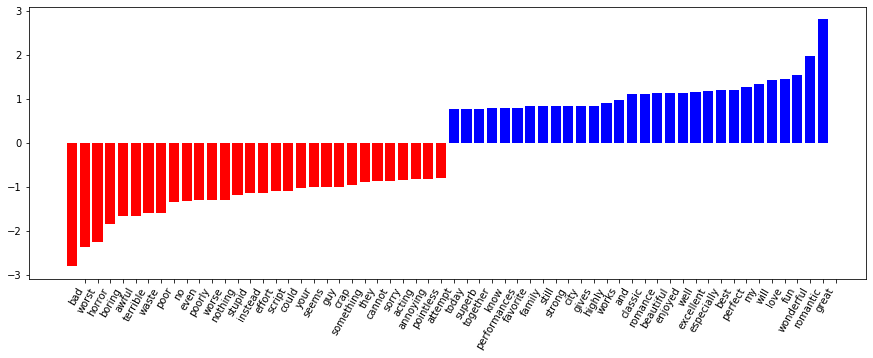}
         \caption{40\% noise in non-rationales}
     \end{subfigure}\\
     \begin{subfigure}[b]{.49\textwidth}
         \centering
         \includegraphics[width=\textwidth]{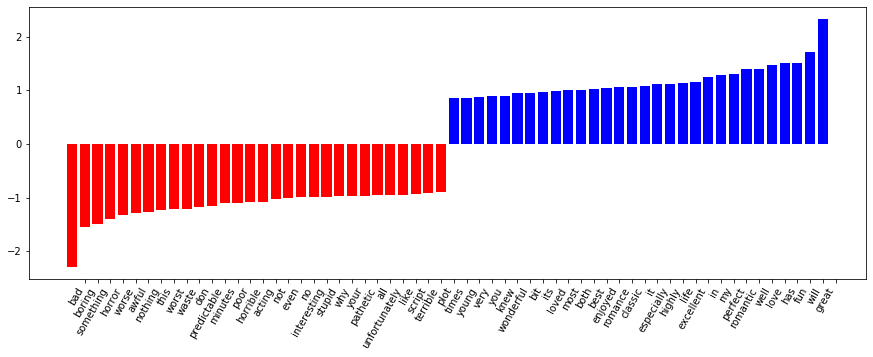}
         \caption{60\% noise in rationales}
         
     \end{subfigure}
     \begin{subfigure}[b]{.49\textwidth}
         \centering
         \includegraphics[width=\textwidth]{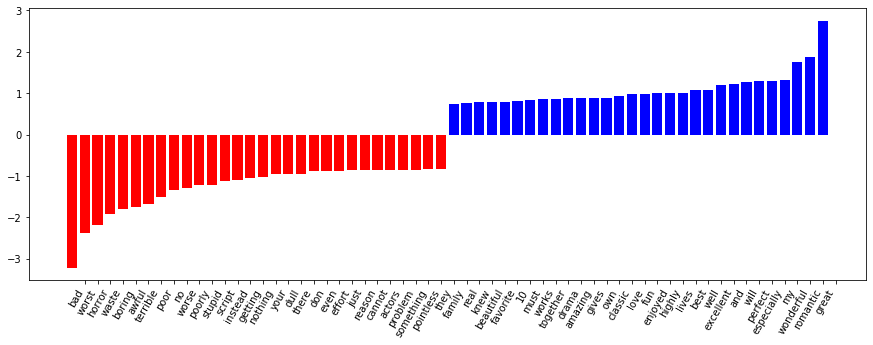}
         \caption{60\% noise in non-rationales}
         
     \end{subfigure}\\
     \begin{subfigure}[b]{.49\textwidth}
         \centering
         \includegraphics[width=\textwidth]{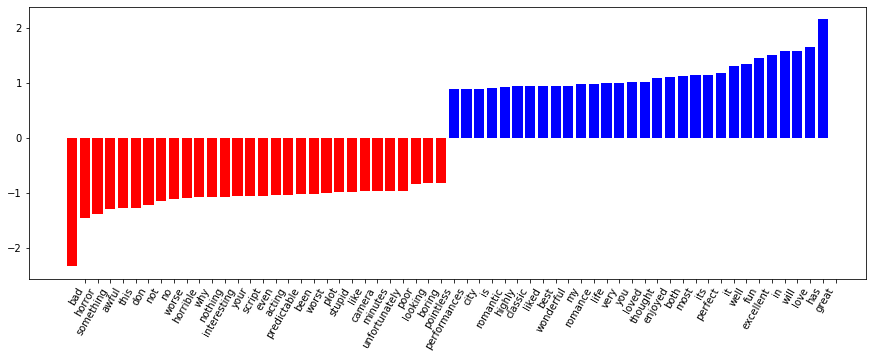}
         \caption{80\% noise in rationales}
         
     \end{subfigure}
     \begin{subfigure}[b]{.49\textwidth}
         \centering
         \includegraphics[width=\textwidth]{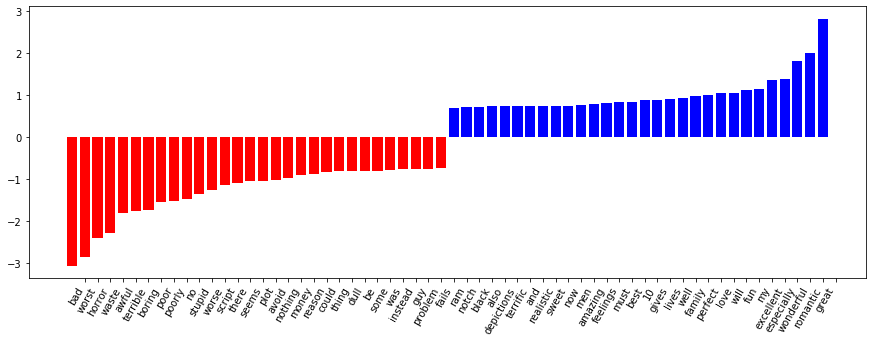}
         \caption{80\% noise in non-rationales}
         
     \end{subfigure}\\
     \begin{subfigure}[b]{.49\textwidth}
         \centering
         \includegraphics[width=\textwidth]{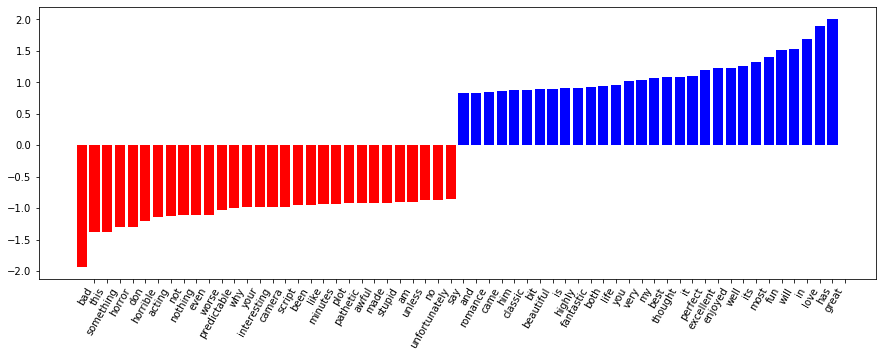}
         \caption{100\% noise in rationales}
         
     \end{subfigure}
     \begin{subfigure}[b]{.49\textwidth}
         \centering
         \includegraphics[width=\textwidth]{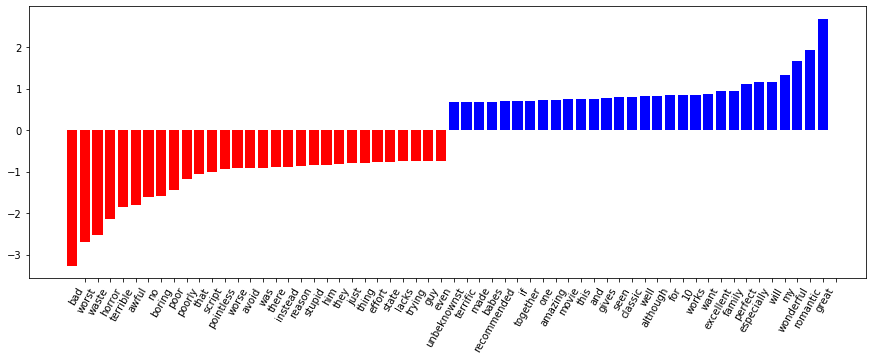}
         \caption{100\% noise in non-rationales}
         
     \end{subfigure}\\
\caption{
Most important features learned by an SVM classifier trained on TF-IDF bag of words. Rationales are identified as tokens attended upon by a BiLSTM with Self Attention model.
\label{iclr_attention}}
\end{figure*}

\begin{figure*}[t!]
    \begin{subfigure}[b]{\textwidth}
         \centering
         \includegraphics[width=0.51\linewidth]{figures/ICLR/orig.png}
         \caption{Trained on the original $1.7k$ IMDb reviews in \citet{kaushik2020learning}}
     \end{subfigure}
    \begin{subfigure}[b]{.49\textwidth}
         \centering
         \includegraphics[width=\linewidth]{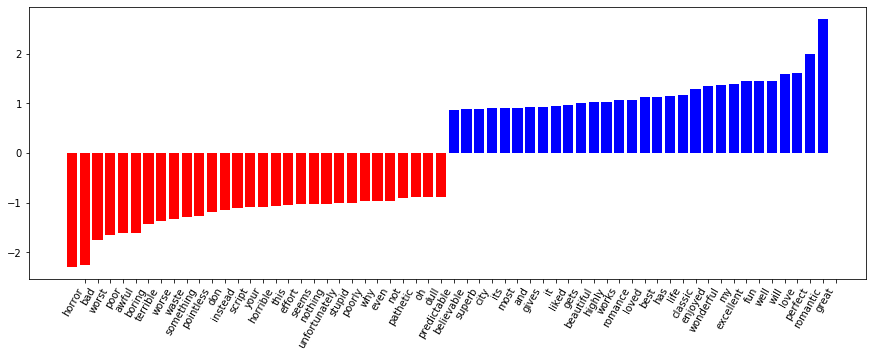}
         \caption{20\% noise in rationales}
     \end{subfigure}
      \begin{subfigure}[b]{.49\textwidth}
          \centering
          \includegraphics[width=\linewidth]{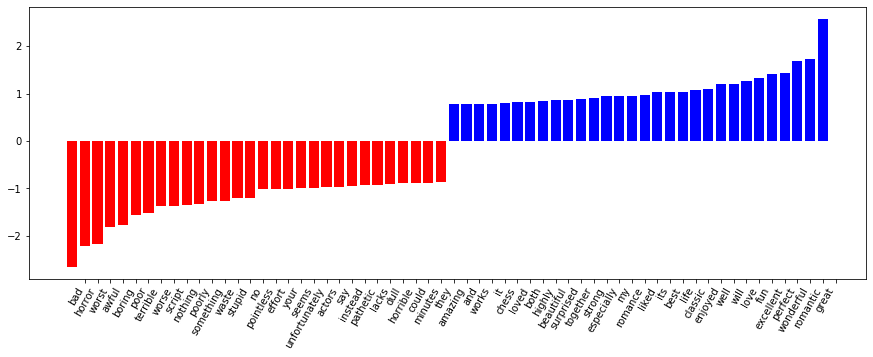}
          \caption{20\% noise in non-rationales}
      \end{subfigure} \\
     \begin{subfigure}[b]{.49\textwidth}
         \centering
         \includegraphics[width=\textwidth]{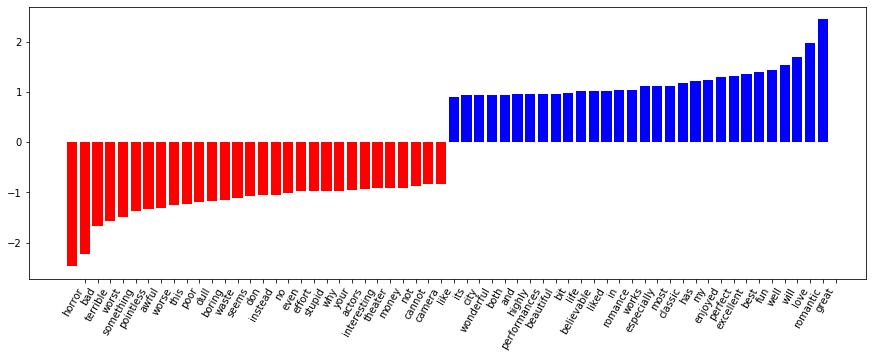}
         \caption{40\% noise in rationales}
     \end{subfigure}
     \begin{subfigure}[b]{.49\textwidth}
         \centering
         \includegraphics[width=\textwidth]{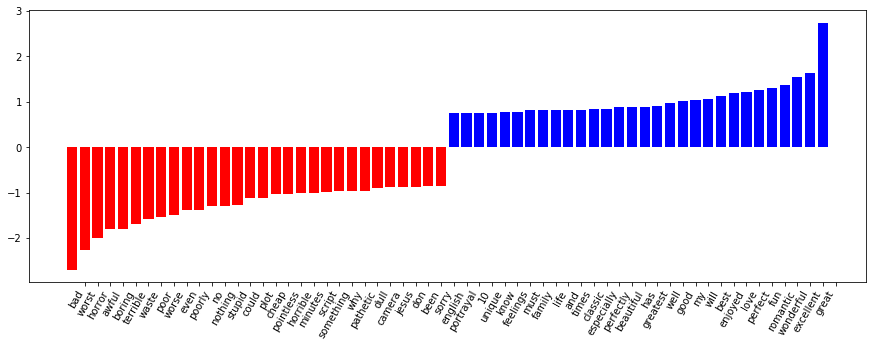}
         \caption{40\% noise in non-rationales}
     \end{subfigure}\\
     \begin{subfigure}[b]{.49\textwidth}
         \centering
         \includegraphics[width=\textwidth]{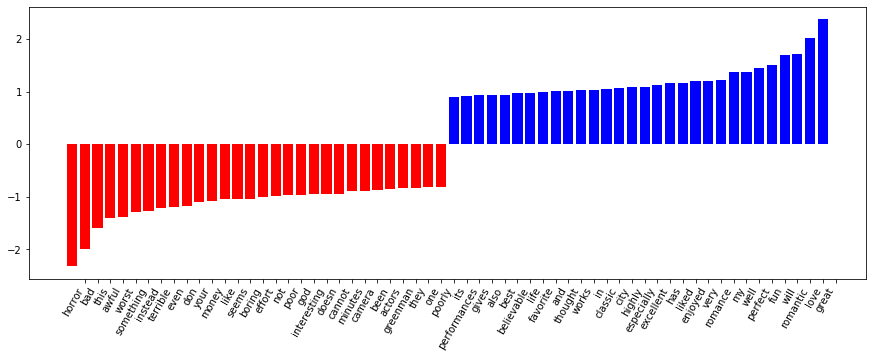}
         \caption{60\% noise in rationales}
         
     \end{subfigure}
     \begin{subfigure}[b]{.49\textwidth}
         \centering
         \includegraphics[width=\textwidth]{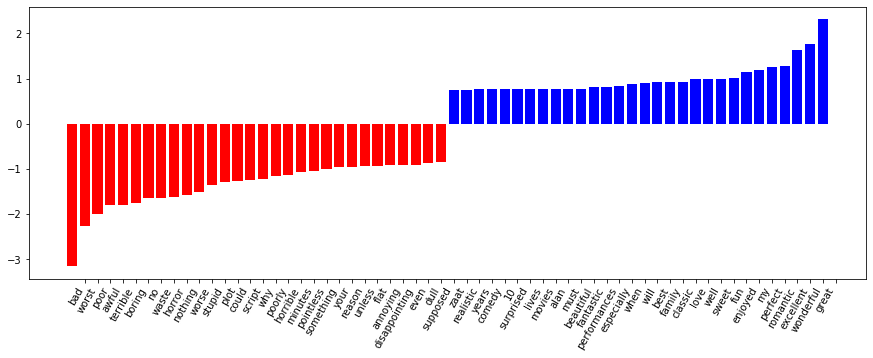}
         \caption{60\% noise in non-rationales}
         
     \end{subfigure}\\
     \begin{subfigure}[b]{.49\textwidth}
         \centering
         \includegraphics[width=\textwidth]{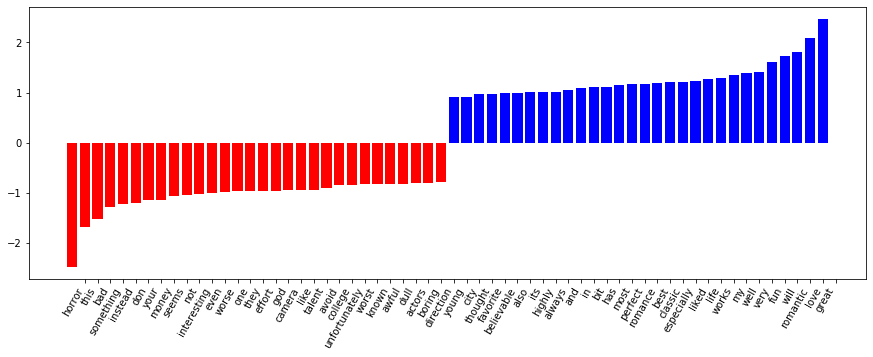}
         \caption{80\% noise in rationales}
         
     \end{subfigure}
     \begin{subfigure}[b]{.49\textwidth}
         \centering
         \includegraphics[width=\textwidth]{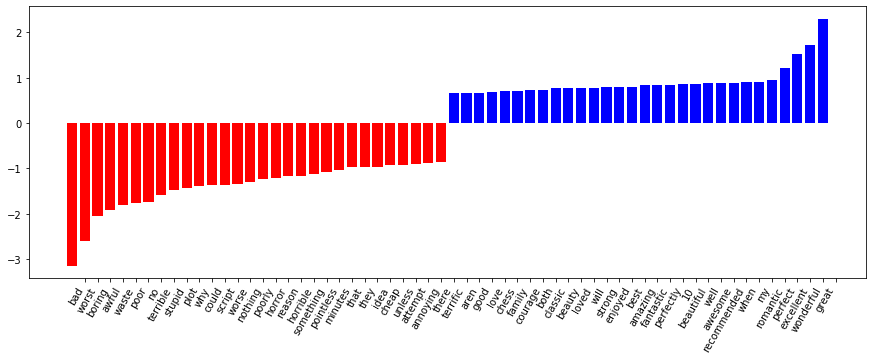}
         \caption{80\% noise in non-rationales}
         
     \end{subfigure}\\
     \begin{subfigure}[b]{.49\textwidth}
         \centering
         \includegraphics[width=\textwidth]{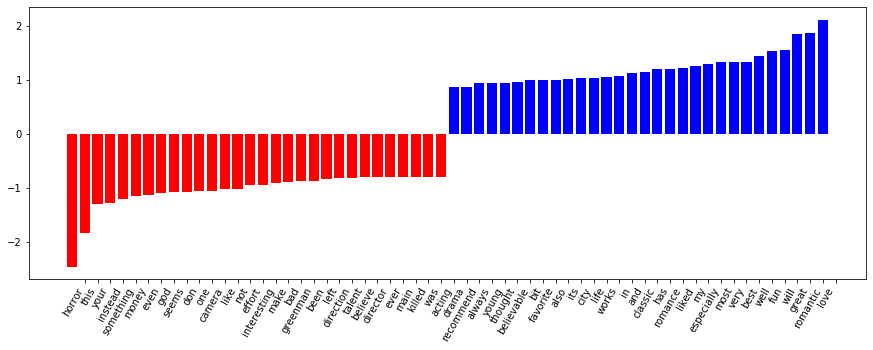}
         \caption{100\% noise in rationales}
         
     \end{subfigure}
     \begin{subfigure}[b]{.49\textwidth}
         \centering
         \includegraphics[width=\textwidth]{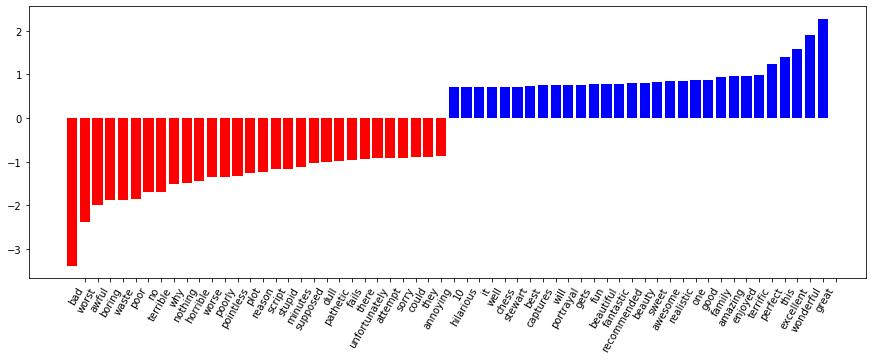}
         \caption{100\% noise in non-rationales}
         
     \end{subfigure}\\
\caption{
Most important features learned by an SVM classifier trained on TF-IDF bag of words. Rationales are identified as tokens marked by the AllenNLP Saliency Interpreter.
\label{iclr_salience}}
\end{figure*}

\begin{figure*}[t!]
    \begin{subfigure}[b]{\textwidth}
         \centering
         \includegraphics[width=0.51\linewidth]{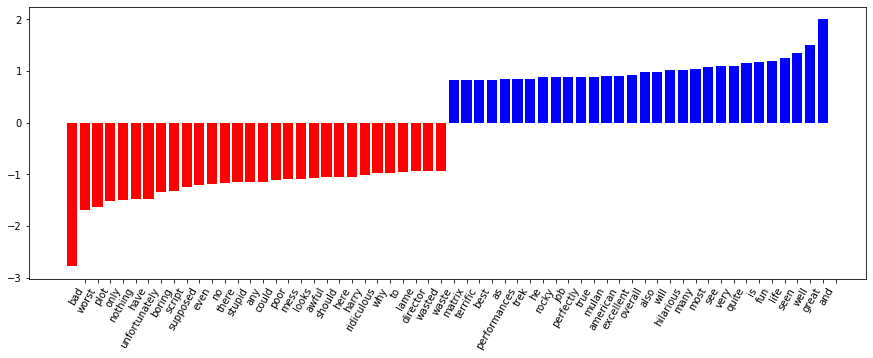}
         \caption{Trained on the original dataset \citep{zaidan2007using}}
         
     \end{subfigure}
    \begin{subfigure}[b]{.49\textwidth}
         \centering
         \includegraphics[width=\linewidth]{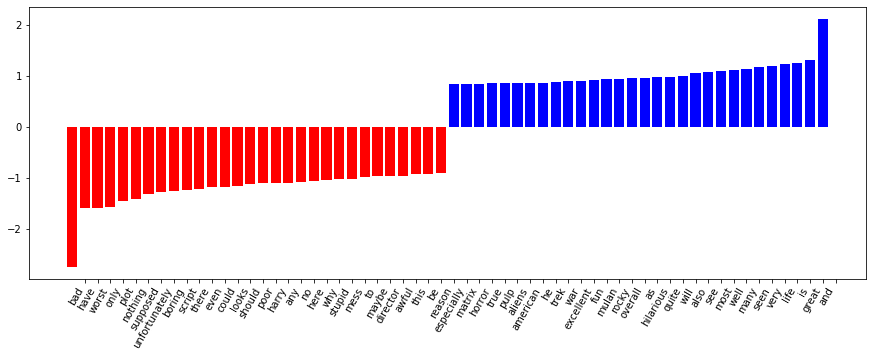}
         \caption{20\% noise in rationales}
         
     \end{subfigure}
      \begin{subfigure}[b]{.49\textwidth}
          \centering
          \includegraphics[width=\linewidth]{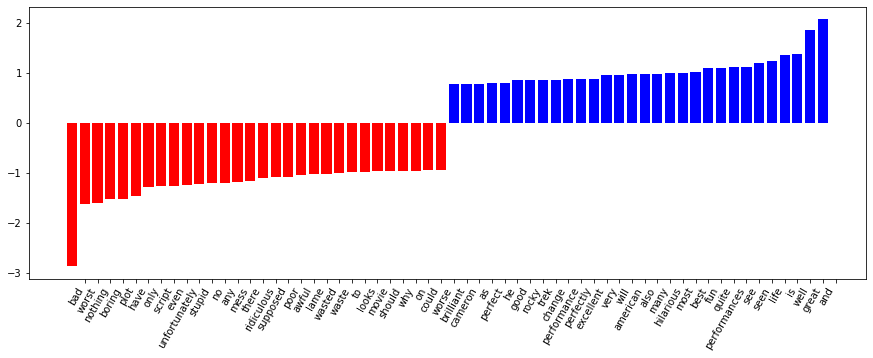}
          \caption{20\% noise in non-rationales}
          
      \end{subfigure} \\
     \begin{subfigure}[b]{.49\textwidth}
         \centering
         \includegraphics[width=\textwidth]{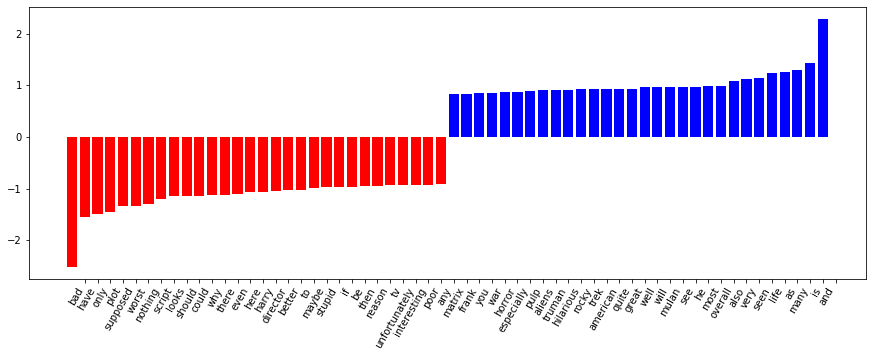}
         \caption{40\% noise in rationales}
         
     \end{subfigure}
     \begin{subfigure}[b]{.49\textwidth}
         \centering
         \includegraphics[width=\textwidth]{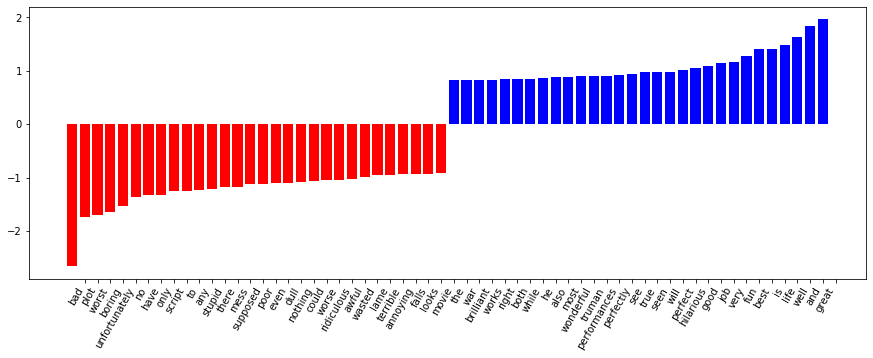}
         \caption{40\% noise in non-rationales}
         
     \end{subfigure}\\
     \begin{subfigure}[b]{.49\textwidth}
         \centering
         \includegraphics[width=\textwidth]{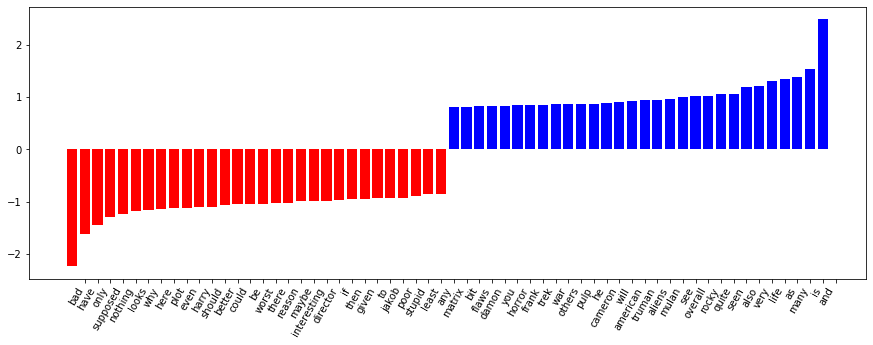}
         \caption{60\% noise in rationales}
         
     \end{subfigure}
     \begin{subfigure}[b]{.49\textwidth}
         \centering
         \includegraphics[width=\textwidth]{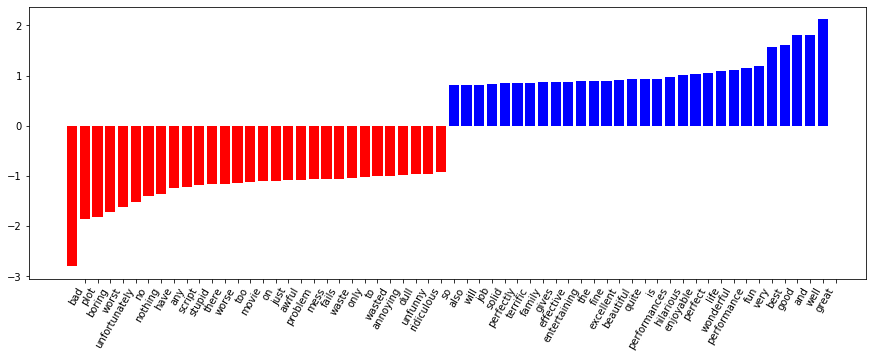}
         \caption{60\% noise in non-rationales}
         
     \end{subfigure}\\
     \begin{subfigure}[b]{.49\textwidth}
         \centering
         \includegraphics[width=\textwidth]{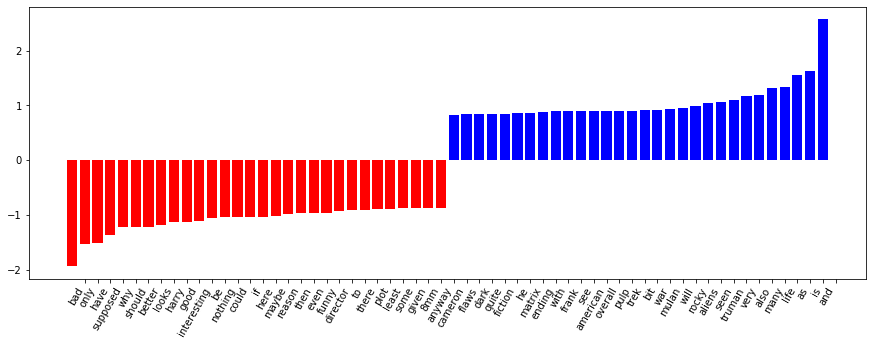}
         \caption{80\% noise in rationales}
         
     \end{subfigure}
     \begin{subfigure}[b]{.49\textwidth}
         \centering
         \includegraphics[width=\textwidth]{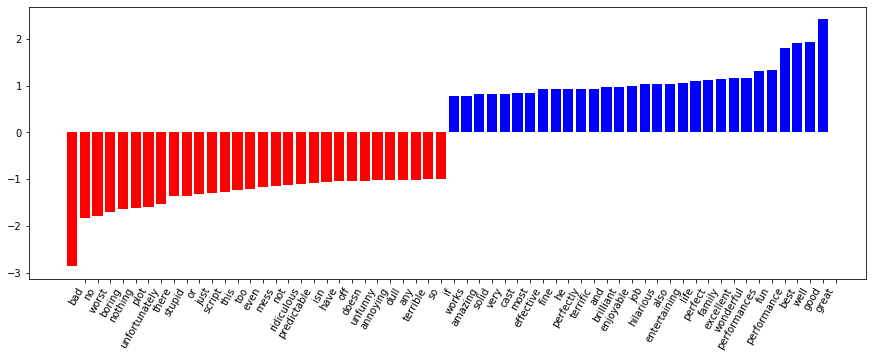}
         \caption{80\% noise in non-rationales}
         
     \end{subfigure}\\
     \begin{subfigure}[b]{.49\textwidth}
         \centering
         \includegraphics[width=\textwidth]{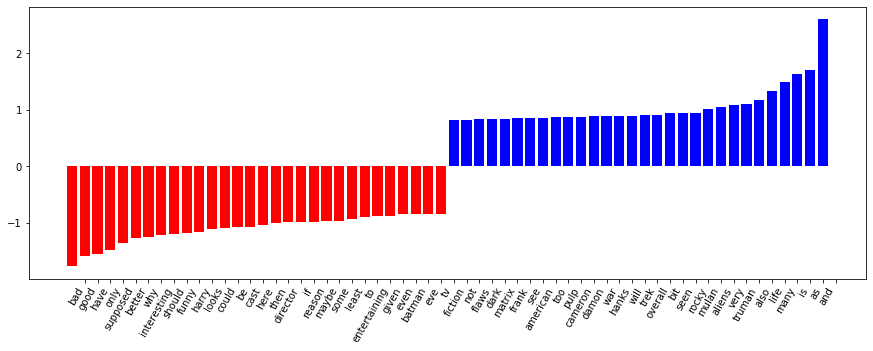}
         \caption{100\% noise in rationales}
         
     \end{subfigure}
     \begin{subfigure}[b]{.49\textwidth}
         \centering
         \includegraphics[width=\textwidth]{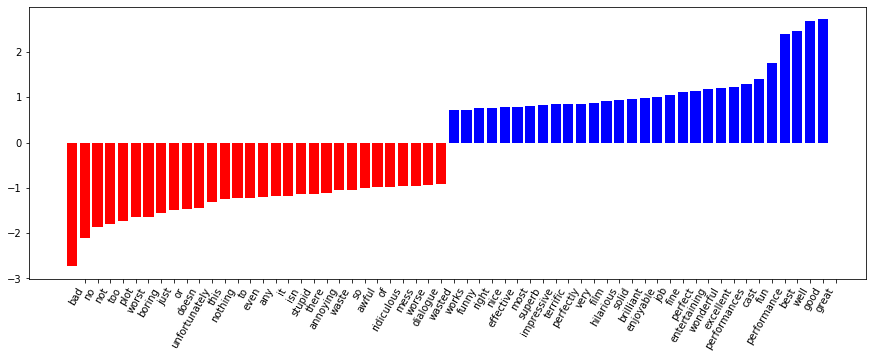}
         \caption{100\% noise in non-rationales}
         
     \end{subfigure}\\
\caption{
Most important features learned by an SVM classifier trained on TF-IDF bag of words. All noise inserted on rationales identified by humans.
\label{zaidan_human}}
\end{figure*}

\begin{figure*}[t!]
    \begin{subfigure}[b]{\textwidth}
         \centering
         \includegraphics[width=0.51\linewidth]{figures/Zaidan/human/orig.png}
         \caption{Trained on the original dataset \citep{zaidan2007using}}
         
     \end{subfigure}
    \begin{subfigure}[b]{.49\textwidth}
         \centering
         \includegraphics[width=\linewidth]{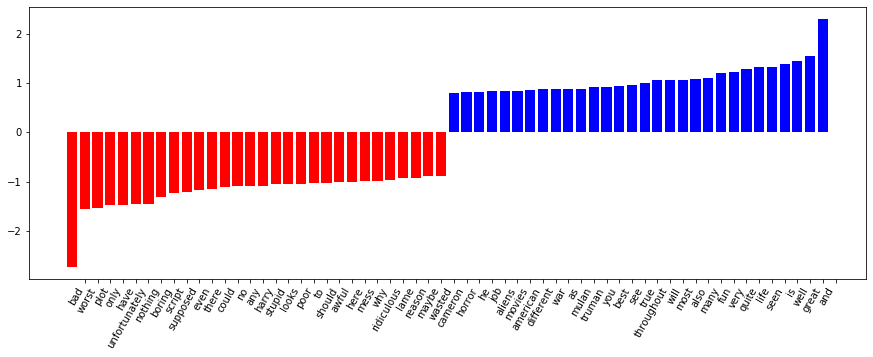}
         \caption{20\% noise in rationales}
         
     \end{subfigure}
      \begin{subfigure}[b]{.49\textwidth}
          \centering
          \includegraphics[width=\linewidth]{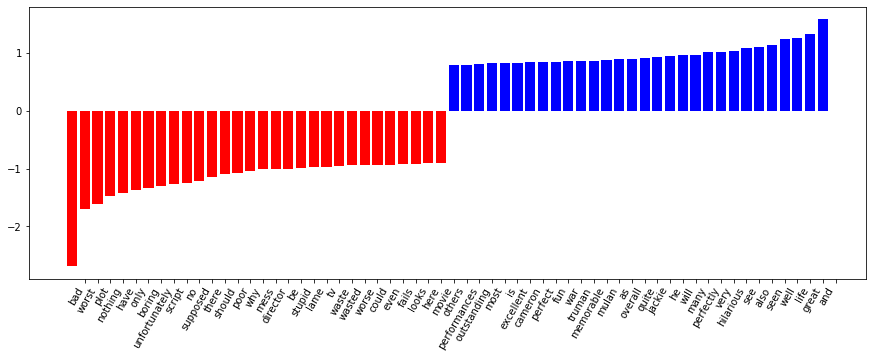}
          \caption{20\% noise in non-rationales}
          
      \end{subfigure} \\
     \begin{subfigure}[b]{.49\textwidth}
         \centering
         \includegraphics[width=\textwidth]{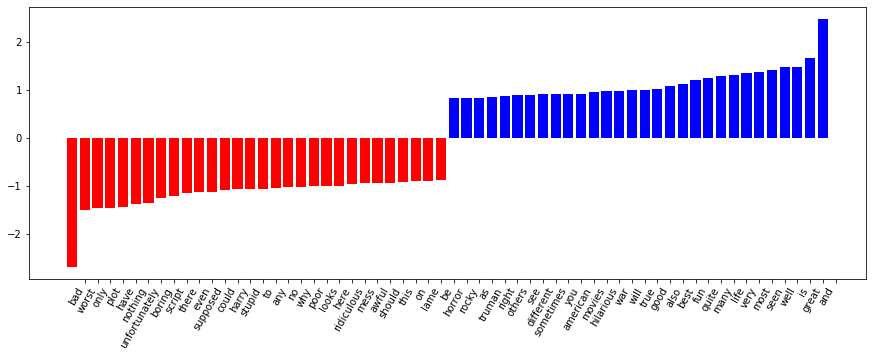}
         \caption{40\% noise in rationales}
         
     \end{subfigure}
     \begin{subfigure}[b]{.49\textwidth}
         \centering
         \includegraphics[width=\textwidth]{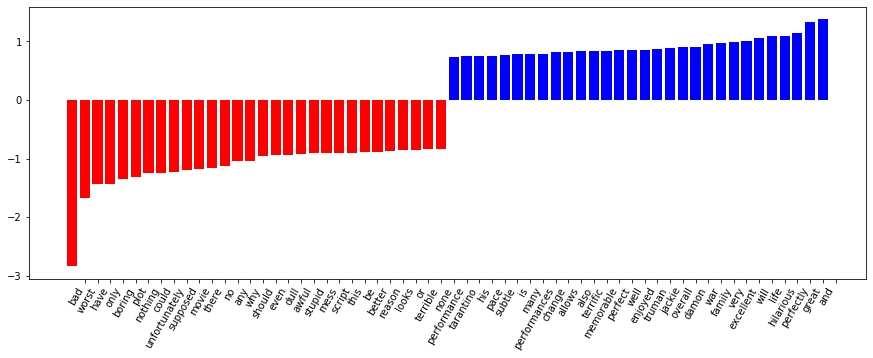}
         \caption{40\% noise in non-rationales}
         
     \end{subfigure}\\
     \begin{subfigure}[b]{.49\textwidth}
         \centering
         \includegraphics[width=\textwidth]{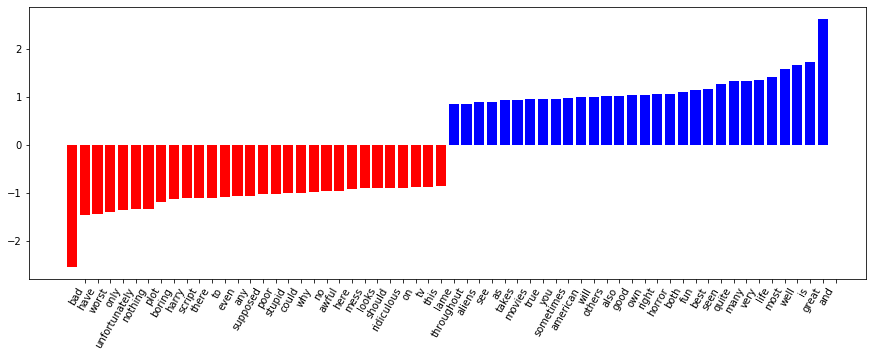}
         \caption{60\% noise in rationales}
         
     \end{subfigure}
     \begin{subfigure}[b]{.49\textwidth}
         \centering
         \includegraphics[width=\textwidth]{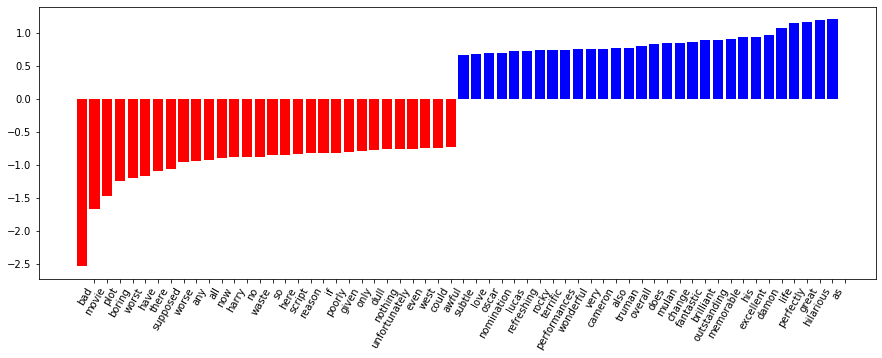}
         \caption{60\% noise in non-rationales}
         
     \end{subfigure}\\
     \begin{subfigure}[b]{.49\textwidth}
         \centering
         \includegraphics[width=\textwidth]{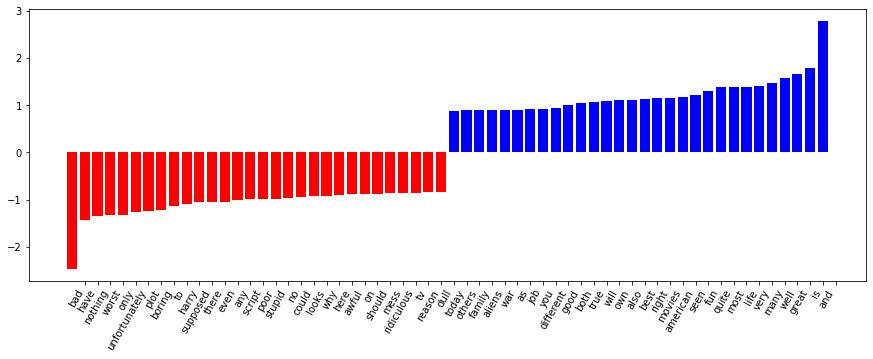}
         \caption{80\% noise in rationales}
         
     \end{subfigure}
     \begin{subfigure}[b]{.49\textwidth}
         \centering
         \includegraphics[width=\textwidth]{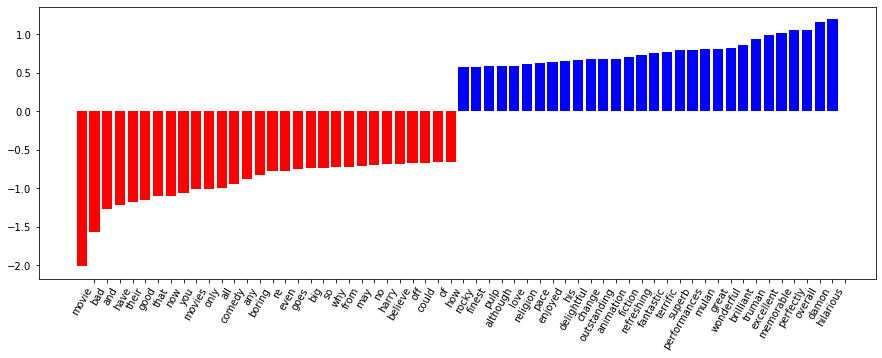}
         \caption{80\% noise in non-rationales}
         
     \end{subfigure}\\
     \begin{subfigure}[b]{.49\textwidth}
         \centering
         \includegraphics[width=\textwidth]{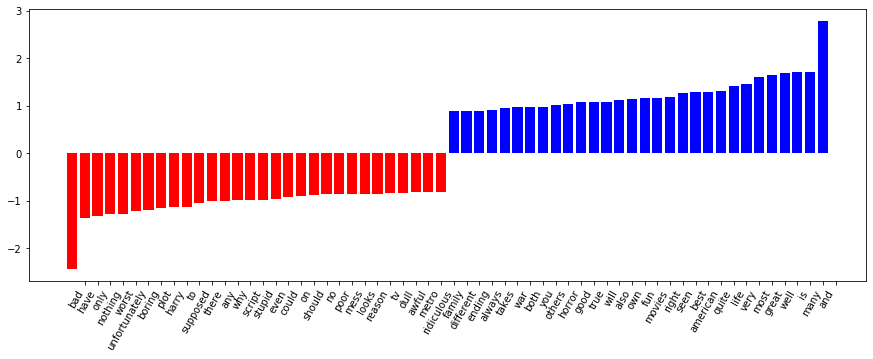}
         \caption{100\% noise in rationales}
         
     \end{subfigure}
     \begin{subfigure}[b]{.49\textwidth}
         \centering
         \includegraphics[width=\textwidth]{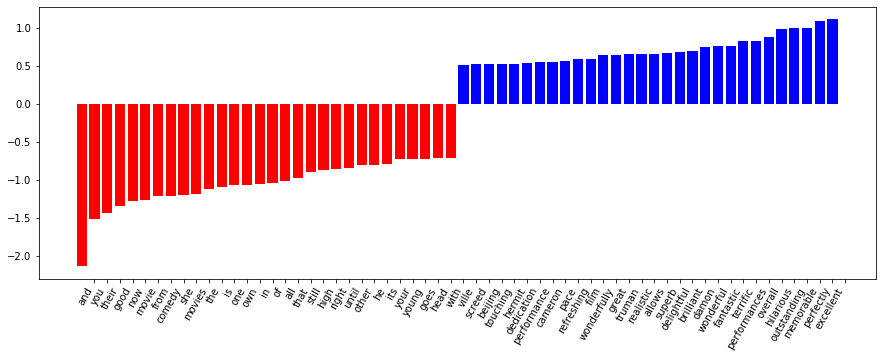}
         \caption{100\% noise in non-rationales}
         
     \end{subfigure}\\
\caption{
Most important features learned by an SVM classifier trained on TF-IDF bag of words. Rationales are identified as tokens attended upon by a BiLSTM with Self Attention model.
\label{zaidan_attention}}
\end{figure*}

\begin{figure*}[t!]
    \begin{subfigure}[b]{\textwidth}
         \centering
         \includegraphics[width=0.51\linewidth]{figures/Zaidan/human/orig.png}
         \caption{Trained on the original dataset \citep{zaidan2007using}}
         
     \end{subfigure}
    \begin{subfigure}[b]{.49\textwidth}
         \centering
         \includegraphics[width=\linewidth]{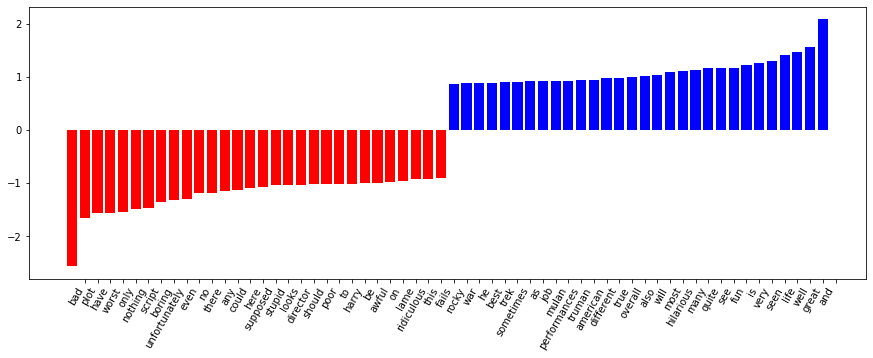}
         \caption{20\% noise in rationales}
         
     \end{subfigure}
      \begin{subfigure}[b]{.49\textwidth}
          \centering
          \includegraphics[width=\linewidth]{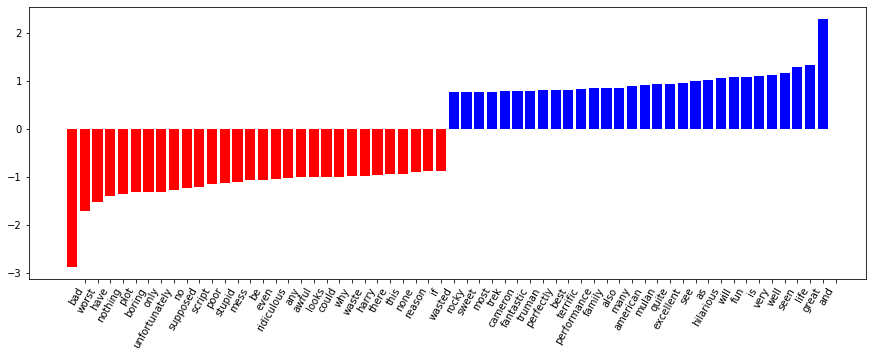}
          \caption{20\% noise in non-rationales}
          
      \end{subfigure} \\
     \begin{subfigure}[b]{.49\textwidth}
         \centering
         \includegraphics[width=\textwidth]{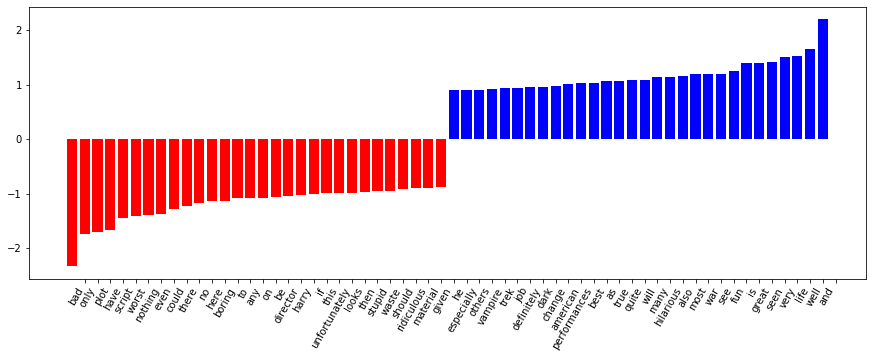}
         \caption{40\% noise in rationales}
         
     \end{subfigure}
     \begin{subfigure}[b]{.49\textwidth}
         \centering
         \includegraphics[width=\textwidth]{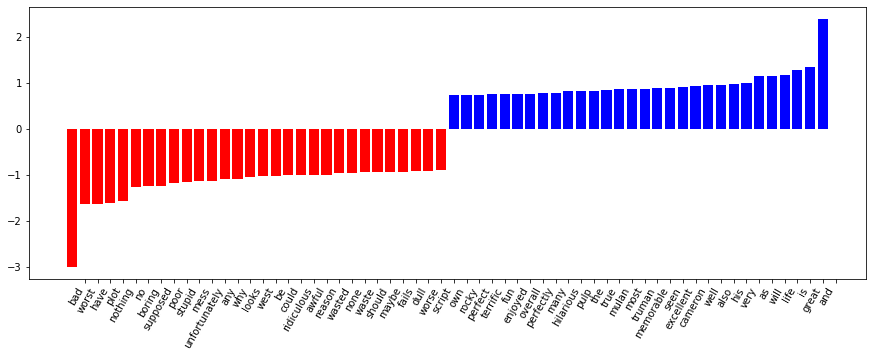}
         \caption{40\% noise in non-rationales}
         
     \end{subfigure}\\
     \begin{subfigure}[b]{.49\textwidth}
         \centering
         \includegraphics[width=\textwidth]{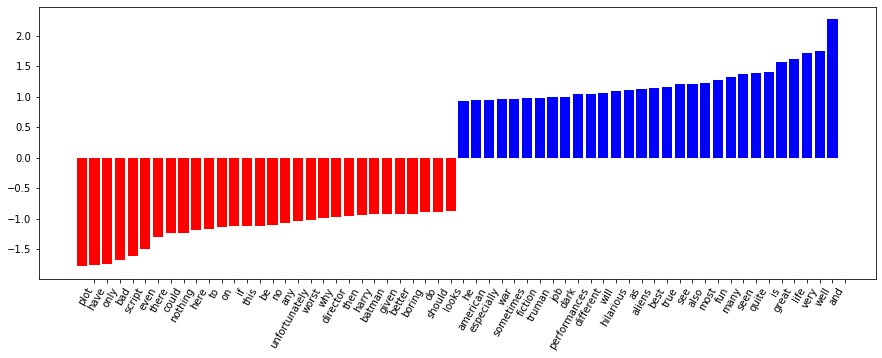}
         \caption{60\% noise in rationales}
         
     \end{subfigure}
     \begin{subfigure}[b]{.49\textwidth}
         \centering
         \includegraphics[width=\textwidth]{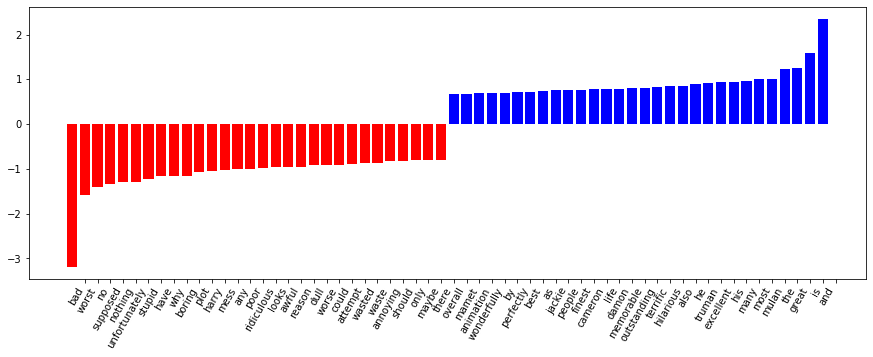}
         \caption{60\% noise in non-rationales}
         
     \end{subfigure}\\
     \begin{subfigure}[b]{.49\textwidth}
         \centering
         \includegraphics[width=\textwidth]{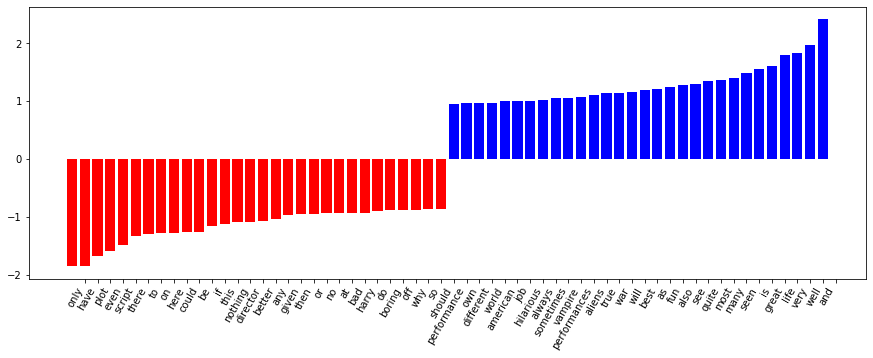}
         \caption{80\% noise in rationales}
         
     \end{subfigure}
     \begin{subfigure}[b]{.49\textwidth}
         \centering
         \includegraphics[width=\textwidth]{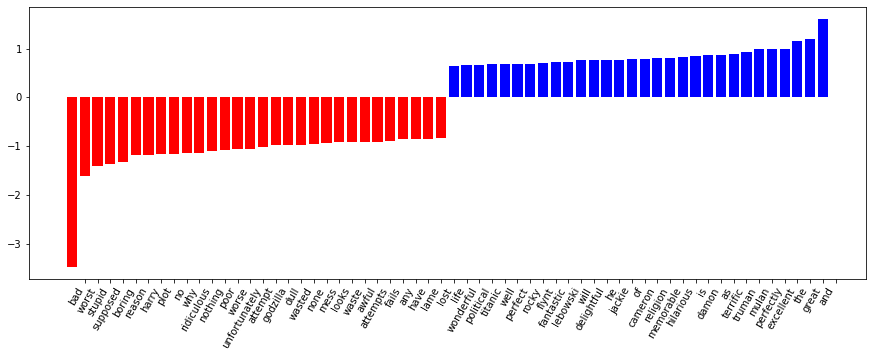}
         \caption{80\% noise in non-rationales}
         
     \end{subfigure}\\
     \begin{subfigure}[b]{.49\textwidth}
         \centering
         \includegraphics[width=\textwidth]{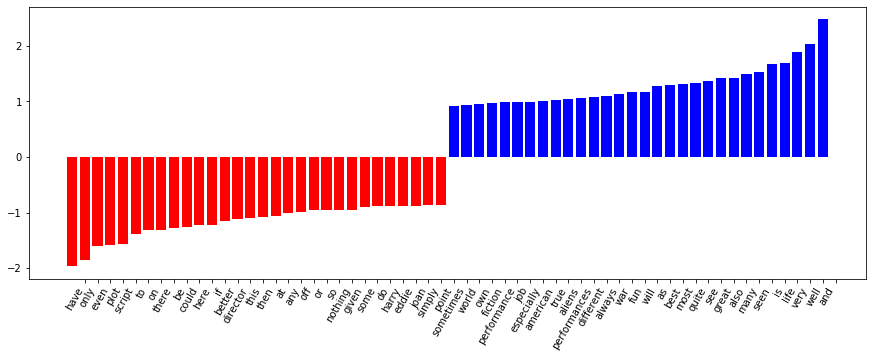}
         \caption{100\% noise in rationales}
         
     \end{subfigure}
     \begin{subfigure}[b]{.49\textwidth}
         \centering
         \includegraphics[width=\textwidth]{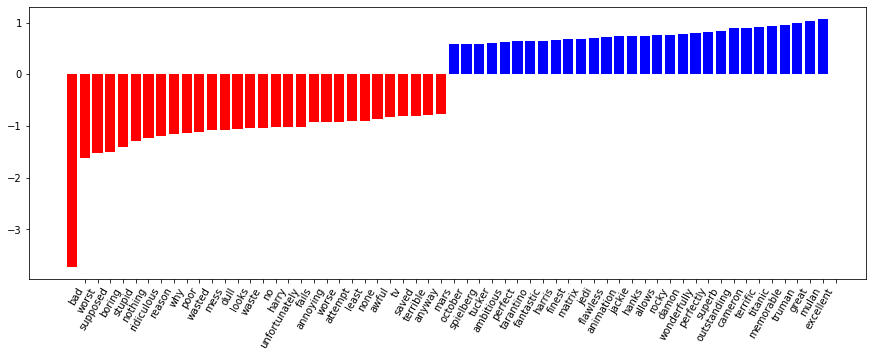}
         \caption{100\% noise in non-rationales}
         
     \end{subfigure}\\
\caption{
Most important features learned by an SVM classifier trained on TF-IDF bag of words. Rationales are identified as tokens marked by the AllenNLP Saliency Interpreter.
\label{zaidan_salience}}
\end{figure*}

\begin{table*}[!th]
  \begin{center}
  \renewcommand{\arraystretch}{1.2}
  \caption{Accuracy of various models for sentiment analysis trained with various datasets. O refers to the in-sample test set from \citet{kaushik2020learning} whereas R refers to the counterfactually revised counterparts of the same.
  \label{tab:style_transfer}}
  \begin{tabular}{ l c c c c c c c c}
    \toprule
    Training data & \multicolumn{2}{c}{SVM} & \multicolumn{2}{c}{NB} & \multicolumn{2}{c}{BiLSTM w/ SA} & \multicolumn{2}{c}{BERT} \\
    \midrule
    & O & R & O & R & O & R & O & R  \\
    \midrule
    Orig. ($1.7k$) & $\textbf{80.0}$& $51.0$&$\textbf{74.9}$ &$47.3$& $\textbf{78.0}$ & $49.4$ & $\textbf{87.4}$ & $82.2$\\
    CRD ($1.7k$) & $58.3$ & $\textbf{91.2}$ & $50.9$ & $\textbf{88.7}$ & $63.8$ & $\textbf{82.0}$ & $80.4$ & $\textbf{90.8}$ \\
    \citet{hu2017toward} & $56.3$ & $68.0$ & $57.1$& $71.1$ & $55.1$ & $67.0$ & $66.3$ & $74.4$\\
    \citet{li2018delete} & $41.3$ & $54.1$ & $37.8$& $58.0$ & $49.9$ & $49.4$ & $37.9$ & $55.9$\\
    \citet{sudhakar2019transforming} & $47.1$ & $55.9$ & $42.6$& $58.8$ & $49.9$ & $49.4$ & $43.7$ & $51.8$\\
    \citet{madaan2020politeness} & $61.2$ & $77.3$ & $50.2$& $75.2$ & $59.3$ & $69.7$ & $70.6$ & $81.6$\\
    \midrule
    CAD ($3.4k$) & $83.7$ & $\textbf{87.3}$ & $\textbf{86.1}$& $\textbf{91.2}$ & $\textbf{80.3}$ & $\textbf{84.8}$ & $88.5$ & $\textbf{95.1}$ \\
    Orig. \& \citeauthor{hu2017toward} ($3.4k$) & $82.1$ & $66.4$ & $81.5$& $55.1$ & $76.4$ & $63.9$ & $87.1$ & $89.3$ \\
    Orig. \& \citeauthor{li2018delete} ($3.4k$) & $73.3$ & $55.7$ & $77.9$& $53.3$ & $69.5$ & $53.3$ & $80.3$ & $79.5$ \\
    Orig. \& \citeauthor{sudhakar2019transforming} ($3.4k$) & $74.1$ & $56.1$ & $79.1$& $51.4$ & $71.4$ & $55.7$ & $89.1$ & $90.8$ \\
    Orig. \& \citeauthor{madaan2020politeness} ($3.4k$) & $83.8$ & $65.4$ & $82.1$& $67.4$ & $75.5$ & $64.1$ & $83.5$ & $81.6$ \\
    Orig. ($3.4k$) & $\textbf{85.1}$ & $54.3$ & $82.4$& $48.2$ & $\textbf{80.1}$ & $57.0$ & $\textbf{90.2}$ & $86.1$\\
\bottomrule
  \end{tabular}
  \end{center}
\end{table*}

\begin{table*}[!th]
  \begin{center}
  \caption{Accuracy of BERT trained on subsample of SNLI \citep{deyoung2020eraser} (where number of rationale tokens and non rationale tokens are within 30\% of one another) as noise is injected on human identified \emph{rationales/non-rationales}. RP and RH are Revised Premise and Revised Hypothesis test sets in \citet{kaushik2020learning}. MNLI-M and MNLI-MM are MNLI \citep{williams2018broad} dev sets.
  \label{tab:nli_results}}
  \begin{tabular}{ l c c c c c c c c c c c}
    \toprule
    & \multicolumn{11}{c}{Percent noise added to train data rationales} \\
    Dataset & $0$ & $10$ & $20$ & $30$ & $40$ & $50$ & $60$ & $70$ & $80$ & $90$ & $100$\\
    \midrule
    In-sample test & $90$ & $87.1$ & $83.5$ & $80.3$ & $80.8$ & $78.9$ & $77.8$ & $77.5$ & $73.5$ & $67.9$ & $69.7$\\
    RP & $66.9$ & $66.4$ & $60.4$ & $57.9$ & $56.9$ & $54.4$ & $52.3$ & $51.3$ & $51.4$ & $51.2$ & $51.5$\\
    RH & $79.2$ & $75$ & $69.8$ & $67$ & $66.5$ & $64.2$ & $63.5$ & $65.7$ & $64.9$ & $61.7$ & $61.8$\\
    MNLI-M & $74.1$ & $66.4$ & $61.9$ & $59.8$ & $59.4$ & $57.4$ & $54.5$ & $56.6$ & $55.7$ & $54.7$ & $54.6$\\
    MNLI-MM & $76.1$ & $66.5$ & $61.4$ & $59$ & $58.5$ & $56.5$ & $53.6$ & $56$ & $55.6$ & $54.2$ & $54.4$\\
    \midrule
    & \multicolumn{11}{c}{Percent noise added to train data non-rationales} \\
    Dataset  & $0$ & $10$ & $20$ & $30$ & $40$ & $50$ & $60$ & $70$ & $80$ & $90$ & $100$\\
    \midrule
    In-sample test & $90$ & $88.7$ & $87.5$ & $85.2$ & $85.7$ & $84.2$ & $83.4$ & $82.2$ & $79.3$ & $77.2$ & $74.9$\\
    RP & $66.9$ & $68.3$ & $66.2$ & $62.7$ & $64.5$ & $63.3$ & $62.6$ & $61.7$ & $61.5$ & $61.5$ & $62.5$\\
    RH & $79.2$ & $78.4$ & $77.4$ & $75.9$ & $74.6$ & $73.4$ & $72.8$ & $72.2$ & $73$ & $70.6$ & $70.8$\\
    MNLI-M & $74.1$ & $67.6$ & $66.5$ & $64.4$ & $65.4$ & $62.8$ & $62.6$ & $62.1$ & $61.9$ & $61.8$ & $61.9$\\
    MNLI-MM & $76.1$ & $68$ & $67.6$ & $65.1$ & $65.1$ & $63.1$ & $63.1$ & $62.3$ & $61.6$ & $61.3$ & $61.1$\\
    \bottomrule
  \end{tabular}
  \end{center}
\end{table*}

\end{document}